\documentclass{article} 
\usepackage{iclr2024_conference,times}

\usepackage[T1]{fontenc}
\usepackage[utf8]{inputenc}
\usepackage{mathtools}

\usepackage{amssymb,mathrsfs}
\usepackage{amsthm}
\usepackage{bm}
\usepackage{scalerel}
\usepackage{nicefrac}
\usepackage{microtype} 
\usepackage[shortlabels]{enumitem}
\usepackage{graphicx}
\usepackage{epstopdf}
\DeclareGraphicsExtensions{.eps,.png,.jpg,.pdf}

\usepackage{url}
\usepackage{colortbl}
\usepackage{booktabs}
\usepackage{multirow}
\usepackage{colortbl,xcolor}
\usepackage[normalem]{ulem}
\usepackage{xparse,xstring}
\usepackage{calc}
\usepackage{etoolbox}

\usepackage{array}
\newcolumntype{L}[1]{>{\raggedright\let\newline\\\arraybackslash\hspace{0pt}}m{#1}}
\newcolumntype{C}[1]{>{\centering\let\newline\\\arraybackslash\hspace{0pt}}m{#1}}
\newcolumntype{R}[1]{>{\raggedleft\let\newline\\\arraybackslash\hspace{0pt}}m{#1}}

\makeatletter
\let\MYcaption\@makecaption
\makeatother
\usepackage[font=footnotesize]{subcaption}
\makeatletter
\let\@makecaption\MYcaption
\makeatother

\usepackage{glossaries}
\makeatletter
\sfcode`\.1006

\let\oldgls\gls
\let\oldglspl\glspl

\newcommand\fussy@ifnextchar[3]{%
	\let\reserved@d=#1%
	\def\reserved@a{#2}%
	\def\reserved@b{#3}%
	\futurelet\@let@token\fussy@ifnch}
\def\fussy@ifnch{%
	\ifx\@let@token\reserved@d
		\let\reserved@c\reserved@a
	\else
		\let\reserved@c\reserved@b
	\fi
	\reserved@c}

\renewcommand{\gls}[1]{%
\oldgls{#1}\fussy@ifnextchar.{\@checkperiod}{\@}}
\renewcommand{\glspl}[1]{%
\oldglspl{#1}\fussy@ifnextchar.{\@checkperiod}{\@}}

\newcommand{\@checkperiod}[1]{%
	\ifnum\sfcode`\.=\spacefactor\else#1\fi
}

\robustify{\gls}
\robustify{\glspl}
\makeatother

\newacronym{wrt}{w.r.t.}{with respect to}
\newacronym{RHS}{R.H.S.}{right-hand side}
\newacronym{LHS}{L.H.S.}{left-hand side}
\newacronym{iid}{i.i.d.}{independent and identically distributed}
\newacronym{SOTA}{SOTA}{state-of-the-art}

\usepackage{float}

\ifx\notloadhyperref\undefined
	\ifx\loadbibentry\undefined
		\usepackage[hidelinks,hypertexnames=false]{hyperref}
	\else
		\usepackage{bibentry}
		\makeatletter\let\saved@bibitem\@bibitem\makeatother
		\usepackage[hidelinks,hypertexnames=false]{hyperref}
		\makeatletter\let\@bibitem\saved@bibitem\makeatother
	\fi
\else
	\ifx\loadbibentry\undefined
		\relax
	\else
		\usepackage{bibentry}
	\fi
\fi

\usepackage[capitalize]{cleveref}
\crefname{equation}{}{}
\Crefname{equation}{}{}
\crefname{claim}{claim}{claims}
\crefname{step}{step}{steps}
\crefname{line}{line}{lines}
\crefname{condition}{condition}{conditions}
\crefname{dmath}{}{}
\crefname{dseries}{}{}
\crefname{dgroup}{}{}

\crefname{Problem}{Problem}{Problems}
\crefformat{Problem}{Problem~#2#1#3}
\crefrangeformat{Problem}{Problems~#3#1#4 to~#5#2#6}

\crefname{Theorem}{Theorem}{Theorems}
\crefname{Corollary}{Corollary}{Corollaries}
\crefname{Proposition}{Proposition}{Propositions}
\crefname{Lemma}{Lemma}{Lemmas}
\crefname{Definition}{Definition}{Definitions}
\crefname{Example}{Example}{Examples}
\crefname{Assumption}{Assumption}{Assumptions}
\crefname{Remark}{Remark}{Remarks}
\crefname{Rem}{Remark}{Remarks}
\crefname{remarks}{Remarks}{Remarks}
\crefname{Appendix}{Appendix}{Appendices}
\crefname{Supplement}{Supplement}{Supplements}
\crefname{Exercise}{Exercise}{Exercises}
\crefname{Theorem_A}{Theorem}{Theorems}
\crefname{Corollary_A}{Corollary}{Corollaries}
\crefname{Proposition_A}{Proposition}{Propositions}
\crefname{Lemma_A}{Lemma}{Lemmas}
\crefname{Definition_A}{Definition}{Definitions}

\usepackage{crossreftools}
\ifx\notloadhyperref\undefined
\else
	\relax
\fi

\usepackage{algorithm}
\usepackage{algpseudocode}

\ifx\loadbreqn\undefined
	\relax
\else
	\usepackage{breqn}
\fi

\makeatletter
\def\cleartheorem#1{%
    \expandafter\let\csname#1\endcsname\relax
    \expandafter\let\csname c@#1\endcsname\relax
}
\def\clearthms#1{ \@for\tname:=#1\do{\cleartheorem\tname} }
\makeatother

\ifx\renewtheorem\undefined
	\ifx\useTheoremCounter\undefined
		\newtheorem{Theorem}{Theorem}
		\newtheorem{Corollary}{Corollary}
		\newtheorem{Proposition}{Proposition}
		
	\else
		\newtheorem{Theorem}{Theorem}
		\newtheorem{Corollary}[Theorem]{Corollary}
		
	\fi

	\newtheorem{Remark}{Remark}

\fi

\theoremstyle{remark}

\theoremstyle{plain}

\newcommand{\qednew}{\nobreak \ifvmode \relax \else
		\ifdim\lastskip<1.5em \hskip-\lastskip
			\hskip1.5em plus0em minus0.5em \fi \nobreak
		\vrule height0.75em width0.5em depth0.25em\fi}



\newcommand{\nn}{\nonumber\\ }

\NewDocumentCommand{\movedownsub}{e{^_}}{%
	\IfNoValueTF{#1}{%
		\IfNoValueF{#2}{^{}}
	}{%
		^{#1}
	}%
	\IfNoValueF{#2}{_{#2}}
}

\let\latexchi\chi
\RenewDocumentCommand{\chi}{}{\latexchi\movedownsub}

\newcommand{\Real}{\mathbb{R}}

\newcommand{\calF}{\mathcal{F}}

\newcommand{\calI}{\mathcal{I}}

\newcommand{\calL}{\mathcal{L}}

\newcommand{\calV}{\mathcal{V}}

\newcommand{\bA}{\mathbf{A}}

\newcommand{\bD}{\mathbf{D}}

\newcommand{\bI}{\mathbf{I}}

\newcommand{\bJ}{\mathbf{J}}

\newcommand{\bL}{\mathbf{L}}

\newcommand{\bP}{\mathbf{P}}

\newcommand{\bR}{\mathbf{R}}

\newcommand{\bS}{\mathbf{S}}

\newcommand{\bV}{\mathbf{V}}

\newcommand{\bW}{\mathbf{W}}
\newcommand{\bx}{\mathbf{x}}
\newcommand{\bX}{\mathbf{X}}

\newcommand{\bY}{\mathbf{Y}}

\DeclareSymbolFont{bsfletters}{OT1}{cmss}{bx}{n}
\DeclareSymbolFont{ssfletters}{OT1}{cmss}{m}{n}
\DeclareMathSymbol{\bsfGamma}{0}{bsfletters}{'000}
\DeclareMathSymbol{\ssfGamma}{0}{ssfletters}{'000}
\DeclareMathSymbol{\bsfDelta}{0}{bsfletters}{'001}
\DeclareMathSymbol{\ssfDelta}{0}{ssfletters}{'001}
\DeclareMathSymbol{\bsfTheta}{0}{bsfletters}{'002}
\DeclareMathSymbol{\ssfTheta}{0}{ssfletters}{'002}
\DeclareMathSymbol{\bsfLambda}{0}{bsfletters}{'003}
\DeclareMathSymbol{\ssfLambda}{0}{ssfletters}{'003}
\DeclareMathSymbol{\bsfXi}{0}{bsfletters}{'004}
\DeclareMathSymbol{\ssfXi}{0}{ssfletters}{'004}
\DeclareMathSymbol{\bsfPi}{0}{bsfletters}{'005}
\DeclareMathSymbol{\ssfPi}{0}{ssfletters}{'005}
\DeclareMathSymbol{\bsfSigma}{0}{bsfletters}{'006}
\DeclareMathSymbol{\ssfSigma}{0}{ssfletters}{'006}
\DeclareMathSymbol{\bsfUpsilon}{0}{bsfletters}{'007}
\DeclareMathSymbol{\ssfUpsilon}{0}{ssfletters}{'007}
\DeclareMathSymbol{\bsfPhi}{0}{bsfletters}{'010}
\DeclareMathSymbol{\ssfPhi}{0}{ssfletters}{'010}
\DeclareMathSymbol{\bsfPsi}{0}{bsfletters}{'011}
\DeclareMathSymbol{\ssfPsi}{0}{ssfletters}{'011}
\DeclareMathSymbol{\bsfOmega}{0}{bsfletters}{'012}
\DeclareMathSymbol{\ssfOmega}{0}{ssfletters}{'012}

\newcommand{\bpi}{\bm{\pi}}

\newcommand{\bPhi}{\bm{\Phi}}

\makeatletter
\newcommand*\rel@kern[1]{\kern#1\dimexpr\macc@kerna}
\newcommand*\widebar[1]{%
  \begingroup
  \def\mathaccent##1##2{%
    \rel@kern{0.8}%
    \overline{\rel@kern{-0.8}\macc@nucleus\rel@kern{0.2}}%
    \rel@kern{-0.2}%
  }%
  \macc@depth\@ne
  \let\math@bgroup\@empty \let\math@egroup\macc@set@skewchar
  \mathsurround\z@ \frozen@everymath{\mathgroup\macc@group\relax}%
  \macc@set@skewchar\relax
  \let\mathaccentV\macc@nested@a
  \macc@nested@a\relax111{#1}%
  \endgroup
}
\makeatother

\DeclareMathOperator{\var}{var}

\DeclareMathOperator{\cov}{cov}

\newcommand{\ifbcdot}[1]{\ifblank{#1}{\cdot}{#1}}

\DeclarePairedDelimiterX\abs[1]{\lvert}{\rvert}{\ifbcdot{#1}}
\DeclarePairedDelimiterX\parens[1]{(}{)}{\ifbcdot{#1}}
\DeclarePairedDelimiterX\brk[1]{[}{]}{\ifbcdot{#1}}
\DeclarePairedDelimiterX\braces[1]{\{}{\}}{\ifbcdot{#1}}
\DeclarePairedDelimiterX\angles[1]{\langle}{\rangle}{\ifblank{#1}{\cdot,\cdot}{#1}}
\DeclarePairedDelimiterX\ip[2]{\langle}{\rangle}{\ifbcdot{#1},\ifbcdot{#2}}
\DeclarePairedDelimiterX\norm[1]{\lVert}{\rVert}{\ifbcdot{#1}}
\DeclarePairedDelimiterX\ceil[1]{\lceil}{\rceil}{\ifbcdot{#1}}
\DeclarePairedDelimiterX\floor[1]{\lfloor}{\rfloor}{\ifbcdot{#1}}

\DeclareFontFamily{U}{matha}{\hyphenchar\font45}
\DeclareFontShape{U}{matha}{m}{n}{
      <5> <6> <7> <8> <9> <10> gen * matha
      <10.95> matha10 <12> <14.4> <17.28> <20.74> <24.88> matha12
      }{}
\DeclareSymbolFont{matha}{U}{matha}{m}{n}
\DeclareFontSubstitution{U}{matha}{m}{n}

\DeclareFontFamily{U}{mathx}{\hyphenchar\font45}
\DeclareFontShape{U}{mathx}{m}{n}{
      <5> <6> <7> <8> <9> <10>
      <10.95> <12> <14.4> <17.28> <20.74> <24.88>
      mathx10
      }{}
\DeclareSymbolFont{mathx}{U}{mathx}{m}{n}
\DeclareFontSubstitution{U}{mathx}{m}{n}

\DeclareMathDelimiter{\vvvert}{0}{matha}{"7E}{mathx}{"17}
\DeclarePairedDelimiterX\vertiii[1]{\vvvert}{\vvvert}{\ifbcdot{#1}}

\DeclarePairedDelimiterXPP\trace[1]{\operatorname{Tr}}{(}{)}{}{\ifbcdot{#1}} 
\DeclarePairedDelimiterXPP\col[1]{\operatorname{col}}{\{}{\}}{}{\ifbcdot{#1}} 
\DeclarePairedDelimiterXPP\row[1]{\operatorname{row}}{\{}{\}}{}{\ifbcdot{#1}} 
\DeclarePairedDelimiterXPP\erf[1]{\operatorname{erf}}{(}{)}{}{\ifbcdot{#1}}
\DeclarePairedDelimiterXPP\erfc[1]{\operatorname{erfc}}{(}{)}{}{\ifbcdot{#1}}
\DeclarePairedDelimiterXPP\KLD[2]{D}{(}{)}{}{\ifbcdot{#1}\, \delimsize\|\, \ifbcdot{#2}} 
\DeclarePairedDelimiterXPP\op[2]{\operatorname{#1}}{(}{)}{}{#2} 

\newcommand{\T}{^{\mkern-1.5mu\mathop\intercal}}
\newcommand{\ud}{\,\mathrm{d}} 

\DeclarePairedDelimiterXPP\indicate[1]{{\bf 1}}{\{}{\}}{}{\ifbcdot{#1}}

\NewDocumentCommand\ofrac{s m}{%
	\IfBooleanTF#1%
	{\dfrac{1}{#2}}%
	{\frac{1}{#2}}%
}
\NewDocumentCommand\ddfrac{s m m}{%
	\IfBooleanTF#1%
	{\dfrac{\mathrm{d} {#2}}{\mathrm{d} {#3}}}%
	{\frac{\mathrm{d} {#2}}{\mathrm{d} {#3}}}%
}
\NewDocumentCommand\ppfrac{s m m}{%
	\IfBooleanTF#1%
	{\dfrac{\partial {#2}}{\partial {#3}}}%
	{\frac{\partial {#2}}{\partial {#3}}}%
}

\providecommand\given{}

\DeclarePairedDelimiterX\Set[2]\{\}{%
\renewcommand\given{\SetSymbol[\delimsize]{#1}}
#2
}
\DeclarePairedDelimiterX\Setc[1]\{\}{%
\renewcommand\given{\SetSymbol{:}}
#1
}

\NewDocumentCommand\set{s o m}{%
	\IfBooleanTF#1%
	{\IfValueTF{#2}{\Set*{#2}{#3}}{\Setc*{#3}}}%
	{\IfValueTF{#2}{\Set{#2}{#3}}{\Setc{#3}}}%
}

\NewDocumentCommand{\evalat}{ s O{\big} m e{_^} }{%
\IfBooleanTF{#1}%
{\left. #3 \right|}{#3#2|}%
\IfValueT{#4}{_{#4}}%
\IfValueT{#5}{^{#5}}%
}

\providecommand\given{}
\DeclarePairedDelimiterXPP\cprob[1]{}(){}{
\renewcommand\given{\nonscript\,\delimsize\vert\allowbreak\nonscript\,\mathopen{}}%
#1%
}
\DeclarePairedDelimiterXPP\cexp[1]{}[]{}{
\renewcommand\given{\nonscript\,\delimsize\vert\allowbreak\nonscript\,\mathopen{}}%
#1%
}

\DeclareDocumentCommand \P { s e{_^} d() g } {%
	\mathbb{P}%
	\IfBooleanTF{#1}%
		{
			\IfValueT{#2}{_{#2}}%
			\IfValueT{#3}{^{#3}}%
			\IfValueTF{#5}{\cprob{#4 \given #5}}{\IfValueT{#4}{\cprob{#4}}}%
		}%
		{
			\IfValueT{#2}{_{#2}}%
			\IfValueT{#3}{^{#3}}%
			\IfValueTF{#5}{\cprob*{#4 \given #5}}{\IfValueT{#4}{\cprob*{#4}}}%
		}%
}

\DeclareDocumentCommand \E { s e{_^} o g } {%
	\mathbb{E}%
	\IfBooleanTF{#1}%
		{
			\IfValueT{#2}{_{#2}}%
			\IfValueT{#3}{^{#3}}%
			\IfValueTF{#5}{\cexp{#4 \given #5}}{\IfValueT{#4}{\cexp{#4}}}%
		}%
		{
			\IfValueT{#2}{_{#2}}%
			\IfValueT{#3}{^{#3}}%
			\IfValueTF{#5}{\cexp*{#4 \given #5}}{\IfValueT{#4}{\cexp*{#4}}}%
		}%
}

\DeclareDocumentCommand \Var { s e{_^} d() g } {%
	\var%
	\IfBooleanTF{#1}%
		{
			\IfValueT{#2}{_{#2}}%
			\IfValueT{#3}{^{#3}}%
			\IfValueTF{#5}{\cprob{#4 \given #5}}{\IfValueT{#4}{\cprob{#4}}}%
		}%
		{
			\IfValueT{#2}{_{#2}}%
			\IfValueT{#3}{^{#3}}%
			\IfValueTF{#5}{\cprob*{#4 \given #5}}{\IfValueT{#4}{\cprob*{#4}}}%
		}%
}

\DeclareDocumentCommand \Cov { s e{_^} d() g } {%
	\cov%
	\IfBooleanTF{#1}%
		{
			\IfValueT{#2}{_{#2}}%
			\IfValueT{#3}{^{#3}}%
			\IfValueTF{#5}{\cprob{#4 \given #5}}{\IfValueT{#4}{\cprob{#4}}}%
		}%
		{
			\IfValueT{#2}{_{#2}}%
			\IfValueT{#3}{^{#3}}%
			\IfValueTF{#5}{\cprob*{#4 \given #5}}{\IfValueT{#4}{\cprob*{#4}}}%
		}%
}

\ExplSyntaxOn
\NewDocumentCommand \dist {m o o} {%
\mathrm{#1}\left(%
	\IfValueT{#3}{%
		\tl_if_blank:nTF{ #3 }{\cdot\, \middle|\, }{#3\, \middle|\, }%
	}
	\IfValueT{#2}{#2}%
\right)%
}
\ExplSyntaxOff

\NewDocumentCommand {\cbrace} {t+ D[]{black} D(){\widthof{#5}} m m } {%
	\begingroup%
		\color{#2}
		\IfBooleanTF{#1}{%
			\overbrace{#4}^%
		}{
			\underbrace{#4}_%
		}%
		{\parbox[c]{#3}{\centering\footnotesize{#5}}}%
	\endgroup%
}

\let\oldforall\forall
\renewcommand{\forall}{\oldforall \, }

\let\oldexist\exists
\renewcommand{\exists}{\oldexist \, }

\makeatletter

\newcommand{\rankcolor}[2]{%
	\expandafter\renewcommand\csname #1\endcsname[1]{%
		\ifblank{##1}{%
			{\color{#2} \textbf{#2}}%
		}{%
			\ifmmode
				\textcolor{#2}{\bm{##1}}%
			\else%
				{\color{#2} \textbf{##1}}%
			\fi	
		}%
	}
}

\providecommand{\first}{}
\providecommand{\second}{}

\rankcolor{first}{red}
\rankcolor{second}{blue}
\rankcolor{third}{cyan}
\makeatother

\graphicspath{{./Figures/}{./figures/}}
\DeclareDocumentCommand{\includeCroppedPdf}{ o O{./Figures/} m }{
	\IfFileExists{#2#3-crop.pdf}{}{%
		\immediate\write18{pdfcrop #2#3.pdf #2#3-crop.pdf}}%
	\includegraphics[#1]{#2#3-crop.pdf}
}

\makeatletter
\newcommand*{\addFileDependency}[1]{
  \typeout{(#1)}
  \@addtofilelist{#1}
  \IfFileExists{#1}{}{\typeout{No file #1.}}
}
\makeatother

\definecolor{gray90}{gray}{0.9}
\def\colorlist{red,blue,brown,cyan,darkgray,gray,lightgray,green,lime,magenta,olive,orange,pink,purple,teal,violet,white,yellow}

\makeatletter
\def\startcomment{[}
\ifx\nohighlights\undefined
	\newcommand{\createcolor}[1]{%
			\expandafter\newcommand\csname #1\endcsname[1]{{\color{#1} ##1}}%
	}
	\newcommand{\msout}[1]{\text{\color{green} \sout{\ensuremath{#1}}}}
	\newcommand{\del}[1]{{\color{green}\ifmmode \msout{#1}\else\sout{#1}\fi}}
\else
	\newcommand{\createcolor}[1]{%
			\expandafter\newcommand\csname #1\endcsname[1]{%
				\noexpandarg%
				\StrChar{##1}{1}[\firstletter]%
				\if\firstletter\startcomment%
					\relax
				\else%
					##1
				\fi
			}%
	}
	\newcommand{\msout}[1]{}
	\newcommand{\del}[1]{}
\fi

\def\@tempa#1,{%
    \ifx\relax#1\relax\else
        \createcolor{#1}%
        \expandafter\@tempa
    \fi
}
\expandafter\@tempa\colorlist,\relax,
\makeatother

\newcommand{\hhide}[1]{}

\newcommand{\txp}[2]{\texorpdfstring{#1}{#2}}

\ifx\diagnoselabel\undefined
	\relax
\else
	\makeatletter
	\def\@testdef #1#2#3{%
		\def\reserved@a{#3}\expandafter \ifx \csname #1@#2\endcsname
			\reserved@a  \else
			\typeout{^^Jlabel #2 changed:^^J%
				\meaning\reserved@a^^J%
				\expandafter\meaning\csname #1@#2\endcsname^^J}%
			\@tempswatrue \fi}
	\makeatother
\fi

\newcommand{\tb}[1]{\textbf{#1}}

\usepackage{tikz}
\usepackage{adjustbox}
\usepackage{wrapfig}

\title{Unleashing the Potential of Fractional Calculus in Graph Neural Networks with FROND}

\author{Qiyu Kang$^1$\thanks{First two authors contributed equally. $^\dagger$Correspondence to: Qiyu Kang <kang0080@e.ntu.edu.sg>. }\footnotemark[1]\; $^\dagger$\;, Kai Zhao$^1$\footnotemark[1]\;, Qinxu Ding$^2$, Feng Ji$^1$, Xuhao Li$^3$,  Wenfei Liang$^1$, Yang Song$^4$, \\
\textbf{Wee Peng Tay$^1$}\\
$^1$Nanyang Technological University 
$^2$Singapore University of Social Sciences\\
$^3$Anhui University
$^4$C3 AI, Singapore
 }

\iclrfinalcopy
\begin{document}

\maketitle

\begin{abstract}
We introduce the FRactional-Order graph Neural Dynamical network (FROND), a new continuous graph neural network (GNN) framework. Unlike traditional continuous GNNs that rely on integer-order differential equations, FROND employs the Caputo fractional derivative to leverage the non-local properties of fractional calculus. This approach enables the capture of long-term dependencies in feature updates, moving beyond the Markovian update mechanisms in conventional integer-order models and offering enhanced capabilities in graph representation learning. 
We offer an interpretation of the node feature updating process in FROND from a non-Markovian random walk perspective when the feature updating is particularly governed by a diffusion process.
We demonstrate analytically that oversmoothing can be mitigated in this setting.
Experimentally, we validate the FROND framework by comparing the fractional adaptations of various established integer-order continuous GNNs, demonstrating their consistently improved performance and underscoring the framework's potential as an effective extension to enhance traditional continuous GNNs.
The code is available at \url{https://github.com/zknus/ICLR2024-FROND}.
\end{abstract}

\section{Introduction} \label{sec.intro}
Graph Neural Networks (GNNs) have excelled in diverse domains, e.g., chemistry \citep{yueBio2019}, finance \citep{AshoorNC2020}, and social media \citep{kipf2017semi,ZhangTKDE2022,WuTNNLS2021}.  The message passing scheme \citep{feng2022khopmessage}, where features are aggregated along edges and iteratively propagated through layers, is crucial for the success of GNNs.
Over the past few years, numerous types of GNNs have been proposed, including Graph Convolutional Networks (GCN) \citep{kipf2017semi}, Graph Attention Networks (GAT) \citep{velickovic2018graph}, and GraphSAGE \citep{hamilton2017inductive}. Recent works, such as \citep{chamberlain2021grand,thorpe2022grand++,rusch2022graph,SonKanWan:C22,choi2023gread,ZhaKanSon:C23,KanZhaSon:C23}, have incorporated various continuous dynamical processes to propagate information over graph nodes, giving rise to a class of continuous GNNs based on integer-order differential equations.
These continuous models have demonstrated notable performance, for instance, in enhancing robustness and addressing heterophilic graphs \citep{han2023continuous}.

Within these integer-order continuous GNNs, the differential operator $\ud^\beta/\ud t^\beta$ has been constrained to \emph{integer values} of $\beta$, primarily 1 or 2. However, over recent decades, the wider scientific community has explored fractional-order differential operators, where $\beta$ can be any \emph{real number}. These expansions have proven pivotal in various applications characterized by non-local and memory-dependent behaviors, with examples including viscoelastic materials \citep{bagley1983theoretical}, anomalous transport mechanisms \citep{gomez2016modeling}, and fractal media \citep{mandelbrot1982fractal}. Unlike conventional integer-order derivatives that measure the function's \emph{instantaneous rate of change} and focus on the local vicinity, fractional-order derivatives \citep{tarasov2011fractional} consider \emph{the entire historical trajectory of the function}. 

We introduce the FRactional-Order graph Neural Dynamical network (FROND) framework, a new approach that broadens the capabilities of traditional integer-order continuous GNNs by incorporating fractional calculus. 
It naturally generalizes the integer-order derivative $\ud^\beta/\ud t^\beta$ in these GNNs to accommodate any positive real number $\beta$. 
This modification gives FROND the ability to incorporate \emph{memory-dependent dynamics} for information propagation and feature updating, enabling refined graph representations and improved performance potentially.
Importantly, this technique assures at least equivalent performance to integer-order models, as setting $\beta$ to integer values reverts the models to their traditional integer-order forms.

Several works like \citep{maskey2023fractional} have combined fractional graph shift operators with integer-order ordinary differential equations (ODEs). These studies are distinct from our research, wherein we focus on incorporating time-fractional derivatives for updating graph node features, modeled as a memory-inclusive dynamical process. Other works like \citep{liu2022regularized} have used fractional calculus in gradient propagation for the training process, which is different from leveraging fractional differential equations (FDEs) in modeling the node feature updating.  We provide a detailed discussion of the differences between FROND and these works in \cref{sec.relatedwork}.

Many real-world graph datasets, such as the World Wide Web, the Internet, and various biological and social networks, are known to exhibit scale-free hierarchical structures. These structures suggest a pervasive self-similarity across different scales, hinting at an underlying fractal behavior \citep{song2005self,kim2007fractality,masters2004fractal}. It has been well-established that dynamical processes with self-similarity on such fractal media are more accurately described using FDEs. For instance, the dispersion of heat or mass over these structures is best modeled using fractional diffusion equations \citep{diaz2022time}. Further investigations have revealed a direct connection between the fractal dimension of these structures and the order $\beta$ in fractional derivatives $\ud^\beta/\ud t^\beta$ \citep{nigmatullin1992fractional,tarasov2011fractional}. This revelation births a compelling insight: the optimal $\beta$ in our models, which may differ from integers, can pave the way for enhanced node classification and potentially unearth insights into the inherent ``fractality'' of graph datasets.

\tb{Main contributions.}
Our objective in this paper is to formulate a generalized fractional-order continuous GNN framework. Our key contributions are summarized as follows:
\begin{itemize}[label=$\bullet$, topsep=-4pt, itemsep=2pt, partopsep=0pt, parsep=0pt,leftmargin=10pt]
    \item We propose a novel, generalized continuous GNN framework that incorporates non-local fractional derivatives $\ud^\beta/\ud t^\beta$. This framework generalizes the prior class of integer-order continuous GNNs, subsuming them as special instances with $\beta$ setting as integers. 
    This approach also lays the groundwork for a diverse new class of GNNs that can accommodate a broad array of learnable memory-dependent feature-updating processes.
    \item We provide an interpretation from the perspective of a non-Markovian graph random walk when the feature-updating dynamics are inspired by the fractional heat diffusion process. Contrasting with the Markovian random walk implicit in traditional integer-order graph neural diffusion models whose convergence to the stationary equilibrium is exponentially swift, we establish that in FROND, convergence follows a slow algebraic rate. This characteristic enhances FROND's ability to mitigate oversmoothing, as verified by our experimental results.
    \item We underscore the compatibility of FROND, emphasizing its capability to be seamlessly integrated to augment the performance of existing integer-order continuous GNNs across diverse datasets. Our exhaustive experiments, encompassing the fractional differential extension of \citep{chamberlain2021grand,thorpe2022grand++,rusch2022graph,SonKanWan:C22,choi2023gread,ZhaKanSon:C23}, substantiate this claim. 
    Through detailed ablation studies, we provide insights into the choice of numerical schemes and parameters.
\end{itemize}

\section{Preliminaries}

In this section, we briefly introduce fractional calculus and integer-order continuous GNNs. For a comprehensive review of fractional calculus, readers are referred to \cref{app.review}.
\subsection{Caputo Fractional Derivative}\label{sec.cap_diff}
The literature offers various fractional derivative definitions, notably by Riemann, Liouville, Chapman, and Caputo \citep{tarasov2011fractional}. Our study leverages the \emph{Caputo} fractional derivative, due to the reasons listed in \cref{ssec.reason}.
The traditional first-order derivative of a scalar function $f(t)$ represents the local rate of change of the function at a point, defined as: $\frac{\ud f(t)}{\ud t}=\lim _{\Delta t \rightarrow 0} \frac{f(t+\Delta t)-f(t)}{\Delta t}$. 
Let $F(s)$ denote the Laplace transform of $f(t)$, assumed to exist on $\left[s_0, \infty\right)$ for some $s_0 \in \mathbb{R}$. Under certain conditions \citep{korn2000mathematical}, the Laplace transform of $\frac{\ud f(t)}{\ud t}$ is given by:
\begin{align}
\mathcal{L}\left\{\frac{\ud f(t)}{\ud t}\right\}=s F(s)-f(0) \label{eq.first_lap}
\end{align}
The Caputo fractional derivative of order $\beta \in (0,1]$ for a function $f(t)$ is defined as follows:
\begin{align} \label{eq.def_fract}
D_t^\beta f(t)=\frac{1}{\Gamma(1-\beta)} \int_0^t(t-\tau)^{-\beta} f^{\prime}(\tau) \ud\tau,
\end{align}
where $\Gamma(\cdot)$ denotes the gamma function, and $f^{\prime}(\tau)$ is the first-order derivative of $f$.
The broader definition for any $\beta>0$ is deferred to \cref{app.review}. 
The Caputo fractional derivative inherently integrates the entire history of the system through the integral term, emphasizing its non-local nature. For $s>\max \left\{0, s_0\right\}$, the Laplace transform of the Caputo fractional derivative is given by \citep{diethelm2010analysis}[Theorem 7.1]:
\begin{align}
\mathcal{L}\left\{D_t^\beta f(t)\right\}=s^\beta F(s)-s^{\beta-1} f(0). \label{eq.fract_lap}
\end{align}
Comparing \cref{eq.first_lap,eq.fract_lap}, it is evident that the Caputo derivative serves as a generalization of the first-order derivative. The alteration in the exponent of $s$ comes from the memory-dependent property in \cref{eq.def_fract}. As $\beta \to 1$, the Laplace transform of the Caputo fractional derivative converges to that of the traditional first-order derivative. When $\beta=1$, $D_t^1f=f^{\prime}$ is uniquely determined through the inverse Laplace transform \citep{Cohen2007}. 

In summary, from the frequency domain using the Laplace transform, we observe that the Caputo fractional derivative can be seen as a natural extension of the traditional first-order derivative. For vector-valued functions, the fractional derivative is defined component-wise for each dimension.

\subsection{Integer-Order Continuous GNNs} 
\label{sec.diff} 
We denote an undirected graph as  $\mathcal{G}=(\mathcal{V}, \bW)$ without self-loops, where $\mathcal{V}$ is the set of $|\mathcal{V}| = N$ nodes.  The feature matrix  $\bX = \left(\left[\bx_{1}\right]\T, \cdots, \left[\bx_{N}\right]\T\right)\T \in \mathbb{R}^{N \times d}$ consists of rows $\bx_{i} \in \mathbb{R}^d$ as node feature vectors and $i$ is the node index. The $N \times N$ matrix $\bW := \left(W_{ij}\right)$ has elements $W_{ij}$ indicating the edge weight between the $i$-th and $j$-th node with $W_{ij} = W_{ji}$. 
The following integer-order continuous GNNs leverage ODEs to facilitate information propagation amongst graph nodes, where features evolve as $\bX(t)$, starting from the initial condition $\bX(0) = \bX$.

\tb{GRAND:} Inspired by the heat diffusion equation, GRAND \citep{chamberlain2021grand} utilizes the following nonlinear autonomous dynamical system:
\begin{align}
\frac{\ud \bX(t)}{\ud t} = (\bA(\bX(t)) - \bI)\bX(t). \label{eq.grand_main} 
\end{align}
where $\bA(\bX(t))\in\Real^{N\times N}$ is a learnable, time-variant attention matrix, calculated using the features $\bX(t)$, and $\bI$ denotes the identity matrix. The feature update outlined in \cref{eq.grand_main} is referred to as the \tb{GRAND-nl} version (due to the nonlinearity in $\bA(\bX(t))$). 
We define $d_i=\sum_{j=1}^n W_{i j}$ and let $\bD$ be a diagonal matrix with $D_{ii} = d_i$. The \emph{random walk Laplacian} is then represented as $\bL = \bI - \bW\bD^{-1}$.
In a simplified context, we employ the following linear dynamical system:
\begin{align}
\frac{\ud \bX(t)}{\ud  t} = ( \bW \bD^{-1} - \bI)\bX(t) = -\bL \bX(t). \label{eq.gra_dif_L}
\end{align}
The feature update process in \cref{eq.gra_dif_L} is the \tb{GRAND-l} version. For implementations of \cref{eq.gra_dif_L}, one may direct set $\mathbf{W D}^{-1}=\mathbf{A}(\mathbf{X}(0))$ as a column-stochastic attention matrix, rather than using a plain weight. Notably, in this time-invariant setting, the attention weight matrix, reliant on the initial node features, stays unchanged throughout the feature evolution period.

\tb{GRAND++} \citep{thorpe2022grand++} adds a source term to GRAND, enhancing learning in scenarios with limited labeled nodes. 
\tb{GraphCON} \citep{rusch2022graph} employs a second-order ODE, which is equivalent to two first-order ODEs, drawing inspiration from oscillator systems.
\tb{CDE} \citep{ZhaKanSon:C23} incorporates convection-diffusion equations into GNNs to address heterophilic graph challenges.
\tb{GREAD} \citep{choi2023gread} introduces a reaction term in the GRAND model, improving its application to heterophilic graphs and formulating a diffusion-reaction equation within GNNs.
The detailed formulation for each model is presented in \cref{ssec.app_odereview} due to space constraints. 

\section{Fractional-Order Graph Neural Dynamical Network}\label{sec.algo}
In this section, we introduce the FROND framework, a novel approach that augments traditional integer-order continuous GNNs by incorporating fractional calculus. We elucidate the fractional counterparts of several well-established integer-order continuous GNNs, including GRAND, GRAND++, GraphCON, CDE, and GREAD, as referenced in \cref{sec.diff}. We provide a detailed study of the fractional extension of GRAND, and present insights into the inherent memory mechanisms in our framework through a random walk interpretation. Our theoretical findings suggest a potential mitigation of oversmoothing due to the model's slow algebraic convergence to stationarity. Subsequently, we outline the numerical FDE solvers required to implement FROND.
\subsection{Framework}
Consider a graph $\mathcal{G}=(\mathcal{V}, \bW)$ 
as defined in \cref{sec.diff}. Analogous to the implementation in traditional integer-order continuous GNNs, a preliminary learnable encoder function $\varphi:\calV\to\Real^d$ that maps each node to a feature vector can be applied. Stacking all the feature vectors together, we obtain $\bX\in\Real^{N\times d}$. Employing the Caputo fractional derivative outlined in \cref{sec.cap_diff}, the information propagation and feature updating dynamics in FROND are characterized by the following FDE:
\begin{align}
D_t^\beta \bX(t)= \calF(\bW,\bX(t)), \quad \beta>0, \label{eq.main_object}
\end{align}
where $\beta$ denotes the fractional order of the derivative, and $\calF$ is a dynamic operator on the graph like the models presented in \cref{sec.diff}. The initial condition for \cref{eq.main_object} is set as $\bX^{[\lceil\beta\rceil-1]}(0) =\ldots = \bX(0)= \bX$ consisting of the preliminary node features, with $\bX^{[i]}(t)$ denoting the $i$-th order derivative and $\lceil \beta \rceil$ is the smallest integer not less than $\beta$, akin to the initial conditions seen in integer-order ODEs.\footnote{See \cref{ssec.supp_capudfe}. We mainly consider $\beta \in (0,1]$ and the initial condition is $\bX(0)= \bX$.}  
Similar to integer-order continuous GNNs, we set an integration time parameter $T$ to get $\bX(T)$. The final node embeddings for downstream tasks are then decoded using a learnable decoder $\psi(\bX(T))$.

When $\beta=1$, \cref{eq.main_object} reverts to the class of integer-order continuous GNNs, with the infinitesimal variation of features dependent only on their present state.
Conversely, when $\beta<1$, the Caputo fractional derivative \cref{eq.def_fract} dictates that the updating process for features encompasses their entire history, not just the present state. 
This paradigm facilitates memory-dependent dynamics in the framework. 

For further insights into memory dependence, readers are directed to \cref{sec.solvers}, which discusses time discretization techniques for numerically solving the system. 
It illustrates how, akin to integer-order neural ODE models, time consistently acts as an analog to the layer index and how the nonlocal properties of fractional derivatives facilitate nontrivial dense or skip connections between layers.
In \cref{eq.random_walk}, when the dynamic operator $\calF$ is designated as the diffusion process in \cref{eq.gra_dif_L}, we offer a memory-dependent \emph{non-Markovian} random walk interpretation of the fractional graph neural diffusion process. Here, as $\beta\rightarrow 1$, the non-Markovian random walk increasingly detaches from the path history, becoming a Markovian walk at $\beta=1$, which is related to the normal diffusion process \citep{thorpe2022grand++}. The parameter $\beta$ provides flexibility to adjust the extent of memorized dynamics embedded in the framework.
From a geometric perspective, as discussed in \cref{sec.intro}, the information propagation dynamics in fractal graph datasets might be more suitably described using FDEs. Choosing a non-integer $\beta$ could reveal the degree of fractality in graph datasets.

\subsubsection{FROND Model Examples}
When $\calF$ in \cref{eq.main_object} is specified to the dynamics depicted in various integer-order continuous GNNs (cf.\ \cref{sec.diff}), we formulate FROND GNN variants such as F-GRAND, F-GRAND++, F-GREAD, F-CDE, and F-GraphCON, serving as fractional differential extensions of the original GNNs.

\tb{F-GRAND}: Mirroring the GRAND model, the fractional-GRAND (F-GRAND) has two versions. The F-GRAND-nl version employs a time-variant FDE as follows:
\begin{align}
D_t^\beta \bX(t) = (\bA(\bX(t)) - \bI)\bX(t), \quad 0<\beta\le 1. \label{eq.frac_grand-nl}
\end{align}
It is computed using $\bX(t)$ and the attention mechanism derived from the Transformer model \citep{vaswani2017attention}. The entries of $\bA(\bX(t))=(a(\bx_i,\bx_j))$ are given by:
\begin{align}
a(\bx_i,\bx_j) = \mathrm{softmax} \left(\left\{\frac{(\mathbf{W}_K \bx_{i}\T )\T \mathbf{W}_Q \bx_{j}\T}{\bar{d}_k}\right\}\right).
\end{align}
In this formulation, $\mathbf{W}_K$ and $\mathbf{W}_Q$ are the learned matrices, and $\bar{d}_k$ signifies a hyperparameter related to the dimensionality of $\bW_K$. In parallel, the F-GRAND-l version stands as the fractional differential extension of \cref{eq.gra_dif_L}:
\vspace{-0.2cm}
\begin{align}
D_t^\beta \bX(t) = - \bL \bX(t), \quad 0<\beta\le 1. \label{eq.frac_gra_dif_L}
\end{align}
 Recall that the initial condition for F-GRAND-nl and F-GRAND-l is $\bX(0)=\bX$ due to $\beta\in(0,1]$.
 
\tb{F-GRAND++, F-GREAD, F-CDE, and F-GraphCON:} Due to space constraints, we direct the reader to \cref{sec.app_moredynamic} for detailed formulations. Succinctly, they represent the fractional differential extensions of GRAND++, GraphCON, CDE, and GREAD. To highlight FROND's compatibility and its potential to enhance the performance of existing integer-order continuous GNNs across a variety of datasets, exhaustive experiments are provided in \cref{sec.exp,sec.app_moredynamic}.

\subsection{Random Walk Perspective of F-GRAND-l} \label{eq.random_walk}
The established Markov interpretation of GRAND-l \cref{eq.gra_dif_L}, as outlined in \citep{thorpe2022grand++}, aligns with F-GRAND-l \cref{eq.frac_gra_dif_L} when $\beta=1$. We herein broaden this interpretation to encompass a non-Markovian random walk that considers the walker's complete path history when $\beta$ is a non-integer, thereby elucidating the memory effects inherent in FROND. 
In contrast to the Markovian walk, whose distribution converges exponentially to equilibrium, our strategy assures algebraic convergence, revealing F-GRAND-l's efficacy in mitigating oversmoothing as evidenced in \cref{sec.over-smooth}.

To begin, we discretize the time domain into time instants as $t_n=n \sigma, \sigma>0, n=0,1,2, \ldots,$ where $\sigma$ is assumed to be small enough to ensure the validity of the approximation.
Let ${\bR(t_n)}$ be a random walk on the graph nodes $\{\bx_j\}_{j=1}^{N}$ that is not necessarily a Markov process and $\bR(t_{n+1})$ may depend on the path history $(\bR(t_0),\bR(t_1),\ldots,\bR(t_n))$ of the random walker. For convenience, we introduce the coefficients $c_k$ for $k\geq1$ and $b_n$ for $n\geq0$ from \citep{gorenflo2002time}, which are used later to define the random walk transition probability:
{\small \begin{align}
c_k(\beta)=(-1)^{k+1}\binom{\beta}{k}=\left|\binom{\beta}{k}\right|, \quad b_n(\beta)=\sum_{k=0}^n(-1)^k\binom{\beta}{k}, \label{eq.ckbm}
\end{align}}%
where the generalized binomial coefficient { $\binom{\beta}{k}=\frac{\Gamma(\beta+1)}{\Gamma(k+1) \Gamma(\beta-k+1)}$} and the gamma function $\Gamma(\cdot)$ are employed in the definition of the coefficients. The sequences $c_k$ and $b_n$ consist of positive numbers, not greater than 1, decreasing strictly monotonically to zero (see supplementary material for details) and satisfy $\sum_{k=1}^{n}c_k+b_n=1$.
Using these coefficients, we define the transition probabilities of the random walk starting from $\bx_{j_0}$ as
{\small \begin{align}
&\P(\bR(t_{n+1})=\bx_{j_{n+1}} \given \bR(t_0)=\bx_{j_0},\bR(t_1)=\bx_{j_{1}},\ldots,\bR(t_n)=\bx_{j_{n}} ) \nn
& =
\begin{cases}
    c_1-\sigma^\beta & \text { if staying at current location with } j_{n+1}= j_{n}, \\
   \sigma^\beta\frac{ W_{ j_{n} j_{n+1}}}{d_{j_n}}& \text { if jumping to neighboring nodes with } j_{n+1}\ne j_{n},\\
    c_{n+1-k} & \text { if revisiting historical positions with } j_{n+1} = j_{k}, 1\le k\le n-1, \\
    b_n & \text { if revisiting historical positions with } j_{n+1} =j_0.
\end{cases} \label{eq.random_pro}
\end{align}}%
This formulation integrates memory effects, considering the walker's time, position, and path history. The transition mechanism of the memory-inclusive random walk between $t_n$ and $t_{n+1}$ is elucidated as follows:
Suppose the walker is at node $j_{n}$ at time $t_n$, having a full path history $(j_{0},j_{1},\ldots,j_{n})$. We generate a  random number $\rho\in[0,1)$ uniformly, and divide the interval $[0, 1)$ into adjacent sub-intervals with lengths $c_1, c_2, \ldots, c_n, b_n$. We further subdivide the first interval (with length $c_1$) into sub-intervals of lengths $c_1 - \sigma^\beta$ and $\sigma^\beta$.
\vspace{-0.2cm}
\begin{enumerate}
    \item If $\rho$ is in the first interval with length $c_1$, the walker either moves to a neighbor $j_{n+1}=k$ with probability $\sigma^\beta\frac{ W_{j_{n} k}}{d_{j_n}}$ or remains at the current position with probability $c_1-\sigma^\beta$.
\item For $\rho$ in subsequent intervals, the walker jumps to a previously visited node in the history $(j_{0},j_{1},\ldots,j_{n-1})$, specifically, to $j_{n+1-k}$ if in $c_k$, or to $j_0$ if in $b_n$.
\end{enumerate}
\vspace{-0.2cm}
When $\beta<1$, the random walk can, with positive probability, revisit its history, restricting extensive drift. 
We denote $\P(\bR(t_{n}))$ as the probability column vector, with its $j$-th element given as $\P(\bR(t_{n}) = \bx_{j})$.
Additionally, we specify $_i\P(\bR(t_{n}))$ to indicate the situation where the random walker initiates from the $i$-th node, i.e., $\bR(0)=\bx_{i}$, with probability 1. In this case, the initial probability vector $_i\P(\bR(0))$ is represented as a one-hot vector with the $i$-th entry marked as 1. 
Using the technique from \citep{gorenflo2002time}, we can prove the following:
\begin{Theorem}\label{thm.rand_walk_int}
Consider the random walk defined in \cref{eq.random_pro},  with the step size $\sigma$ and number of steps $n$. Under the conditions that $n \to \infty$ and $n\sigma = t$, the limiting probability distribution $\bP(t)\coloneqq\lim_{n \to \infty}\P(\bR(t_{n}))$ satisfies \cref{eq.frac_gra_dif_L}. In other words, 
\begin{align}
    D_t^\beta \bP(t)=-\mathbf{L} \bP(t) \label{eq.zdfva}
\end{align}
\end{Theorem}
Considering that initial conditions and dimensions affect the solutions of FDEs, $\bP(t)$ and $\mathbf{X}(t)$ are not equivalent. However, due to the linearity of FDEs, the following conclusion is straightforward:
\begin{Corollary}\label{cor.rand_walk_int}
    Under the conditions that $n \to \infty$ and $n\sigma = t$,  we have $\lim_{n \to \infty}\sum_i {}_i\mathbb{P}(\bR(t_{n}))\bx_i = \bX(t)$, i.e., $\sum_i {}_i\bP(t)\bx_i = \bX(t)$ with ${}_i \bP(t)\coloneqq\lim_{n \to \infty}{}_i\P(\bR(t_{n}))$ , where $\bX(t)$ is the solution to \cref{eq.frac_gra_dif_L} with the initial condition $\bX(0)= \bX$. 
\end{Corollary}
\begin{Remark}
\cref{thm.rand_walk_int,cor.rand_walk_int} relate F-GAND-l \cref{eq.frac_gra_dif_L} to the non-Markovian random walk in \cref{eq.random_pro}, illustrating memory dependence in FROND.  As $\beta\rightarrow 1$, this process reverts to the Markovian random walk found in GRAND-l \citep{thorpe2022grand++} in \cref{eq.conv_random_pro}. This underscores the FROND framework's capability to apprehend more complex dynamics than integer-order continuous GNNs.
\begin{align}
&\P(\bR(t_{n+1})=\bx_{j_{n+1}} \given \bR(t_0)=\bx_{j_0},\bR(t_1)=\bx_{j_{1}},\ldots,\bR(t_n)=\bx_{j_{n}}) \vspace{-0.2cm}\label{eq.conv_random_pro}\\
&= \P (\bR(t_{n+1})=\bx_{j_{n+1}} \given \bR(t_n)=\bx_{j_{n}} ) = 
\begin{cases}
 1 -\sigma & \text{\scriptsize if staying at current location with } j_{n+1}= j_{n} \\
 \sigma\frac{ W_{ j_{n} j_{n+1}}}{d_{j_n}} & \text{\scriptsize if jumping to neighbors with } j_{n+1}\ne j_{n}
\end{cases}\nonumber
\end{align}%
since we have that all these coefficients vanishing except $c_1=1$, i.e.,
\begin{align}
c_1 =1,\quad \lim_{\beta\rightarrow 1} c_k(\beta)=0, \quad k\ge 2,\quad \lim_{\beta\rightarrow 1} b_n(\beta)=0, \quad n\ge 1. \label{eq.ckbm_limit}
\end{align}
\end{Remark}

\subsubsection{Oversmoothing Mitigation of F-GRAND-l Compared to GRAND-l}\label{ssec.over-smooth}
The seminal research \citep{oonograph}[Corollary 3. and Remark 1] has highlighted that, when considering a GNN as a layered dynamical system, oversmoothing is a broad expression of the \emph{exponential convergence} to stationary states that only retain information about graph connected components and node degrees. 
Under certain conditions, the stationary distribution for the Markovian random walk \cref{eq.conv_random_pro} is given by {\small $\bpi=( \frac{d_1}{\sum_{j=1}^{N}d_j}, \ldots, \frac{d_N}{\sum_{j=1}^{N}d_j})$} \citep{thorpe2022grand++}, with an \emph{exponentially rapid convergence rate} $\left\| \P(\bR(t_{n}))- \bpi\T\right\|_{2} \sim O(e^{-r'n})$ \footnote{We use the asymptotic order notations from \citep{bigo} in this paper.}, where $r'>0$ relates to the eigenvalues of the matrix $\bL$ \citep{chung1997spectral}, and $\|\cdot\|_2$  denotes the $\ell^2$ norm.
This behavior extends to the continuous limit, akin to a first-order linear ODE solution, exhibiting \emph{exponential convergence} with some $r>0$:
\begin{align}
\left\| \bP(t)- \bpi\T\right\|_{2} \sim O(e^{-rt}).  
\end{align}
In contrast, we next prove that the non-Markovian random walk \cref{eq.random_pro} converges to the stationary distribution at a \emph{slow algebraic rate}, thereby helping to mitigate oversmoothing. 
As $\beta\rightarrow 0$, the convergence is expected to be \emph{arbitrarily slow}. In real-world scenarios where we operate within a finite horizon, this slower rate of convergence may be sufficient to alleviate oversmoothing, particularly when it is imperative for a deep model to extract distinctive features instead of achieving exponentially fast convergence to a stationary equilibrium. 
\begin{Theorem}\label{thm.rate}
Under the assumption that the graph is strongly connected and aperiodic, the stationary probability for the non-Markovian random walk \cref{eq.random_pro}, with $0<\beta<1$, is still $\bpi$, which is unique. This mirrors the stationary probability of the Markovian random walk as defined by \cref{eq.conv_random_pro} when $\beta=1$. Notably, when $\beta<1$, the convergence of the distribution (distinct from $\boldsymbol{\pi}$) to $\boldsymbol{\pi}$  is algebraic:
  \begin{align}
      \left\|\bP(t) - \bpi\T \right\|_2  \sim  \Theta( t^{-\beta}). \label{eq.algebraic_conv}
    \end{align}
\end{Theorem}
\begin{Remark} \cref{cor.rand_walk_int,thm.rate} indicate that $\bX(t)=\sum_i {}_i\bP(t)\bx_i$, as the solution to F-GRAND-l \cref{eq.frac_gra_dif_L}, converges to $\sum_i \bpi\T\bx_i = \bpi\T \sum_i \bx_i$ at a slow algebraic rate since $\left\|_i\bP(t) - \bpi\T \right\|_2  \sim  \Theta( t^{-\beta})$ for all $i$. Notably, $\bpi\T \sum_i \bx_i$ forms a rank 1 invariant subspace under the dynamics of \cref{eq.frac_gra_dif_L}, due to $\bpi$ being stationary. This underscores the difference in convergence rates, contrasting the slow algebraic rate in our case with the fast exponential rate \citep{oonograph,ZhaKanSon:C23b}.
\end{Remark}

\subsection{Solving FROND} \label{sec.solvers}
\vspace{-0.2cm}

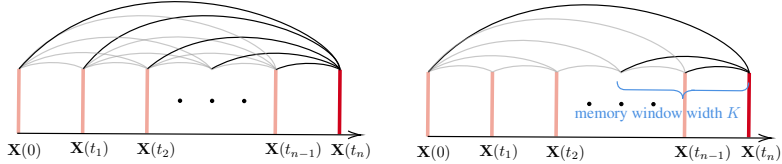
\begin{figure}[H]
\vspace{-0.6cm}
    \centering
    \adjustbox{scale=0.6,center}{

\tikzset{every picture/.style={line width=0.75pt}} 

\begin{tikzpicture}[x=0.75pt,y=0.75pt,yscale=-1,xscale=1]

\draw [color={rgb, 255:red, 241; green, 168; blue, 156 }  ,draw opacity=1 ][line width=2.25]    (83.13,100.77) -- (82.64,154.97) ;
\draw [color={rgb, 255:red, 241; green, 168; blue, 156 }  ,draw opacity=1 ][line width=2.25]    (137.09,100.77) -- (136.6,154.97) ;
\draw [color={rgb, 255:red, 241; green, 168; blue, 156 }  ,draw opacity=1 ][line width=2.25]    (245.03,101.47) -- (244.54,155.67) ;
\draw [color={rgb, 255:red, 208; green, 2; blue, 27 }  ,draw opacity=1 ][line width=2.25]    (299,101.47) -- (298.51,155.67) ;
\draw [color={rgb, 255:red, 155; green, 155; blue, 155 }  ,draw opacity=0.56 ][line width=0.75]    (83.13,100.77) .. controls (98.83,93.15) and (126.71,94.31) .. (137.09,100.77) ;
\draw [color={rgb, 255:red, 155; green, 155; blue, 155 }  ,draw opacity=0.56 ][line width=0.75]    (83.13,100.77) .. controls (128.51,75.05) and (184.28,93.73) .. (191.06,101.01) ;
\draw [color={rgb, 255:red, 155; green, 155; blue, 155 }  ,draw opacity=0.56 ][line width=0.75]    (137.09,100.77) .. controls (152.79,93.15) and (180.68,94.31) .. (191.06,100.77) ;
\draw [color={rgb, 255:red, 155; green, 155; blue, 155 }  ,draw opacity=0.56 ][line width=0.75]    (191.06,101.47) .. controls (206.76,93.85) and (234.65,95.02) .. (245.03,101.47) ;
\draw [color={rgb, 255:red, 0; green, 0; blue, 0 }  ,draw opacity=1 ][line width=0.75]    (245.03,101.47) .. controls (260.73,93.85) and (288.62,95.02) .. (299,101.47) ;
\draw [color={rgb, 255:red, 155; green, 155; blue, 155 }  ,draw opacity=0.56 ][line width=0.75]    (137.09,100.77) .. controls (182.48,75.05) and (238.24,93.73) .. (245.03,101.01) ;
\draw [color={rgb, 255:red, 0; green, 0; blue, 0 }  ,draw opacity=1 ][line width=0.75]    (191.06,100.77) .. controls (236.45,75.05) and (292.21,93.73) .. (299,101.01) ;
\draw [color={rgb, 255:red, 155; green, 155; blue, 155 }  ,draw opacity=0.56 ][line width=0.75]    (83.13,100.54) .. controls (140.2,51.12) and (234.65,82.06) .. (245.03,101.47) ;
\draw [color={rgb, 255:red, 0; green, 0; blue, 0 }  ,draw opacity=1 ][line width=0.75]    (137.09,100.77) .. controls (194.17,51.35) and (288.62,82.29) .. (299,101.71) ;
\draw [color={rgb, 255:red, 0; green, 0; blue, 0 }  ,draw opacity=1 ][line width=0.75]    (83.13,100.77) .. controls (146.5,17.84) and (296.71,69.8) .. (299,101.47) ;
\draw    (218.45,127.82) ;
\draw [shift={(218.45,127.82)}, rotate = 0] [color={rgb, 255:red, 0; green, 0; blue, 0 }  ][fill={rgb, 255:red, 0; green, 0; blue, 0 }  ][line width=0.75]      (0, 0) circle [x radius= 1.34, y radius= 1.34]   ;
\draw    (191.47,127.71) ;
\draw [shift={(191.47,127.71)}, rotate = 0] [color={rgb, 255:red, 0; green, 0; blue, 0 }  ][fill={rgb, 255:red, 0; green, 0; blue, 0 }  ][line width=0.75]      (0, 0) circle [x radius= 1.34, y radius= 1.34]   ;
\draw    (164.5,127.59) ;
\draw [shift={(164.5,127.59)}, rotate = 0] [color={rgb, 255:red, 0; green, 0; blue, 0 }  ][fill={rgb, 255:red, 0; green, 0; blue, 0 }  ][line width=0.75]      (0, 0) circle [x radius= 1.34, y radius= 1.34]   ;
\draw [color={rgb, 255:red, 241; green, 168; blue, 156 }  ,draw opacity=1 ][line width=2.25]    (29.16,100.77) -- (28.67,154.97) ;
\draw [color={rgb, 255:red, 155; green, 155; blue, 155 }  ,draw opacity=0.56 ][line width=0.75]    (29.16,100.77) .. controls (44.86,93.15) and (72.74,94.31) .. (83.13,100.77) ;
\draw [color={rgb, 255:red, 155; green, 155; blue, 155 }  ,draw opacity=0.56 ][line width=0.75]    (29.16,100.19) .. controls (74.54,74.47) and (130.31,93.15) .. (137.09,100.42) ;
\draw [color={rgb, 255:red, 155; green, 155; blue, 155 }  ,draw opacity=0.56 ][line width=0.75]    (29.16,100.77) .. controls (86.23,51.35) and (180.68,82.29) .. (191.06,101.71) ;
\draw [color={rgb, 255:red, 155; green, 155; blue, 155 }  ,draw opacity=0.55 ][line width=0.75]    (29.16,100.77) .. controls (92.53,17.84) and (242.74,69.8) .. (245.03,101.47) ;
\draw [color={rgb, 255:red, 0; green, 0; blue, 0 }  ,draw opacity=1 ][line width=0.75]    (29.16,100.77) .. controls (122.21,-5.62) and (294.91,63.84) .. (299,101.94) ;
\draw [color={rgb, 255:red, 241; green, 168; blue, 156 }  ,draw opacity=1 ][line width=2.25]    (427.13,103.44) -- (426.64,157.63) ;
\draw [color={rgb, 255:red, 241; green, 168; blue, 156 }  ,draw opacity=1 ][line width=2.25]    (481.09,103.44) -- (480.6,157.63) ;
\draw [color={rgb, 255:red, 241; green, 168; blue, 156 }  ,draw opacity=1 ][line width=2.25]    (589.03,104.14) -- (588.54,158.33) ;
\draw [color={rgb, 255:red, 208; green, 2; blue, 27 }  ,draw opacity=1 ][line width=2.25]    (643,104.14) -- (642.51,158.33) ;
\draw [color={rgb, 255:red, 155; green, 155; blue, 155 }  ,draw opacity=0.56 ][line width=0.75]    (427.13,103.44) .. controls (442.83,95.81) and (470.71,96.98) .. (481.09,103.44) ;
\draw [color={rgb, 255:red, 155; green, 155; blue, 155 }  ,draw opacity=0.56 ][line width=0.75]    (481.09,103.44) .. controls (496.79,95.81) and (524.68,96.98) .. (535.06,103.44) ;
\draw [color={rgb, 255:red, 155; green, 155; blue, 155 }  ,draw opacity=0.56 ][line width=0.75]    (535.06,104.14) .. controls (550.76,96.51) and (578.65,97.68) .. (589.03,104.14) ;
\draw [color={rgb, 255:red, 0; green, 0; blue, 0 }  ,draw opacity=1 ][line width=0.75]    (589.03,104.14) .. controls (604.73,96.51) and (632.62,97.68) .. (643,104.14) ;
\draw [color={rgb, 255:red, 0; green, 0; blue, 0 }  ,draw opacity=1 ][line width=0.75]    (535.06,103.44) .. controls (580.45,77.72) and (636.21,96.4) .. (643,103.67) ;
\draw    (562.45,130.49) ;
\draw [shift={(562.45,130.49)}, rotate = 0] [color={rgb, 255:red, 0; green, 0; blue, 0 }  ][fill={rgb, 255:red, 0; green, 0; blue, 0 }  ][line width=0.75]      (0, 0) circle [x radius= 1.34, y radius= 1.34]   ;
\draw    (535.47,130.37) ;
\draw [shift={(535.47,130.37)}, rotate = 0] [color={rgb, 255:red, 0; green, 0; blue, 0 }  ][fill={rgb, 255:red, 0; green, 0; blue, 0 }  ][line width=0.75]      (0, 0) circle [x radius= 1.34, y radius= 1.34]   ;
\draw    (508.5,130.26) ;
\draw [shift={(508.5,130.26)}, rotate = 0] [color={rgb, 255:red, 0; green, 0; blue, 0 }  ][fill={rgb, 255:red, 0; green, 0; blue, 0 }  ][line width=0.75]      (0, 0) circle [x radius= 1.34, y radius= 1.34]   ;
\draw [color={rgb, 255:red, 241; green, 168; blue, 156 }  ,draw opacity=1 ][line width=2.25]    (373.16,103.44) -- (372.67,157.63) ;
\draw [color={rgb, 255:red, 155; green, 155; blue, 155 }  ,draw opacity=0.56 ][line width=0.75]    (373.16,103.44) .. controls (388.86,95.81) and (416.74,96.98) .. (427.13,103.44) ;
\draw [color={rgb, 255:red, 155; green, 155; blue, 155 }  ,draw opacity=0.56 ][line width=0.75]    (373.16,102.86) .. controls (418.54,77.13) and (474.31,95.81) .. (481.09,103.09) ;
\draw [color={rgb, 255:red, 155; green, 155; blue, 155 }  ,draw opacity=0.56 ][line width=0.75]    (373.16,103.44) .. controls (430.23,54.02) and (524.68,84.96) .. (535.06,104.37) ;
\draw  [color={rgb, 255:red, 74; green, 144; blue, 226 }  ,draw opacity=1 ] (532,111.25) .. controls (531.96,115.92) and (534.27,118.27) .. (538.94,118.31) -- (577.13,118.66) .. controls (583.8,118.72) and (587.11,121.08) .. (587.06,125.75) .. controls (587.11,121.08) and (590.46,118.78) .. (597.13,118.84)(594.13,118.81) -- (635.44,119.19) .. controls (640.11,119.23) and (642.46,116.92) .. (642.5,112.25) ;
\draw [color={rgb, 255:red, 0; green, 0; blue, 0 }  ,draw opacity=1 ]   (27.67,154.97) -- (314.67,156.32) ;
\draw [shift={(316.67,156.33)}, rotate = 180.27] [color={rgb, 255:red, 0; green, 0; blue, 0 }  ,draw opacity=1 ][line width=0.75]    (10.93,-3.29) .. controls (6.95,-1.4) and (3.31,-0.3) .. (0,0) .. controls (3.31,0.3) and (6.95,1.4) .. (10.93,3.29)   ;
\draw [color={rgb, 255:red, 0; green, 0; blue, 0 }  ,draw opacity=1 ]   (372.67,157.63) -- (659.67,158.99) ;
\draw [shift={(661.67,159)}, rotate = 180.27] [color={rgb, 255:red, 0; green, 0; blue, 0 }  ,draw opacity=1 ][line width=0.75]    (10.93,-3.29) .. controls (6.95,-1.4) and (3.31,-0.3) .. (0,0) .. controls (3.31,0.3) and (6.95,1.4) .. (10.93,3.29)   ;
\draw [color={rgb, 255:red, 0; green, 0; blue, 0 }  ,draw opacity=1 ][line width=0.75]    (373.16,103.44) .. controls (466.21,-2.96) and (638.91,66.51) .. (643,104.61) ;
\draw [color={rgb, 255:red, 155; green, 155; blue, 155 }  ,draw opacity=0.55 ][line width=0.75]    (373.16,103.44) .. controls (436.53,20.51) and (586.74,72.46) .. (589.03,104.14) ;

\draw (18,161.4) node [anchor=north west][inner sep=0.75pt]  [font=\normalsize]  {$\mathbf{X}( 0)$};
\draw (70.67,160.4) node [anchor=north west][inner sep=0.75pt]  [font=\normalsize]  {$\mathbf{X}( t_{1})$};
\draw (126.67,161.07) node [anchor=north west][inner sep=0.75pt]  [font=\normalsize]  {$\mathbf{X}( t_{2})$};
\draw (234.67,161.07) node [anchor=north west][inner sep=0.75pt]  [font=\normalsize]  {$\mathbf{X}( t_{n-1})$};
\draw (289.67,161.73) node [anchor=north west][inner sep=0.75pt]  [font=\normalsize]  {$\mathbf{X}( t_{n})$};
\draw (362,163.4) node [anchor=north west][inner sep=0.75pt]  [font=\normalsize]  {$\mathbf{X}( 0)$};
\draw (414.67,162.4) node [anchor=north west][inner sep=0.75pt]  [font=\normalsize]  {$\mathbf{X}( t_{1})$};
\draw (470.67,163.07) node [anchor=north west][inner sep=0.75pt]  [font=\normalsize]  {$\mathbf{X}( t_{2})$};
\draw (578.67,163.07) node [anchor=north west][inner sep=0.75pt]  [font=\normalsize]  {$\mathbf{X}( t_{n-1})$};
\draw (634.67,163.73) node [anchor=north west][inner sep=0.75pt]  [font=\normalsize]  {$\mathbf{X}( t_{n})$};
\draw (495,131.67) node [anchor=north west][inner sep=0.75pt]  [color={rgb, 255:red, 74; green, 144; blue, 226 }  ,opacity=1 ] [align=left] {memory window width $\displaystyle K$ };

\end{tikzpicture}

    }\vspace{-0.2cm}
    \caption{\small Diagrams of fractional Adams–Bashforth–Moulton method with full (left) and short (right) memory. 
    }\label{fig.block1}
\end{figure}
\vspace{-0.4cm}
The studies by \citep{chen2018neural,quaglino2019snode,yan2019robustness} introduce numerical solvers specifically designed for integer-order neural ODE models.
Our research, in contrast, engages with fractional-order ODEs, entities inherently more intricate than integer-order ODEs.
To address the scenario \emph{where $\beta$ is non-integer}, we introduce the \emph{fractional explicit Adams–Bashforth–Moulton solver}, incorporating three variants employed in this study: the \textbf{basic predictor} discussed in this section, the \textbf{predictor-corrector} elaborated in \cref{subsec:supp_corrector}, and the \textbf{short memory principle} detailed in \cref{subsec:supp_Short_memory}. Additionally, we present one \textbf{implicit L1} solver in \cref{subs:l1solver}. These methods exemplify how time still acts as a continuous analog to the layer index and elucidate how memory dependence manifests as nontrivial dense or skip connections between layers (see \cref{fig.block1,fig.block_inf}), stemming from the non-local properties of fractional derivatives.

\tb{Basic Predictor:} We first employ a preliminary numerical solver  called ``predictor'' \citep{diethelm2004detailed}  through time discretisation.  Let $h$ be a small positive discretization parameter. We have
\begin{align}
{}_{\mathrm{P}}\bX^{(k)}= \sum_{j=0}^{\lceil\beta\rceil-1} \frac{t_{k}^j}{j!} \bX^{[j]}(0)+\frac{1}{\Gamma(\beta)} \sum_{j=0}^{k-1} \mu_{j, k} \calF(\bW,\bX^{(j)}),\label{eq.pre}
\end{align}
where $\mu_{j, k}= \frac{h^\beta}{\beta}\left((k-j)^\beta-(k-1-j)^\beta\right)$, $k$ denotes the discrete time index (iteration), and $t_k=kh$ represents the discretized time steps. $\bX^{(k)}$ is the numerical approximation of  $\bX(t_k)$.
When $\beta=1$, this method simplifies to the Euler solver in \citep{chen2018neural,chamberlain2021grand} as $\mu_{j, n}\equiv h$, yielding ${}_{\mathrm{P}}\bX^{(k)} = \bX^{(k-1)}  + h\calF(\bW,\bX^{(k-1)})$. 
Thus, our basic predictor can be considered as the fractional Euler method or fractional Adams–Bashforth method, which is a generalization of the Euler method used in \citep{chen2018neural,chamberlain2021grand}. However, when $\beta<1$, we need to utilize the full memory $\{\calF(\bW,\bX^{(j))}\}_{j=0}^{k-1}$. 
The block diagram in \cref{fig.block1} shows the basic predictor and the short memory variant, highlighting the inclusion of nontrivial dense or skip connections in our framework. 
A more refined visualization is conveyed in \cref{fig.block_inf}, elucidating the manner in which information propagates through layers and the graph's spatial domain.

\section{Experiments} \label{sec.exp}
We execute a series of experiments to illustrate that continuous GNNs formulated within the FROND framework using $D_t^\beta$ outperform their traditional counterparts based on integer-order derivatives.
Importantly, our primary aim is not to achieve state-of-the-art results, but rather to demonstrate the additional effectiveness of the FROND framework when applied to existing integer-order continuous GNNs.
In the main paper, we detail the impressive results achieved by F-GRAND, particularly emphasizing its efficacy on tree-structured data, and F-CDE, highlighting its proficiency in managing large heterophilic datasets. We also validate the slow algebraic convergence, as discussed in \cref{thm.rate}, by constructing deeper GNNs with non-integer $\beta<1$. To maintain consistency in the experiments presented in the main paper, the basic predictor solver is used instead of other solvers when $\beta<1$.

\tb{More Experiments In the Appendix:} The \cref{sec:supp_more_exp} section provides additional details covering various aspects such as experimental settings, described in \cref{subsec:supp_dataset,subsec.graph_class_set,subsec:supp_imp_det}, the computational complexity of F-GRAND in \cref{subsec:supp_computetime}, and analysis of F-GRAND’s robustness against adversarial attacks in \cref{subsec:supp_robust_exp}.
Furthermore, results related to other FROND-based continuous GNNs are extensively presented in the \cref{sec.app_moredynamic}. In the main paper, we utilize the basic predictor, as delineated in \cref{eq.pre}, while the exploration of its variants is reserved for the \cref{subsec:supp_solver_exp}. Additional insights into the optimal fractional-derivative order $\beta$ and fractality in graph datasets are explored in Section \cref{sec.app_fractal}.

\subsection{Node Classification of F-GRAND}
\tb{Datasets and splitting.} 
We utilize datasets with varied topologies, including citation networks (Cora \citep{McCallum2004AutomatingTC}, Citeseer \citep{SenNamata2008}, Pubmed \citep{namata:mlg12-wkshp}), tree-structured datasets (Disease and Airport \citep{chami2019hyperbolic_GCNN}), coauthor and co-purchasing graphs (CoauthorCS~\citep{shchur2018pitfalls}, Computer and Photo~\citep{McAuleySIGIR2015}), and the ogbn-arxiv dataset \citep{hu2020open}. We follow the same data splitting and pre-processing in \citep{chami2019hyperbolic_GCNN} for Disease and Airport datasets. Consistent with experiment settings in GRAND \citep{chamberlain2021grand}, we use random splits for the largest connected component of each other dataset.  We also incorporate the large-scale Ogbn-Products dataset \citep{hu2021ogbdataset} to demonstrate the scalability of the FROND framework, with the results displayed in \cref{tab:ogbn-products}.\\
\tb{Methods.} 
For a comprehensive performance comparison, we select several prominent GNN models as baselines, including GCN \citep{kipf2017semi}, and GAT \citep{velickovic2018graph}. Given the inclusion of tree-structured datasets, we also incorporate well-suited baselines: HGCN\citep{chami2019hyperbolic_GCNN} and GIL \citep{zhu2020GIL}. To highlight the benefits of memorized dynamics in FROND, we include GRAND \citep{chamberlain2021grand} as a special case of F-GRAND with $\beta=1$.
In line with \citep{chamberlain2021grand}, we examine two F-GRAND variants: F-GRAND-nl \cref{eq.frac_grand-nl} and F-GRAND-l \cref{eq.frac_gra_dif_L}.
Graph rewiring is not explored in this study. 
Where available, results from the paper \citep{chamberlain2021grand} are used. \\

\vspace{-0.7cm}
\begin{table*}[!htb]\small
\centering
\caption{Node classification results(\%) for random train-val-test splits. The best and the second-best results are highlighted in \first{} and \second{}, respectively.}
\label{tab:noderesults}
 \resizebox{1\textwidth}{!}{
 \begin{tabular}{c|cccccccc|cc}
\toprule
Method   & Cora      & Citeseer           &  Pubmed      & CoauthorCS             & Computer            & Photo          & CoauthorPhy &  ogbn-arxiv & Airport & Disease   \\

\midrule
GCN    &  81.5$\pm$1.3        &  71.9$\pm$1.9   &  77.8$\pm$2.9   &  91.1$\pm$0.5     & 82.6$\pm$2.4  &  91.2$\pm$1.2   &  92.8$\pm$1.0   &  72.2$\pm$0.3 &  81.6$\pm$0.6  &  69.8$\pm$0.5   \\

GAT    &  81.8$\pm$1.3       &  71.4$\pm$1.9   &  78.7$\pm$2.3   &  90.5$\pm$0.6  &  78.0$\pm$19.0  & 85.7$\pm$20.3   &  92.5$\pm$0.90  &  \first{73.7$\pm$0.1}  &  81.6$\pm$0.4  &  70.4$\pm$0.5   \\

HGCN &  78.7$\pm$1.0 &  65.8$\pm$2.0 &  76.4$\pm$0.8  & 90.6$\pm$0.3   & 80.6$\pm$1.8   & 88.2$\pm$1.4   & 90.8$\pm$1.5   &  59.6$\pm$0.4 & 85.4$\pm$0.7  &  89.9$\pm$1.1\\
GIL  &   82.1$\pm$1.1  &  71.1$\pm$1.2  &   77.8$\pm$0.6 &   89.4$\pm$1.5    & -- &   89.6$\pm$1.3 & -- & -- &    91.5$\pm$1.7 & \second{90.8$\pm$0.5} \\
\midrule
GRAND-l    &  \second{83.6$\pm$1.0}        &  73.4$\pm$0.5   &  78.8$\pm$1.7   & 92.9$\pm$0.4     & 83.7$\pm$1.2  & 92.3$\pm$0.9   &  93.5$\pm$0.9  &  71.9$\pm$0.2  &  80.5$\pm$9.6  &  74.5$\pm$3.4   \\

GRAND-nl    &  82.3$\pm$1.6        &  70.9$\pm$1.0   &  77.5$\pm$1.8   & 92.4$\pm$0.3     &  82.4$\pm$2.1  & 92.4$\pm$0.8   &  91.4$\pm$1.3  &  71.2$\pm$0.2  &  90.9$\pm$1.6  &  81.0$\pm$6.7   \\

\midrule
F-GRAND-l   &  \first{84.8$\pm$1.1}        &  \second{74.0$\pm$1.5}   &  \first{79.4$\pm$1.5}   & \first{93.0$\pm$0.3}    & \first{84.4$\pm$1.5}  & \second{92.8$\pm$0.6}   &  \first{94.5$\pm$0.4}  & \second{72.6$\pm$0.1 }    &  \first{98.1$\pm$0.2} &  \first{92.4$\pm$3.9}  \\
\gray{$\beta$ for F-GRAND-l} &\gray{0.9}  & \gray{0.9} & \gray{0.9}& \gray{0.7}& \gray{0.98} & \gray{0.9}&  \gray{0.6} & \gray{0.7} & \gray{0.5} &\gray{0.6} \\
F-GRAND-nl   &  83.2$\pm$1.1  &  \first{74.7$\pm$1.9}   &  \second{79.2$\pm$0.7} & \second{92.9$\pm$0.4}    & 
 \second{84.1$\pm$0.9}  & \first{93.1$\pm$0.9}   &  \second{93.9$\pm$0.5}   &  71.4$\pm$0.3  &  \second{96.1$\pm$0.7}  &  85.5$\pm$2.5   \\
\gray{$\beta$ for F-GRAND-nl} &  \gray{0.9}& \gray{0.9} &\gray{0.4} & \gray{0.6}& \gray{0.85}&\gray{0.8} &\gray{0.4} & \gray{0.7} & \gray{0.1} &\gray{0.7}  \\
\bottomrule
\end{tabular}}
\end{table*}
   \vspace{-0.5cm}
\begin{figure}[!htb]
\hspace{-8mm}
 \begin{minipage}{0.5\linewidth}
    \includegraphics[width=\textwidth,trim=0 5 0 12mm,clip]{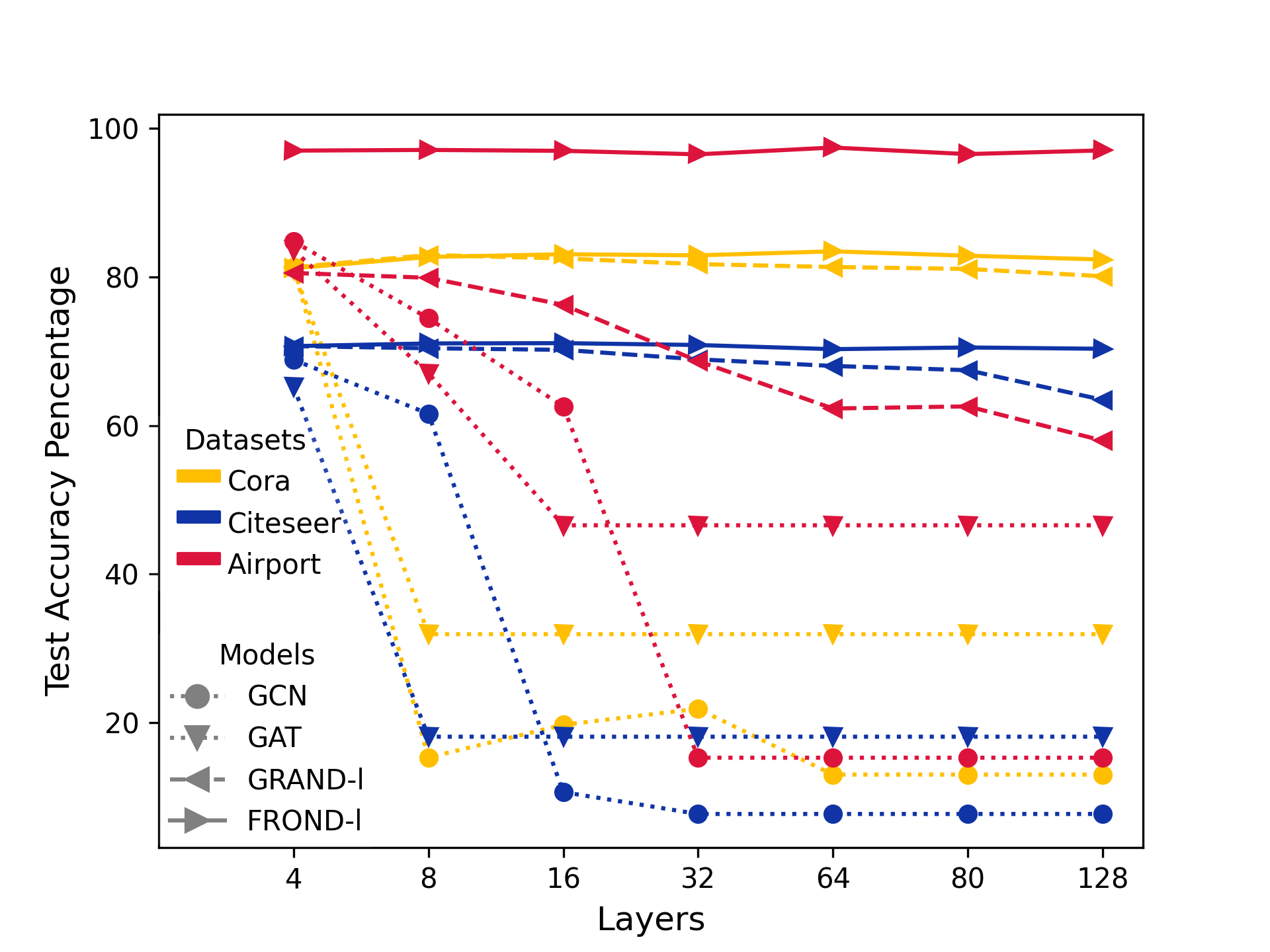}
    \vspace{-8mm}
    \caption{oversmoothing mitigation.}
    \label{fig:over-smooth}
    \vspace{2mm}
\end{minipage}
\hspace{-7mm}
\begin{minipage}{0.5\linewidth}
    \centering
    \vspace{-11mm}
        \captionof{table}{Graph classification results.}
    \label{tab.graphclas}
    \resizebox{1.2\textwidth}{!}{
\begin{tabular}{l|c|c|c|c|c|ccc}
\hline
& \multicolumn{3}{c|}{\textbf{POL}} & \multicolumn{3}{c}{\textbf{GOS}} \\
\textbf{Feature} & \textbf{Profile} & \textbf{word2vec} & \textbf{BERT}  & \textbf{Profile} & \textbf{word2vec} & \textbf{BERT} \\
\midrule
GraphSage & 77.60$\pm$0.68 & 80.36$\pm$0.68 & 81.22$\pm$4.81 &  92.10$\pm$0.08  & 96.58$\pm$0.22  &  \second{97.07$\pm$0.23}  \\
GCN & \second{78.28$\pm$0.52}  & 83.89$\pm$0.53 &  83.44$\pm$0.38 &   89.53$\pm$0.49  & 96.28$\pm$0.08 &  95.96$\pm$0.75 \\
GAT & 74.03$\pm$0.53  &78.69$\pm$0.78   & 82.71$\pm$0.19 &  91.18$\pm$0.23   & 96.57$\pm$0.34 & 96.61$\pm$0.45 \\
\midrule
GRAND-l & 77.83$\pm$0.37 & \second{86.57$\pm$1.13}  & \second{85.97$\pm$0.74}  &  \second{96.11$\pm$0.26} & \second{97.04$\pm$0.55}  & 96.77$\pm$0.34   \\
\midrule
F-GRAND-l & \first{79.49$\pm$0.43}  & \first{88.69$\pm$0.37}  & \first{89.29$\pm$0.93}  &  \first{96.40$\pm$0.19}  & \first{97.40$\pm$0.03} &  \first{97.53$\pm$0.14}   \\

\bottomrule
\end{tabular}}
    \centering
     \captionof{table}{Node classification accuracy of F-GRAND-l under different value of $\beta$ when time $T=8$.}
    \label{tab:nodeclasbeta}
     \resizebox{1.2\textwidth}{!}{\vspace{0.3cm}
    \begin{tabular}{c|c|c|c|c|c|ccccc}
    \toprule
      $\beta$  & 0.1  & 0.3  & 0.5  & 0.7  & 0.9 & 1.0  \\
      \midrule
      Cora & 74.80$\pm$0.42  & 77.0$\pm$0.98    & 79.60$\pm$0.91   & 81.56$\pm$0.30   & 82.68$\pm$0.64 &  82.37$\pm$0.59    \\
      \midrule
      Airport & 97.09$\pm$0.87   & 95.80$\pm$2.03  & 91.66$\pm$6.34  & 84.36$\pm$8.04 & 78.73$\pm$6.33 &  78.88$\pm$9.67 \\
    \bottomrule
    \end{tabular}}
    \vspace{-4mm}
  \end{minipage}
  \vspace{-4mm}
\end{figure}

\tb{Performance.}  The results for graph node classification are summarized in \cref{tab:noderesults}, which also report the optimal $\beta$ obtained via hyperparameter tuning. Consistent with our expectations, F-GRAND surpasses GRAND across nearly all datasets, given that GRAND represents a special case of FROND with $\beta=1$. This underscores the consistent performance enhancement offered by the integration of memorized dynamics. This advantage is particularly noticeable on tree-structured datasets such as Airports and Disease, where F-GRAND markedly outperforms the baselines. For instance, F-GRAND-l outperforms both GRAND and GIL by approximately 7\% on the Airport dataset. Interestingly, our experiments indicate a smaller $\beta$ (signifying greater dynamic memory) is preferable for such fractal-structured datasets, aligning with previous studies on FDEs in biological and chemical systems \citep{nigmatullin1986realization,ionescu2017role}. Further discussion on $\beta$ and its relation to the fractal dimension of graph datasets can be found in \cref{sec.beta,sec.app_fractal}.

 \subsection{Graph Classification of F-GRAND}
We employ the Fake-NewsNet datasets \citep{dou2021upfd}, constructed from Politifact and Gossipcop fact-checking data. More details can be found in the \cref{subsec.graph_class_set}. This dataset features three types of node features: 768-dimensional BERT features, and 300-dimensional spaCy features, both extracted using pre-trained models, and 10-dimensional profile features from Twitter accounts. The graphs in the dataset exhibit a hierarchical tree structure. 
From \cref{tab.graphclas}, we observe that F-GRAND consistently outperforms GRAND with a notable edge on the POL dataset.
\subsection{Oversmoothing of F-GRAND} \label{sec.over-smooth}
To validate that F-GRAND mitigates the oversmoothing issue and performs well with numerous layers, we conducted an experiment using the basic predictor in the \emph{Adams Bashforth Moulton} method as defined in \cref{eq.pre}. This allows us to generate architectures of varying depths. In this context, we utilize the fixed data splitting as described in \citep{chami2019hyperbolic_GCNN}.
As illustrated in \cref{fig:over-smooth}, optimal performance on the Cora dataset is attained with a network depth of 64 layers.
When compared to GRAND-l, F-GRAND-l maintains a consistent performance level across all datasets as the number of layers increases, with virtually no performance drop observed up to 128 layers.
This observation is consistent with our expectations, given that \cref{thm.rate} predicts a slow algebraic convergence. In contrast, GRAND exhibits a faster rate of performance degradation particularly on the Airport~dataset. Further details on oversmoothing mitigation are in \cref{subsec.over-smooth_set}.
\subsection{Ablation Study: Selection of \txp{$\beta$}{beta}} \label{sec.beta}  
In \cref{tab:nodeclasbeta}, we investigate the influence of $\beta$ across various graph datasets. Notably, for the Cora dataset, a larger $\beta$ is optimal, whereas, for tree-structured data, a smaller $\beta$ is preferable. This suggests that the quantity of memorized dynamics should be tailored to the dataset's topology, and a default setting of memoryless graph diffusion with $\beta=1$ may not be optimal. More comprehensive details concerning the variations in $\beta$ can be found in the appendix, specifically in \cref{tab:betat8}. 
\subsection{More integer-order continuous GNNs in FROND framework}
Our FROND framework can be seamlessly applied to various other integer-order continuous GNNs, as elaborated in \cref{sec.app_moredynamic}. Specifically, here we outline the node classification results of FROND based on the CDE model in \cref{tab:res_cde_hetero}. It is evident from the results that F-CDE enhances the performance of the CDE model across almost all large heterophilic datasets. The optimal $\beta$ is determined through hyperparameter tuning. When $\beta=1$, F-CDE seamlessly reverts to CDE, and the results from the original paper are reported.
Additionally, we conduct comprehensive experiments detailed in \cref{sec.app_moredynamic}. The results for F-GRAND++, F-GREAD, and F-GraphCON are available in \cref{tab:res_fgrand++}, \cref{tab:res_f_gread}, and \cref{tab:res_GraphCON}, respectively. Collectively, these results demonstrate that our FROND framework can significantly bolster the performance of integer-order continuous GNNs, without introducing any additional training parameters to the backbones.
\begin{table}[H]
    \caption{Node classification accuracy(\%) of large heterophilic datasets}
    \label{tab:res_cde_hetero}
    \small
    \centering
    \resizebox{0.9\textwidth}{!}{
    \begin{tabular}{cccccccc}
    \toprule
        Model & Roman-empire & Wiki-cooc & Minesweeper & Questions & Workers & Amazon-ratings \\

    \midrule
        CDE &   91.64$\pm$0.28 &  97.99$\pm$0.38  &   95.50$\pm$5.23  &  75.17$\pm$0.99 &  80.70$\pm$1.04 &  47.63$\pm$0.43    \\
    F-CDE  &   \tb{93.06$\pm$0.55} &  \tb{98.73$\pm$0.68}  &   \tb{96.04$\pm$0.25}  & 75.17$\pm$0.99 &  \tb{82.68$\pm$0.86} &  \tb{49.01$\pm$0.56}    \\

    \gray{$\beta$ for F-CDE} &  \gray{0.9}& \gray{0.6 } &\gray{0.6} & \gray{1.0 }& \gray{0.4}&\gray{0.1 } \\

    \bottomrule
    \end{tabular}
    }
\end{table}
\section{Conclusion}
We have introduced FROND, a novel graph learning framework that incorporates Caputo fractional derivatives to capture long-term memory in the graph feature updating dynamics. This approach has demonstrated superior performance compared to various traditional integer-order continuous GNNs. The resulting framework represents a significant advancement in graph representation learning, addressing key challenges in the field, such as oversmoothing. Our results highlight the potential of fractional calculus in enabling more effective graph learning algorithms.

\section*{Acknowledgments and Disclosure of Funding}
This research is supported by the Singapore Ministry of Education Academic Research Fund Tier 2 grant MOE-T2EP20220-0002, and the National Research Foundation, Singapore and Infocomm Media Development Authority under its Future Communications Research and Development Programme. The computational work for this article was partially performed on resources of the National Supercomputing Centre, Singapore (https://www.nscc.sg). 
Xuhao Li is supported by the National Natural Science Foundation of China (Grant No. 12301491) and the Anhui Provincial Natural Science Foundation (Grant No. 2208085QA02). 
To improve the readability, parts of this paper have been grammatically revised using ChatGPT \citep{openai2022chatgpt4}.

\bibliography{IEEEabrv,StringDefinitions,adv_dnn}

\begin{thebibliography}{138}
\providecommand{\natexlab}[1]{#1}
\providecommand{\url}[1]{\texttt{#1}}
\expandafter\ifx\csname urlstyle\endcsname\relax
  \providecommand{\doi}[1]{doi: #1}\else
  \providecommand{\doi}{doi: \begingroup \urlstyle{rm}\Url}\fi

\bibitem[Almeida et~al.(2016)Almeida, Bastos, and Monteiro]{almeida2016modeling}
Ricardo Almeida, Nuno~RO Bastos, and M~Teresa~T Monteiro.
\newblock Modeling some real phenomena by fractional differential equations.
\newblock \emph{Mathematical Methods in the Applied Sciences}, 39\penalty0 (16):\penalty0 4846--4855, 2016.

\bibitem[Alon \& Yahav(2021)Alon and Yahav]{alonbottleneck}
Uri Alon and Eran Yahav.
\newblock On the bottleneck of graph neural networks and its practical implications.
\newblock In \emph{Proc. Int. Conf. Learn. Representations}, 2021.

\bibitem[Antil et~al.(2020)Antil, Khatri, L{\"o}hner, and Verma]{antil2020fractional}
Harbir Antil, Ratna Khatri, Rainald L{\"o}hner, and Deepanshu Verma.
\newblock Fractional deep neural network via constrained optimization.
\newblock \emph{Mach. Learn.: Sci. Technol.}, 2\penalty0 (1):\penalty0 015003, 2020.

\bibitem[Ashoor et~al.(2020)Ashoor, Chen, Rosikiewicz, Wang, Cheng, Wang, Ruan, and Li]{AshoorNC2020}
Haitham Ashoor, Xiaowen Chen, Wojciech Rosikiewicz, Jiahui Wang, Albert Cheng, Ping Wang, Yijun Ruan, and Sheng Li.
\newblock Graph embedding and unsupervised learning predict genomic sub-compartments from hic chromatin interaction data.
\newblock \emph{Nat. Commun.}, 11, 2020.

\bibitem[Atkinson et~al.(2011)Atkinson, Han, and Stewart]{atkinson2011numerical}
Kendall Atkinson, Weimin Han, and David~E Stewart.
\newblock \emph{Numerical solution of ordinary differential equations}.
\newblock John Wiley \& Sons, 2011.

\bibitem[Avelar et~al.(2019)Avelar, Tavares, Gori, and Lamb]{avelar2019discrete}
Pedro~HC Avelar, Anderson~R Tavares, Marco Gori, and Luis~C Lamb.
\newblock Discrete and continuous deep residual learning over graphs.
\newblock \emph{arXiv preprint arXiv:1911.09554}, 2019.

\bibitem[Bagley \& Torvik(1983)Bagley and Torvik]{bagley1983theoretical}
Ronald~L Bagley and PJ~Torvik.
\newblock A theoretical basis for the application of fractional calculus to viscoelasticity.
\newblock \emph{J. Rheology}, 27\penalty0 (3):\penalty0 201--210, 1983.

\bibitem[Bo et~al.(2021)Bo, Wang, Shi, and Shen]{bo2021fagcn}
Deyu Bo, Xiao Wang, Chuan Shi, and Huawei Shen.
\newblock Beyond low-frequency information in graph convolutional networks.
\newblock In \emph{Proceedings of the AAAI Conference on Artificial Intelligence}, volume~35, pp.\  3950--3957, 2021.

\bibitem[Bodnar et~al.(2022)Bodnar, Giovanni, Chamberlain, Li{\`o}, and Bronstein]{crifraben:sheaf2022}
Cristian Bodnar, Francesco~Di Giovanni, Benjamin~Paul Chamberlain, Pietro Li{\`o}, and Michael~M. Bronstein.
\newblock Neural sheaf diffusion: A topological perspective on heterophily and oversmoothing in {GNN}s.
\newblock In \emph{Advances Neural Inf. Process. Syst.}, 2022.

\bibitem[Brockmann et~al.(2006)Brockmann, Hufnagel, and Geisel]{brockmann2006scaling}
Dirk Brockmann, Lars Hufnagel, and Theo Geisel.
\newblock The scaling laws of human travel.
\newblock \emph{Nature}, 439\penalty0 (7075):\penalty0 462--465, 2006.

\bibitem[Chamberlain et~al.(2021{\natexlab{a}})Chamberlain, Rowbottom, Gorinova, Bronstein, Webb, and Rossi]{chamrowgor:grand2021}
Ben Chamberlain, James Rowbottom, Maria~I Gorinova, Michael Bronstein, Stefan Webb, and Emanuele Rossi.
\newblock Grand: Graph neural diffusion.
\newblock In \emph{Proc. Int. Conf. Mach. Learn.}, pp.\  1407--1418, 2021{\natexlab{a}}.

\bibitem[Chamberlain et~al.(2021{\natexlab{b}})Chamberlain, Rowbottom, Eynard, Di~Giovanni, Dong, and Bronstein]{charoweyn:blend2021}
Benjamin Chamberlain, James Rowbottom, Davide Eynard, Francesco Di~Giovanni, Xiaowen Dong, and Michael Bronstein.
\newblock Beltrami flow and neural diffusion on graphs.
\newblock In \emph{Advances Neural Inf. Process. Syst.}, pp.\  1594--1609, 2021{\natexlab{b}}.

\bibitem[Chamberlain et~al.(2021{\natexlab{c}})Chamberlain, Rowbottom, Goronova, Webb, Rossi, and Bronstein]{chamberlain2021grand}
Benjamin~Paul Chamberlain, James Rowbottom, Maria Goronova, Stefan Webb, Emanuele Rossi, and Michael~M Bronstein.
\newblock Grand: Graph neural diffusion.
\newblock In \emph{Proc. Int. Conf. Mach. Learn.}, 2021{\natexlab{c}}.

\bibitem[Chami et~al.(2019)Chami, Ying, R{\'e}, and Leskovec]{chami2019hyperbolic_GCNN}
Ines Chami, Zhitao Ying, Christopher R{\'e}, and Jure Leskovec.
\newblock Hyperbolic graph convolutional neural networks.
\newblock In \emph{Advances Neural Inf. Process. Syst.}, 2019.

\bibitem[Chen et~al.(2018{\natexlab{a}})Chen, Wu, Xu, Chen, Zheng, and Xuan]{Chen2018FastGA}
Jinyin Chen, Yangyang Wu, Xuanheng Xu, Yixian Chen, Haibin Zheng, and Qi~Xuan.
\newblock Fast gradient attack on network embedding.
\newblock \emph{ArXiv}, 2018{\natexlab{a}}.

\bibitem[Chen et~al.(2020)Chen, Wei, Huang, Ding, and Li]{chen2020simple}
Ming Chen, Zhewei Wei, Zengfeng Huang, Bolin Ding, and Yaliang Li.
\newblock Simple and deep graph convolutional networks.
\newblock In \emph{Proc. Int. Conf. Mach. Learn.}, pp.\  1725--1735, 2020.

\bibitem[Chen et~al.(2018{\natexlab{b}})Chen, Rubanova, Bettencourt, and Duvenaud]{chen2018neural}
Ricky~TQ Chen, Yulia Rubanova, Jesse Bettencourt, and David Duvenaud.
\newblock Neural ordinary differential equations.
\newblock In \emph{Advances Neural Inf. Process. Syst.}, 2018{\natexlab{b}}.

\bibitem[Chien et~al.(2020)Chien, Peng, Li, and Milenkovic]{chien2020gprgnn}
Eli Chien, Jianhao Peng, Pan Li, and Olgica Milenkovic.
\newblock Adaptive universal generalized pagerank graph neural network.
\newblock \emph{arXiv preprint arXiv:2006.07988}, 2020.

\bibitem[Choi et~al.(2023)Choi, Hong, Park, and Cho]{choi2023gread}
Jeongwhan Choi, Seoyoung Hong, Noseong Park, and Sung-Bae Cho.
\newblock Gread: Graph neural reaction-diffusion networks.
\newblock In \emph{Proc. Int. Conf. Mach. Learn.}, 2023.

\bibitem[Chung(1997)]{chung1997spectral}
Fan~RK Chung.
\newblock \emph{Spectral graph theory}, volume~92.
\newblock American Mathematical Soc., 1997.

\bibitem[Cohen(2007)]{Cohen2007}
Alan~M. Cohen.
\newblock \emph{Inversion Formulae and Practical Results}, pp.\  23--44.
\newblock Springer US, Boston, MA, 2007.

\bibitem[Coleman \& Noll(1961)Coleman and Noll]{coleman1961foundations}
Bernard~D Coleman and Walter Noll.
\newblock Foundations of linear viscoelasticity.
\newblock \emph{Rev. Modern Phys.}, 33\penalty0 (2):\penalty0 239, 1961.

\bibitem[Deng(2007)]{deng2007short}
Weihua Deng.
\newblock Short memory principle and a predictor--corrector approach for fractional differential equations.
\newblock \emph{J. Comput. Appl. Math.}, 206\penalty0 (1):\penalty0 174--188, 2007.

\bibitem[Di~Giovanni et~al.(2022)Di~Giovanni, Rowbottom, Chamberlain, Markovich, and Bronstein]{di2022graff}
Francesco Di~Giovanni, James Rowbottom, Benjamin~P Chamberlain, Thomas Markovich, and Michael~M Bronstein.
\newblock Graph neural networks as gradient flows.
\newblock \emph{arXiv preprint arXiv:2206.10991}, 2022.

\bibitem[Di~Giovanni et~al.(2023)Di~Giovanni, Rowbottom, Chamberlain, Markovich, and Bronstein]{di2023understanding}
Francesco Di~Giovanni, James Rowbottom, Benjamin~Paul Chamberlain, Thomas Markovich, and Michael~M Bronstein.
\newblock Understanding convolution on graphs via energies.
\newblock \emph{Tran. Mach. Learn. Res.}, 2023.

\bibitem[Diaz-Diaz \& Estrada(2022)Diaz-Diaz and Estrada]{diaz2022time}
Fernando Diaz-Diaz and Ernesto Estrada.
\newblock Time and space generalized diffusion equation on graph/networks.
\newblock \emph{Chaos, Solitons and Fractals}, 156:\penalty0 111791, 2022.

\bibitem[Diethelm(2010)]{diethelm2010analysis}
Kai Diethelm.
\newblock \emph{The analysis of fractional differential equations: an application-oriented exposition using differential operators of Caputo type}, volume 2004.
\newblock Springer, 2010.

\bibitem[Diethelm \& Ford(2002)Diethelm and Ford]{diethelm2002analysis}
Kai Diethelm and Neville~J Ford.
\newblock Analysis of fractional differential equations.
\newblock \emph{J. Math. Anal. Appl.}, 265\penalty0 (2):\penalty0 229--248, 2002.

\bibitem[Diethelm et~al.(2004)Diethelm, Ford, and Freed]{diethelm2004detailed}
Kai Diethelm, Neville~J Ford, and Alan~D Freed.
\newblock Detailed error analysis for a fractional adams method.
\newblock \emph{Numer. Algorithms}, 36:\penalty0 31--52, 2004.

\bibitem[Dou et~al.(2021)Dou, Shu, Xia, Yu, and Sun]{dou2021upfd}
Yingtong Dou, Kai Shu, Congying Xia, Philip~S. Yu, and Lichao Sun.
\newblock User preference-aware fake news detection.
\newblock In \emph{Proc. Int. ACM SIGIR Conf. Res. Develop. Inform. Retrieval}, 2021.

\bibitem[Du et~al.(2017)Du, Zhang, Wu, Moura, and Kar]{Du2017TopologyAG}
Jian Du, Shanghang Zhang, Guanhang Wu, Jos{\'e} M.~F. Moura, and Soummya Kar.
\newblock Topology adaptive graph convolutional networks.
\newblock \emph{ArXiv}, abs/1710.10370, 2017.

\bibitem[Du et~al.(2022)Du, Shi, Fu, Ma, Liu, Han, and Zhang]{du2022gbkgnn}
Lun Du, Xiaozhou Shi, Qiang Fu, Xiaojun Ma, Hengyu Liu, Shi Han, and Dongmei Zhang.
\newblock Gbk-gnn: Gated bi-kernel graph neural networks for modeling both homophily and heterophily.
\newblock In \emph{Proceedings of the ACM Web Conference 2022}, pp.\  1550--1558, 2022.

\bibitem[Dupont et~al.(2019)Dupont, Doucet, and Teh]{dupont2019augmented}
Emilien Dupont, Arnaud Doucet, and Yee~Whye Teh.
\newblock Augmented neural odes.
\newblock In \emph{Advances Neural Inf. Process. Syst.}, pp.\  1--11, 2019.

\bibitem[Feng et~al.(2022)Feng, Chen, Li, Sarkar, and Zhang]{feng2022khopmessage}
Jiarui Feng, Yixin Chen, Fuhai Li, Anindya Sarkar, and Muhan Zhang.
\newblock How powerful are k-hop message passing graph neural networks.
\newblock \emph{Advances in Neural Information Processing Systems}, 35:\penalty0 4776--4790, 2022.

\bibitem[Gao \& Sun(2011)Gao and Sun]{gao2011compact}
Guang-hua Gao and Zhi-zhong Sun.
\newblock A compact finite difference scheme for the fractional sub-diffusion equations.
\newblock \emph{Journal of Computational Physics}, 230\penalty0 (3):\penalty0 586--595, 2011.

\bibitem[Gao \& Ji(2019)Gao and Ji]{gao2019graph}
Hongyang Gao and Shuiwang Ji.
\newblock Graph u-nets.
\newblock In \emph{Proc. Int. Conf. Mach. Learn.}, pp.\  2083--2092, 2019.

\bibitem[G{\'o}mez-Aguilar et~al.(2016)G{\'o}mez-Aguilar, Miranda-Hern{\'a}ndez, L{\'o}pez-L{\'o}pez, Alvarado-Mart{\'\i}nez, and Baleanu]{gomez2016modeling}
Jos{\'e}~Francisco G{\'o}mez-Aguilar, Margarita Miranda-Hern{\'a}ndez, MG~L{\'o}pez-L{\'o}pez, Victor~Manuel Alvarado-Mart{\'\i}nez, and Dumitru Baleanu.
\newblock Modeling and simulation of the fractional space-time diffusion equation.
\newblock \emph{Commun. Nonlinear Sci. Numer. Simul.}, 30\penalty0 (1-3):\penalty0 115--127, 2016.

\bibitem[Gorenflo \& Mainardi(2003)Gorenflo and Mainardi]{gorenflo2003fractional}
Rudolf Gorenflo and Francesco Mainardi.
\newblock Fractional diffusion processes: probability distributions and continuous time random walk.
\newblock In \emph{Process. Long-Range Correlations: Theory Appl.}, pp.\  148--166. Springer, 2003.

\bibitem[Gorenflo et~al.(2002)Gorenflo, Mainardi, Moretti, and Paradisi]{gorenflo2002time}
Rudolf Gorenflo, Francesco Mainardi, Daniele Moretti, and Paolo Paradisi.
\newblock Time fractional diffusion: a discrete random walk approach.
\newblock \emph{Nonlinear Dynamics}, 29:\penalty0 129--143, 2002.

\bibitem[Gravina et~al.(2022)Gravina, Bacciu, and Gallicchio]{gravina2022anti}
Alessio Gravina, Davide Bacciu, and Claudio Gallicchio.
\newblock Anti-symmetric dgn: A stable architecture for deep graph networks.
\newblock In \emph{Proc. Int. Conf. Learn. Representations}, 2022.

\bibitem[Guo et~al.(2022)Guo, Wu, Yu, and Zhou]{guo2022monte}
Ling Guo, Hao Wu, Xiaochen Yu, and Tao Zhou.
\newblock Monte carlo fpinns: Deep learning method for forward and inverse problems involving high dimensional fractional partial differential equations.
\newblock \emph{Comput. Methods Appl. Mechanics Eng.}, 400:\penalty0 115523, 2022.

\bibitem[Gustafson et~al.(2017)Gustafson, Bayati, and Eckhoff]{gustafson2017fractional}
Kyle~B Gustafson, Basil~S Bayati, and Philip~A Eckhoff.
\newblock Fractional diffusion emulates a human mobility network during a simulated disease outbreak.
\newblock \emph{Frontiers Ecology Evol.}, 5:\penalty0 35, 2017.

\bibitem[Gutteridge et~al.(2023)Gutteridge, Dong, Bronstein, and Di~Giovanni]{gutteridge2023drew}
Benjamin Gutteridge, Xiaowen Dong, Michael~M Bronstein, and Francesco Di~Giovanni.
\newblock Drew: Dynamically rewired message passing with delay.
\newblock In \emph{Proc. Int. Conf. Mach. Learn.}, pp.\  12252--12267, 2023.

\bibitem[Haber \& Ruthotto(2017)Haber and Ruthotto]{haber2017stable}
Eldad Haber and Lars Ruthotto.
\newblock Stable architectures for deep neural networks.
\newblock \emph{Inverse Problems}, 34\penalty0 (1):\penalty0 1--23, December 2017.

\bibitem[Hamilton et~al.(2017)Hamilton, Ying, and Leskovec]{hamilton2017inductive}
William~L. Hamilton, Rex Ying, and Jure Leskovec.
\newblock Inductive representation learning on large graphs.
\newblock In \emph{Advances Neural Inf. Process. Syst.}, 2017.

\bibitem[Han et~al.(2023)Han, Shi, Lin, and Gao]{han2023continuous}
Andi Han, Dai Shi, Lequan Lin, and Junbin Gao.
\newblock From continuous dynamics to graph neural networks: Neural diffusion and beyond.
\newblock \emph{arXiv preprint arXiv:2310.10121}, 2023.

\bibitem[Hartman(2002)]{hartman2002ordinary}
Philip Hartman.
\newblock \emph{Ordinary differential equations}.
\newblock SIAM, 2002.

\bibitem[He et~al.(2016)He, Zhang, Ren, and Sun]{HeCVPR2016}
Kaiming He, Xiangyu Zhang, Shaoqing Ren, and Jian Sun.
\newblock Deep residual learning for image recognition.
\newblock In \emph{Proc. Conf. Comput. Vision Pattern Recognition}, 2016.

\bibitem[Horn \& Johnson(2012)Horn and Johnson]{horn2012matrix}
Roger~A Horn and Charles~R Johnson.
\newblock \emph{Matrix analysis}.
\newblock Cambridge university press, New York, 2012.

\bibitem[Hu et~al.(2020)Hu, Fey, Zitnik, Dong, Ren, Liu, Catasta, and Leskovec]{hu2020open}
Weihua Hu, Matthias Fey, Marinka Zitnik, Yuxiao Dong, Hongyu Ren, Bowen Liu, Michele Catasta, and Jure Leskovec.
\newblock Open graph benchmark: Datasets for machine learning on graphs.
\newblock \emph{arXiv:2005.00687}, 2020.

\bibitem[Hu et~al.(2021)Hu, Fey, Zitnik, Dong, Ren, Liu, Catasta, and Leskovec]{hu2021ogbdataset}
Weihua Hu, Matthias Fey, Marinka Zitnik, Yuxiao Dong, Hongyu Ren, Bowen Liu, Michele Catasta, and Jure Leskovec.
\newblock Open graph benchmark: Datasets for machine learning on graphs, 2021.

\bibitem[Huang et~al.(2017)Huang, Liu, Van Der~Maaten, and Weinberger]{huang2017densely}
Gao Huang, Zhuang Liu, Laurens Van Der~Maaten, and Kilian~Q Weinberger.
\newblock Densely connected convolutional networks.
\newblock In \emph{Proc. Conf. Comput. Vision Pattern Recognition}, 2017.

\bibitem[Hussain et~al.(2022)Hussain, Cao, Sikdar, Helic, Lex, Strohmaier, and Kern]{hussain2022adversarial}
Hussain Hussain, Meng Cao, Sandipan Sikdar, Denis Helic, Elisabeth Lex, Markus Strohmaier, and Roman Kern.
\newblock Adversarial inter-group link injection degrades the fairness of graph neural networks.
\newblock \emph{arXiv preprint arXiv:2209.05957}, 2022.

\bibitem[Ionescu et~al.(2017)Ionescu, Lopes, Copot, Machado, and Bates]{ionescu2017role}
C~Ionescu, A~Lopes, Dana Copot, JA~Tenreiro Machado, and Jason~HT Bates.
\newblock The role of fractional calculus in modeling biological phenomena: A review.
\newblock \emph{Commun. Nonlinear Sci. Numer. Simul.}, 51:\penalty0 141--159, 2017.

\bibitem[Javadi et~al.(2023)Javadi, Mesgarani, Nikan, and Avazzadeh]{javadi2023solving}
Rana Javadi, Hamid Mesgarani, Omid Nikan, and Zakieh Avazzadeh.
\newblock Solving fractional order differential equations by using fractional radial basis function neural network.
\newblock \emph{Symmetry}, 15\penalty0 (6):\penalty0 1275, 2023.

\bibitem[Jin et~al.(2017)Jin, Li, and Zhou]{jin2017correction}
Bangti Jin, Buyang Li, and Zhi Zhou.
\newblock Correction of high-order bdf convolution quadrature for fractional evolution equations.
\newblock \emph{SIAM J. Sci. Comput.}, 39\penalty0 (6):\penalty0 A3129--A3152, 2017.

\bibitem[Jin et~al.(2020)Jin, Ma, Liu, Tang, Wang, and Tang]{jin2020prognn}
Wei Jin, Yao Ma, Xiaorui Liu, Xianfeng Tang, Suhang Wang, and Jiliang Tang.
\newblock Graph structure learning for robust graph neural networks.
\newblock In \emph{Proc. Int. Conf. Knowl. Discovery Data Mining}, pp.\  66--74, 2020.

\bibitem[Kang et~al.(2021)Kang, Song, Ding, and Tay]{kang2021Neurips}
Qiyu Kang, Yang Song, Qinxu Ding, and Wee~Peng Tay.
\newblock Stable neural {ODE} with {Lyapunov-stable} equilibrium points for defending against adversarial attacks.
\newblock In \emph{Advances Neural Inf. Process. Syst.}, 2021.

\bibitem[Kang et~al.(2023)Kang, Zhao, Song, Wang, and Tay]{KanZhaSon:C23}
Qiyu Kang, Kai Zhao, Yang Song, Sijie Wang, and Wee~Peng Tay.
\newblock Node embedding from neural {Hamiltonian} orbits in graph neural networks.
\newblock In \emph{Proc. International Conference on Machine Learning}, pp.\  15786--15808, 2023.

\bibitem[Kang et~al.(2024)Kang, Zhao, Song, Xie, Zhao, Wang, She, and Tay]{ZhaKanSon:C24}
Qiyu Kang, Kai Zhao, Yang Song, Yihang Xie, Yanan Zhao, Sijie Wang, Rui She, and Wee~Peng Tay.
\newblock Coupling graph neural networks with fractional order continuous dynamics: {A} robustness study.
\newblock In \emph{Proc. AAAI Conference on Artificial Intelligence}, Vancouver, Canada, 2024.

\bibitem[Kim et~al.(2007)Kim, Goh, Salvi, Oh, Kahng, and Kim]{kim2007fractality}
JS~Kim, K-I Goh, G~Salvi, E~Oh, B~Kahng, and D~Kim.
\newblock Fractality in complex networks: Critical and supercritical skeletons.
\newblock \emph{Physical Review E}, 75\penalty0 (1):\penalty0 016110, 2007.

\bibitem[Kingma \& Ba(2014)Kingma and Ba]{kingma2014adam}
Diederik~P Kingma and Jimmy Ba.
\newblock Adam: A method for stochastic optimization.
\newblock \emph{arXiv preprint arXiv:1412.6980}, 2014.

\bibitem[Kipf \& Welling(2017)Kipf and Welling]{kipf2017semi}
Thomas~N. Kipf and Max Welling.
\newblock Semi-supervised classification with graph convolutional networks.
\newblock In \emph{Proc. Int. Conf. Learn. Representations}, 2017.

\bibitem[Korn \& Korn(2000)Korn and Korn]{korn2000mathematical}
Granino~Arthur Korn and Theresa~M Korn.
\newblock \emph{Mathematical handbook for scientists and engineers: definitions, theorems, and formulas for reference and review}.
\newblock Courier Corporation, 2000.

\bibitem[Krapf(2015)]{krapf2015mechanisms}
Diego Krapf.
\newblock Mechanisms underlying anomalous diffusion in the plasma membrane.
\newblock \emph{Current Topics Membranes}, 75:\penalty0 167--207, 2015.

\bibitem[Li et~al.(2019)Li, Muller, Thabet, and Ghanem]{li2019deepgcns}
Guohao Li, Matthias Muller, Ali Thabet, and Bernard Ghanem.
\newblock Deepgcns: Can gcns go as deep as cnns?
\newblock In \emph{Proc. Int. Conf. Learn. Representations}, pp.\  9267--9276, 2019.

\bibitem[Li et~al.(2020{\natexlab{a}})Li, Xiong, Thabet, and Ghanem]{li2020deepergcn}
Guohao Li, Chenxin Xiong, Ali Thabet, and Bernard Ghanem.
\newblock Deepergcn: All you need to train deeper gcns.
\newblock \emph{arXiv preprint arXiv:2006.07739}, 2020{\natexlab{a}}.

\bibitem[Li et~al.(2022)Li, Zhu, Cheng, Shan, Luo, Li, and Qian]{li2022Glognn}
Xiang Li, Renyu Zhu, Yao Cheng, Caihua Shan, Siqiang Luo, Dongsheng Li, and Weining Qian.
\newblock Finding global homophily in graph neural networks when meeting heterophily.
\newblock In \emph{International Conference on Machine Learning}, pp.\  13242--13256. PMLR, 2022.

\bibitem[Li et~al.(2020{\natexlab{b}})Li, Jin, Xu, and Tang]{li2020deeprobust}
Yaxin Li, Wei Jin, Han Xu, and Jiliang Tang.
\newblock Deeprobust: A pytorch library for adversarial attacks and defenses.
\newblock \emph{arXiv preprint arXiv:2005.06149}, 2020{\natexlab{b}}.

\bibitem[Lim et~al.(2021)Lim, Hohne, Li, Huang, Gupta, Bhalerao, and Lim]{lim2021LINKX}
Derek Lim, Felix Hohne, Xiuyu Li, Sijia~Linda Huang, Vaishnavi Gupta, Omkar Bhalerao, and Ser~Nam Lim.
\newblock Large scale learning on non-homophilous graphs: New benchmarks and strong simple methods.
\newblock \emph{Advances in Neural Information Processing Systems}, 34:\penalty0 20887--20902, 2021.

\bibitem[Liu et~al.(2022)Liu, Wang, Luo, and Luo]{liu2022regularized}
Zijian Liu, Yaning Wang, Yang Luo, and Chunbo Luo.
\newblock A regularized graph neural network based on approximate fractional order gradients.
\newblock \emph{Mathematics}, 10\penalty0 (8):\penalty0 1320, 2022.

\bibitem[Luan et~al.(2022)Luan, Hua, Lu, Zhu, Zhao, Zhang, Chang, and Precup]{luan2022acmgcn}
Sitao Luan, Chenqing Hua, Qincheng Lu, Jiaqi Zhu, Mingde Zhao, Shuyuan Zhang, Xiao-Wen Chang, and Doina Precup.
\newblock Revisiting heterophily for graph neural networks.
\newblock \emph{Advances in neural information processing systems}, 35:\penalty0 1362--1375, 2022.

\bibitem[Lv \& Xu(2016)Lv and Xu]{lv2016error}
Chunwan Lv and Chuanju Xu.
\newblock Error analysis of a high order method for time-fractional diffusion equations.
\newblock \emph{SIAM J. Sci. Comput.}, 38\penalty0 (5):\penalty0 A2699--A2724, 2016.

\bibitem[Machado et~al.(2011)Machado, Kiryakova, and Mainardi]{machado2011recent}
J~Tenreiro Machado, Virginia Kiryakova, and Francesco Mainardi.
\newblock Recent history of fractional calculus.
\newblock \emph{Communications in nonlinear science and numerical simulation}, 16\penalty0 (3):\penalty0 1140--1153, 2011.

\bibitem[Mandelbrot \& Mandelbrot(1982)Mandelbrot and Mandelbrot]{mandelbrot1982fractal}
Benoit~B Mandelbrot and Benoit~B Mandelbrot.
\newblock \emph{The fractal geometry of nature}, volume~1.
\newblock WH freeman New York, 1982.

\bibitem[Maskey et~al.(2023)Maskey, Paolino, Bacho, and Kutyniok]{maskey2023fractional}
Sohir Maskey, Raffaele Paolino, Aras Bacho, and Gitta Kutyniok.
\newblock A fractional graph laplacian approach to oversmoothing.
\newblock \emph{arXiv preprint arXiv:2305.13084}, 2023.

\bibitem[Masters(2004)]{masters2004fractal}
Barry~R Masters.
\newblock Fractal analysis of the vascular tree in the human retina.
\newblock \emph{Annu. Rev. Biomed. Eng.}, 6:\penalty0 427--452, 2004.

\bibitem[McAuley et~al.(2015)McAuley, Targett, Shi, and van~den Hengel]{McAuleySIGIR2015}
Julian McAuley, Christopher Targett, Qinfeng Shi, and Anton van~den Hengel.
\newblock Image-based recommendations on styles and substitutes.
\newblock In \emph{Proc. Int. ACM SIGIR Conf. Res. Develop. Inform. Retrieval}, pp.\  43–52, 2015.

\bibitem[McCallum et~al.(2004)McCallum, Nigam, Rennie, and Seymore]{McCallum2004AutomatingTC}
Andrew McCallum, Kamal Nigam, Jason D.~M. Rennie, and Kristie Seymore.
\newblock Automating the construction of internet portals with machine learning.
\newblock \emph{Inf. Retrieval}, 3:\penalty0 127--163, 2004.

\bibitem[Monti et~al.(2017)Monti, Boscaini, Masci, Rodola, Svoboda, and Bronstein]{monti2017geometric}
Federico Monti, Davide Boscaini, Jonathan Masci, Emanuele Rodola, Jan Svoboda, and Michael~M Bronstein.
\newblock Geometric deep learning on graphs and manifolds using mixture model cnns.
\newblock In \emph{Proc. Conf. Comput. Vision Pattern Recognition}, pp.\  5115--5124, 2017.

\bibitem[Namata et~al.(2012)Namata, London, Getoor, and Huang]{namata:mlg12-wkshp}
Galileo~Mark Namata, Ben London, Lise Getoor, and Bert Huang.
\newblock Query-driven active surveying for collective classification.
\newblock In \emph{Workshop Mining Learn. Graphs}, 2012.

\bibitem[Nigmatullin(1992)]{nigmatullin1992fractional}
Ravil'Rashidovich Nigmatullin.
\newblock Fractional integral and its physical interpretation.
\newblock \emph{Theoretical and Mathematical Physics}, 90\penalty0 (3):\penalty0 242--251, 1992.

\bibitem[Nigmatullin(1986)]{nigmatullin1986realization}
RR~Nigmatullin.
\newblock The realization of the generalized transfer equation in a medium with fractal geometry.
\newblock \emph{Physica status solidi (b)}, 133\penalty0 (1):\penalty0 425--430, 1986.

\bibitem[Notations(2023)]{bigo}
Bachmann–Landau~Order Notations.
\newblock Big o notation, 2023.
\newblock URL \url{https://en.wikipedia.org/wiki/Big_O_notation}.
\newblock Accessed: Sep 1, 2023.

\bibitem[Oono \& Suzuki(2020)Oono and Suzuki]{oonograph}
Kenta Oono and Taiji Suzuki.
\newblock Graph neural networks exponentially lose expressive power for node classification.
\newblock In \emph{Proc. Int. Conf. Learn. Representations}, 2020.

\bibitem[OpenAI(2022)]{openai2022chatgpt4}
OpenAI.
\newblock Chatgpt-4, 2022.
\newblock Available at: \url{https://www.openai.com} (Accessed: 10 April 2024).

\bibitem[Pang et~al.(2019)Pang, Lu, and Karniadakis]{pang2019fpinns}
Guofei Pang, Lu~Lu, and George~Em Karniadakis.
\newblock fpinns: Fractional physics-informed neural networks.
\newblock \emph{SIAM J. Sci. Comput.}, 41\penalty0 (4):\penalty0 A2603--A2626, 2019.

\bibitem[Paszke et~al.(2017)Paszke, Gross, Chintala, Chanan, Yang, DeVito, Lin, Desmaison, Antiga, and Lerer]{paszke2017automatic}
Adam Paszke, Sam Gross, Soumith Chintala, Gregory Chanan, Edward Yang, Zachary DeVito, Zeming Lin, Alban Desmaison, Luca Antiga, and Adam Lerer.
\newblock Automatic differentiation in pytorch.
\newblock In \emph{Advances Neural Inf. Process. Syst.}, 2017.

\bibitem[Pei et~al.(2020)Pei, Wei, Chang, Lei, and Yang]{pei2020geomgcn}
Hongbin Pei, Bingzhe Wei, Kevin Chen-Chuan Chang, Yu~Lei, and Bo~Yang.
\newblock Geom-gcn: Geometric graph convolutional networks.
\newblock \emph{arXiv preprint arXiv:2002.05287}, 2020.

\bibitem[Podlubny(1994)]{podlubny1994fractional}
Igor Podlubny.
\newblock Fractional-order systems and fractional-order controllers.
\newblock \emph{Institute of Experimental Physics, Slovak Academy of Sciences, Kosice}, 12\penalty0 (3):\penalty0 1--18, 1994.

\bibitem[Podlubny(1999)]{podlubny1999fractional}
Igor Podlubny.
\newblock \emph{Fractional Differential Equations}.
\newblock Academic Press, 1999.

\bibitem[Poli et~al.(2019)Poli, Massaroli, Park, Yamashita, Asama, and Park]{poli2019graph}
Michael Poli, Stefano Massaroli, Junyoung Park, Atsushi Yamashita, Hajime Asama, and Jinkyoo Park.
\newblock Graph neural ordinary differential equations.
\newblock \emph{arXiv preprint arXiv:1911.07532}, 2019.

\bibitem[Quaglino et~al.(2019)Quaglino, Gallieri, Masci, and Koutn{\'\i}k]{quaglino2019snode}
Alessio Quaglino, Marco Gallieri, Jonathan Masci, and Jan Koutn{\'\i}k.
\newblock Snode: Spectral discretization of neural odes for system identification.
\newblock In \emph{Proc. Int. Conf. Learn. Representations}, 2019.

\bibitem[Radwan et~al.(2008)Radwan, Elwakil, and Soliman]{radwan2008fractional}
Ahmed~Gomaa Radwan, Ahmed~S Elwakil, and Ahmed~M Soliman.
\newblock Fractional-order sinusoidal oscillators: design procedure and practical examples.
\newblock \emph{IEEE Tran. Circuits and Syst.}, 55\penalty0 (7):\penalty0 2051--2063, 2008.

\bibitem[Rusch et~al.(2022)Rusch, Chamberlain, Rowbottom, Mishra, and Bronstein]{rusch2022graph}
T~Konstantin Rusch, Ben Chamberlain, James Rowbottom, Siddhartha Mishra, and Michael Bronstein.
\newblock Graph-coupled oscillator networks.
\newblock In \emph{Proc. Int. Conf. Mach. Learn.}, 2022.

\bibitem[Scalas et~al.(2000)Scalas, Gorenflo, and Mainardi]{scalas2000fractional}
Enrico Scalas, Rudolf Gorenflo, and Francesco Mainardi.
\newblock Fractional calculus and continuous-time finance.
\newblock \emph{Physica A: Statistical Mechanics and its Applications}, 284\penalty0 (1-4):\penalty0 376--384, 2000.

\bibitem[Sen et~al.(2008)Sen, Namata, Bilgic, Getoor, Galligher, and Eliassi-Rad]{SenNamata2008}
Prithviraj Sen, Galileo Namata, Mustafa Bilgic, Lise Getoor, Brian Galligher, and Tina Eliassi-Rad.
\newblock Collective classification in network data.
\newblock \emph{AI Magazine}, 29\penalty0 (3):\penalty0 93, Sep. 2008.

\bibitem[Shchur et~al.(2018)Shchur, Mumme, Bojchevski, and G{\"u}nnemann]{shchur2018pitfalls}
Oleksandr Shchur, Maximilian Mumme, Aleksandar Bojchevski, and Stephan G{\"u}nnemann.
\newblock Pitfalls of graph neural network evaluation.
\newblock \emph{Relational Representation Learning Workshop, {Advances Neural Inf. Process. Syst.},}, 2018.

\bibitem[She et~al.(2023{\natexlab{a}})She, Kang, Wang, Tay, Guan, Navarro, and Hartmannsgruber]{SheKanWan:J23}
Rui She, Qiyu Kang, Sijie Wang, Wee~Peng Tay, Yong~Liang Guan, Diego~Navarro Navarro, and Andreas Hartmannsgruber.
\newblock Image patch-matching with graph-based learning in street scenes.
\newblock \emph{{IEEE} Trans. Image Process.}, 32:\penalty0 3465--3480, 2023{\natexlab{a}}.

\bibitem[She et~al.(2023{\natexlab{b}})She, Kang, Wang, Y{\'a}ng, Zhao, Song, and Tay]{she2023robustmat}
Rui She, Qiyu Kang, Sijie Wang, Yu{\'a}n-Ru{\`\i} Y{\'a}ng, Kai Zhao, Yang Song, and Wee~Peng Tay.
\newblock Robustmat: Neural diffusion for street landmark patch matching under challenging environments.
\newblock \emph{{IEEE} Trans. Image Process.}, 2023{\natexlab{b}}.

\bibitem[She et~al.(2024{\natexlab{a}})She, Kang, Wang, Tay, Zhao, Song, Geng, Xu, Navarro, and Hartmannsgruber]{SheKanWan:J24}
Rui She, Qiyu Kang, Sijie Wang, Wee~Peng Tay, Kai Zhao, Yang Song, Tianyu Geng, Yi~Xu, Diego~Navarro Navarro, and Andreas Hartmannsgruber.
\newblock {PointDifformer: Robust} point cloud registration with neural diffusion and transformer.
\newblock \emph{{IEEE} Transactions on Geoscience and Remote Sensing}, 62:\penalty0 1 -- 15, 2024{\natexlab{a}}.
\newblock \doi{10.1109/TGRS.2024.3351286}.

\bibitem[She et~al.(2024{\natexlab{b}})She, Wang, Kang, Zhao, Song, Tay, Geng, and Jian]{SheWanKan:C24}
Rui She, Sijie Wang, Qiyu Kang, Kai Zhao, Yang Song, Wee~Peng Tay, Tianyu Geng, and Xingchao Jian.
\newblock {PosDiffNet: Positional} neural diffusion for point cloud registration in a large field of view with perturbations.
\newblock In \emph{Proc. AAAI Conference on Artificial Intelligence}, Vancouver, Canada, 2024{\natexlab{b}}.

\bibitem[Song et~al.(2005)Song, Havlin, and Makse]{song2005self}
Chaoming Song, Shlomo Havlin, and Hernan~A Makse.
\newblock Self-similarity of complex networks.
\newblock \emph{Nature}, 433\penalty0 (7024):\penalty0 392--395, 2005.

\bibitem[Song et~al.(2007)Song, Gallos, Havlin, and Makse]{song2007calculate}
Chaoming Song, Lazaros~K Gallos, Shlomo Havlin, and Hern{\'a}n~A Makse.
\newblock How to calculate the fractal dimension of a complex network: the box covering algorithm.
\newblock \emph{J. Stat. Mech. Theory Exp.}, 2007\penalty0 (03):\penalty0 P03006, 2007.

\bibitem[Song et~al.(2022)Song, Kang, Wang, Zhao, and Tay]{SonKanWan:C22}
Yang Song, Qiyu Kang, Sijie Wang, Kai Zhao, and Wee~Peng Tay.
\newblock On the robustness of graph neural diffusion to topology perturbations.
\newblock In \emph{Advances Neural Inf. Process. Syst.}, 2022.

\bibitem[Sornette(2006)]{sornette2006critical}
Didier Sornette.
\newblock \emph{Critical phenomena in natural sciences: chaos, fractals, selforganization and disorder: concepts and tools}.
\newblock Springer Science \& Business Media, 2006.

\bibitem[Sun \& Wu(2006)Sun and Wu]{sun2006fully}
Zhi-zhong Sun and Xiaonan Wu.
\newblock A fully discrete difference scheme for a diffusion-wave system.
\newblock \emph{Applied Numerical Mathematics}, 56\penalty0 (2):\penalty0 193--209, 2006.

\bibitem[Tarasov(2011)]{tarasov2011fractional}
Vasily~E Tarasov.
\newblock \emph{Fractional dynamics: applications of fractional calculus to dynamics of particles, fields and media}.
\newblock Springer Science \& Business Media, 2011.

\bibitem[Thorpe et~al.(2022)Thorpe, Xia, Nguyen, Strohmer, Bertozzi, Osher, and Wang]{thorpe2022grand++}
Matthew Thorpe, Hedi Xia, Tan Nguyen, Thomas Strohmer, Andrea Bertozzi, Stanley Osher, and Bao Wang.
\newblock Grand++: Graph neural diffusion with a source term.
\newblock In \emph{Proc. Int. Conf. Learn. Representations}, 2022.

\bibitem[Tian et~al.(2015)Tian, Zhou, and Deng]{tian2015class}
WenYi Tian, Han Zhou, and Weihua Deng.
\newblock A class of second order difference approximations for solving space fractional diffusion equations.
\newblock \emph{Math. Comput.}, 84\penalty0 (294):\penalty0 1703--1727, 2015.

\bibitem[Vaswani et~al.(2017)Vaswani, Shazeer, Parmar, Uszkoreit, Jones, Gomez, Kaiser, and Polosukhin]{vaswani2017attention}
Ashish Vaswani, Noam Shazeer, Niki Parmar, Jakob Uszkoreit, Llion Jones, Aidan~N Gomez, {\L}ukasz Kaiser, and Illia Polosukhin.
\newblock Attention is all you need.
\newblock \emph{Advances in neural information processing systems}, 30, 2017.

\bibitem[Veli{\v{c}}kovi{\'{c}} et~al.(2018)Veli{\v{c}}kovi{\'{c}}, Cucurull, Casanova, Romero, Li{\`{o}}, and Bengio]{velickovic2018graph}
Petar Veli{\v{c}}kovi{\'{c}}, Guillem Cucurull, Arantxa Casanova, Adriana Romero, Pietro Li{\`{o}}, and Yoshua Bengio.
\newblock Graph attention networks.
\newblock In \emph{Proc. Int. Conf. Learn. Representations}, pp.\  1--12, 2018.

\bibitem[Wang et~al.(2020)Wang, Luo, Suya, Li, Yang, and Zheng]{Wang2020ScalableAO}
Jihong Wang, Minnan Luo, Fnu Suya, Jundong Li, Zijiang Yang, and Qinghua Zheng.
\newblock Scalable attack on graph data by injecting vicious nodes.
\newblock \emph{Data Mining Knowl. Discovery}, pp.\  1 -- 27, 2020.

\bibitem[Wang et~al.(2022{\natexlab{a}})Wang, Zhang, and Jiang]{wang2022fractional}
Shupeng Wang, Hui Zhang, and Xiaoyun Jiang.
\newblock Fractional physics-informed neural networks for time-fractional phase field models.
\newblock \emph{Nonlinear Dyn.}, 110\penalty0 (3):\penalty0 2715--2739, 2022{\natexlab{a}}.

\bibitem[Wang et~al.(2023)Wang, Kang, She, Tay, Hartmannsgruber, and Navarro]{WanKanShe:C23}
Sijie Wang, Qiyu Kang, Rui She, Wee~Peng Tay, Andreas Hartmannsgruber, and Diego~Navarro Navarro.
\newblock {RobustLoc}: {Robust} camera pose regression in challenging driving environments.
\newblock In \emph{Proc. AAAI Conference on Artificial Intelligence}, Feb. 2023.

\bibitem[Wang et~al.(2022{\natexlab{b}})Wang, Yi, Liu, Wang, and Jin]{wang2022acmp}
Yuelin Wang, Kai Yi, Xinliang Liu, Yu~Guang Wang, and Shi Jin.
\newblock Acmp: Allen-cahn message passing with attractive and repulsive forces for graph neural networks.
\newblock In \emph{Proc. Int. Conf. Learn. Representations}, 2022{\natexlab{b}}.

\bibitem[Waniek et~al.(2018)Waniek, Michalak, Wooldridge, and Rahwan]{WaniekNHB2018}
Marcin Waniek, Tomasz~P. Michalak, Michael~J. Wooldridge, and Talal Rahwan.
\newblock Hiding individuals and communities in a social network.
\newblock \emph{Nature Human Behaviour}, 2\penalty0 (1):\penalty0 139--147, 2018.

\bibitem[Weinan(2017)]{weinan2017proposal}
Ee~Weinan.
\newblock A proposal on machine learning via dynamical systems.
\newblock \emph{Commun. Math. Statist.}, 1\penalty0 (5):\penalty0 1--11, 2017.

\bibitem[Wikipedia(2023)]{feller1991introduction}
Wikipedia.
\newblock Hardy–littlewood tauberian theorem, 2023.
\newblock URL \url{https://en.wikipedia.org/wiki/Hardy%E2%80%93Littlewood_Tauberian_theorem}.
\newblock Accessed: Sep 1, 2023.

\bibitem[Wu et~al.(2021)Wu, Pan, Chen, Long, Zhang, and Yu]{WuTNNLS2021}
Zonghan Wu, Shirui Pan, Fengwen Chen, Guodong Long, Chengqi Zhang, and Philip~S. Yu.
\newblock A comprehensive survey on graph neural networks.
\newblock \emph{{IEEE} Trans. Neural Netw. Learn. Syst.}, 32\penalty0 (1):\penalty0 4--24, 2021.

\bibitem[Xhonneux et~al.(2020)Xhonneux, Qu, and Tang]{xhonneux2020continuous}
Louis-Pascal Xhonneux, Meng Qu, and Jian Tang.
\newblock Continuous graph neural networks.
\newblock In \emph{Proc. Int. Conf. Mach. Learn.}, pp.\  10432--10441, 2020.

\bibitem[Xu et~al.(2018)Xu, Li, Tian, Sonobe, Kawarabayashi, and Jegelka]{xu2018representation}
Keyulu Xu, Chengtao Li, Yonglong Tian, Tomohiro Sonobe, Ken-ichi Kawarabayashi, and Stefanie Jegelka.
\newblock Representation learning on graphs with jumping knowledge networks.
\newblock In \emph{Proc. Int. Conf. Mach. Learn.}, pp.\  5453--5462, 2018.

\bibitem[Yan et~al.(2018)Yan, Du, Tan, and Feng]{yan2019robustness}
Hanshu Yan, Jiawei Du, Vincent~YF Tan, and Jiashi Feng.
\newblock On robustness of neural ordinary differential equations.
\newblock In \emph{Advances Neural Inf. Process. Syst.}, pp.\  1--13, 2018.

\bibitem[Yan et~al.(2022)Yan, Hashemi, Swersky, Yang, and Koutra]{yan2022ggcn}
Yujun Yan, Milad Hashemi, Kevin Swersky, Yaoqing Yang, and Danai Koutra.
\newblock Two sides of the same coin: Heterophily and oversmoothing in graph convolutional neural networks.
\newblock In \emph{2022 IEEE International Conference on Data Mining (ICDM)}, pp.\  1287--1292. IEEE, 2022.

\bibitem[Yue et~al.(2019)Yue, Wang, Huang, Parthasarathy, Moosavinasab, Huang, Lin, Zhang, Zhang, and Sun]{yueBio2019}
Xiang Yue, Zhen Wang, Jingong Huang, Srinivasan Parthasarathy, Soheil Moosavinasab, Yungui Huang, Simon~M Lin, Wen Zhang, Ping Zhang, and Huan Sun.
\newblock Graph embedding on biomedical networks: methods, applications and evaluations.
\newblock \emph{Bioinformatics}, 36\penalty0 (4):\penalty0 1241--1251, 2019.

\bibitem[Yuste \& Acedo(2005)Yuste and Acedo]{yuste2005explicit}
Santos~B Yuste and Luis Acedo.
\newblock An explicit finite difference method and a new von neumann-type stability analysis for fractional diffusion equations.
\newblock \emph{SIAM Journal on Numerical Analysis}, 42\penalty0 (5):\penalty0 1862--1874, 2005.

\bibitem[Zeng et~al.(2020)Zeng, Zhou, Srivastava, Kannan, and Prasanna]{zeng2020graphsaint}
Hanqing Zeng, Hongkuan Zhou, Ajitesh Srivastava, Rajgopal Kannan, and Viktor Prasanna.
\newblock Graphsaint: Graph sampling based inductive learning method, 2020.

\bibitem[Zhang et~al.(2022)Zhang, Cui, and Zhu]{ZhangTKDE2022}
Ziwei Zhang, Peng Cui, and Wenwu Zhu.
\newblock Deep learning on graphs: A survey.
\newblock \emph{{IEEE} Trans. Knowl. Data Eng.}, 34\penalty0 (1):\penalty0 249--270, Jan 2022.

\bibitem[Zhao et~al.(2023{\natexlab{a}})Zhao, Kang, Song, She, Wang, and Tay]{ZhaKanSon:C23}
Kai Zhao, Qiyu Kang, Yang Song, Rui She, Sijie Wang, and Wee~Peng Tay.
\newblock Graph neural convection-diffusion with heterophily.
\newblock In \emph{Proc. Inter. Joint Conf. Artificial Intell.}, Macao, China, 2023{\natexlab{a}}.

\bibitem[Zhao et~al.(2023{\natexlab{b}})Zhao, Kang, Song, She, Wang, and Tay]{ZhaKanSon:C23b}
Kai Zhao, Qiyu Kang, Yang Song, Rui She, Sijie Wang, and Wee~Peng Tay.
\newblock Adversarial robustness in graph neural networks: A {Hamiltonian} energy conservation approach.
\newblock In \emph{Advances in Neural Information Processing Systems}, New Orleans, USA, 2023{\natexlab{b}}.

\bibitem[Zheng et~al.(2022)Zheng, Fei, Li, Liu, Hu, and Sun]{speit_attack}
Qinkai Zheng, Yixiao Fei, Yanhao Li, Qingmin Liu, Minhao Hu, and Qibo Sun.
\newblock Kdd cup 2020 ml track 2 adversarial attacks and defense on academic graph 1st place solution, 2022.
\newblock URL \url{https://github.com/Stanislas0/KDD_CUP_2020_MLTrack2_SPEIT}.
\newblock Accessed: May 1, 2022.

\bibitem[Zhu et~al.(2020{\natexlab{a}})Zhu, Yan, Zhao, Heimann, Akoglu, and Koutra]{zhu2020h2gcn}
Jiong Zhu, Yujun Yan, Lingxiao Zhao, Mark Heimann, Leman Akoglu, and Danai Koutra.
\newblock Beyond homophily in graph neural networks: Current limitations and effective designs.
\newblock \emph{Advances in neural information processing systems}, 33:\penalty0 7793--7804, 2020{\natexlab{a}}.

\bibitem[Zhu et~al.(2021)Zhu, Rossi, Rao, Mai, Lipka, Ahmed, and Koutra]{zhu2021cpgnn}
Jiong Zhu, Ryan~A Rossi, Anup Rao, Tung Mai, Nedim Lipka, Nesreen~K Ahmed, and Danai Koutra.
\newblock Graph neural networks with heterophily.
\newblock In \emph{Proceedings of the AAAI conference on artificial intelligence}, volume~35, pp.\  11168--11176, 2021.

\bibitem[Zhu et~al.(2020{\natexlab{b}})Zhu, Pan, Zhou, Wu, Cao, and Wang]{zhu2020GIL}
Shichao Zhu, Shirui Pan, Chuan Zhou, Jia Wu, Yanan Cao, and Bin Wang.
\newblock Graph geometry interaction learning.
\newblock In \emph{Advances Neural Inf. Process. Syst.}, 2020{\natexlab{b}}.

\bibitem[Zhuang et~al.(2019)Zhuang, Dvornek, Li, and Duncan]{zhuang2019ordinary}
Juntang Zhuang, Nicha Dvornek, Xiaoxiao Li, and James~S Duncan.
\newblock Ordinary differential equations on graph networks.
\newblock 2019.

\bibitem[Zou et~al.(2021)Zou, Zheng, Dong, Guan, Kharlamov, Lu, and Tang]{zouKDD2021}
Xu~Zou, Qinkai Zheng, Yuxiao Dong, Xinyu Guan, Evgeny Kharlamov, Jialiang Lu, and Jie Tang.
\newblock Tdgia: Effective injection attacks on graph neural networks.
\newblock In \emph{Proc. Int. Conf. Knowl. Discovery Data Mining}, pp.\  2461–2471, 2021.

\bibitem[Z{\"u}gner \& G{\"u}nnemann(2019)Z{\"u}gner and G{\"u}nnemann]{zugner_adversarial_2019}
Daniel Z{\"u}gner and Stephan G{\"u}nnemann.
\newblock Adversarial attacks on graph neural networks via meta learning.
\newblock In \emph{Proc. Int. Conf. Learn. Representations}, 2019.

\bibitem[Z\"{u}gner et~al.(2018)Z\"{u}gner, Akbarnejad, and G\"{u}nnemann]{zugnerKDD2018}
Daniel Z\"{u}gner, Amir Akbarnejad, and Stephan G\"{u}nnemann.
\newblock Adversarial attacks on neural networks for graph data.
\newblock In \emph{Proc. Int. Conf. Knowl. Discovery Data Mining}, 2018.

\end{thebibliography}
\bibliographystyle{iclr2024_conference}

\newpage

\appendix

This appendix complements the main body of our paper, providing additional details and supporting evidence for the assertions made therein. The structure of this document is as follows:

\begin{enumerate}
\item We discuss related work in \cref{sec.relatedwork}.
\item We offer a concise review of fractional calculus in \cref{app.review}.
\item We include more solver details and variants in \cref{sec.supp_solvers_frond}.
\item We present dataset statistics, experimental settings, and additional experimental results in \cref{sec:supp_more_exp}.
\item We introduce more dynamics within the FROND framework in \cref{sec.app_moredynamic}.
\item We provide proofs for all theoretical assertions made in the main paper in \cref{sec:supp_proof}.
\item We discuss the limitations of our work and its broader impact in the final section of this supplementary material.
\end{enumerate}

\section{Related Work} \label{sec.relatedwork}

\tb{Fractional Calculus and Its Applications}

The field of fractional calculus has seen a notable surge in interest recently due to its wide-ranging applications across various domains. These include, but are not limited to, numerical analysis \citep{yuste2005explicit}, viscoelastic materials \citep{coleman1961foundations}, population growth models \citep{almeida2016modeling}, control theory \citep{podlubny1994fractional}, signal processing \citep{machado2011recent}, financial mathematics \citep{scalas2000fractional}, and particularly in the representation of porous and fractal phenomena \citep{nigmatullin1986realization,mandelbrot1982fractal,ionescu2017role}.
Within these contexts, FDEs have been developed as a powerful extension to the conventional integer-ordered differential equations, offering a resilient mathematical framework for system analysis \citep{diethelm2002analysis}. To illustrate, in studies related to diffusion processes, researchers have utilized fractional calculus for delineating various natural and synthetic systems, from protein diffusion in cellular membranes \citep{krapf2015mechanisms}, to animal migration patterns \citep{brockmann2006scaling}, human mobility networks \citep{gustafson2017fractional}, and even biological phenomena pertinent to respiratory tissues and neuroscience \citep{ionescu2017role}.
Interestingly, the occurrence of subdiffusion, as modeled by FDEs, has been observed in scenarios where diffusing entities encounter intermittent obstructions due to the complex geometrical structure or interaction dynamics of the environment \citep{diaz2022time,sornette2006critical}.

Within the realm of deep learning, \citep{liu2022regularized} proposes a novel approach to GNN parameter optimization using the fractional derivative. This marks a significant shift from the conventional integer-order derivative employed in optimization algorithms like SGD or Adam \citep{kingma2014adam} with respect to the weights. The essence of their work fundamentally differs from ours, which focuses on the fractional-order evolution of node embeddings, not gradient optimization.
A detailed examination of the study by \citep{liu2022regularized} is pivotal as it adopts fractional derivatives instead of the standard first-order derivatives \emph{during the weight updating phase of a GNN in the gradient descent.} 
Specifically, attention is drawn to equation (16) in \citep{liu2022regularized}, elucidating that the fractional derivative is operational on the loss function. This stands in stark contrast to the FROND framework proposed in this work. As delineated in equation \cref{eq.main_object} of our paper, the fractional derivative is applied to the evolving node feature, representing an implementation of a fractional-order feature updating process, thereby showcasing a clear distinction in the application of fractional derivatives.

Additionally, \citep{antil2020fractional} incorporates insights from fractional calculus and its L1 approximation of the fractional derivative to craft a densely connected neural network. Their aim is to adeptly handle non-smooth data and counteract the vanishing gradient problem. While our research operates within a similar sphere, we have introduced fractional calculus into integer-order continuous GNNs. Our work examines the potential of fractional derivatives in node embedding evolution to address the oversmoothing issue and establishes a connection to non-Markovian dynamic processes. 
Our framework paves the way for a new class of GNNs, enabling a wide spectrum of learnable feature-updating processes influenced by memory effects.

From the perspective of physics-informed machine learning, another line of research is dedicated to crafting neural networks rooted in physical laws to solve fractional PDEs. A pioneering work in this domain is the Fractional Physics Informed Neural Networks (fPINNs) \citep{pang2019fpinns}. Subsequent research, such as \citep{guo2022monte,javadi2023solving,wang2022fractional}, has evolved in this direction. It is worth noting that this line of research is starkly different from our problem formulation.

\tb{Integer-Order Continuous GNNs}

Recent research has illuminated a fascinating intersection between differential equations and neural networks. The concept of continuous dynamical systems as a framework for deep learning has been initially explored by \citep{weinan2017proposal}.
The seminal work of \citep{chen2018neural} introduces neural ODEs with open-source solvers to model continuous residual layers, which has subsequently been applied to the field of GNNs. By utilizing neural ODEs, we can align the inputs and outputs of a neural network with specific physical laws, enhancing the network’s explainability \citep{weinan2017proposal,chamberlain2021grand}. 
Additionally, separate advancements in this domain have led to improvements in neural network performance \citep{dupont2019augmented}, robustness\citep{yan2019robustness,kang2021Neurips}, and gradient stability \citep{haber2017stable,gravina2022anti}. In practical applications, neural ODEs are demonstrating superior performance \citep{SheKanWan:J24,SheWanKan:C24,she2023robustmat,WanKanShe:C23,SheKanWan:J23}. 
In a similar vein, \citep{avelar2019discrete} models continuous residual layers in GCN, leveraging neural ODE solvers to produce output. Further, the work of \citep{poli2019graph} proposes a model that considers a continuum of GNN layers, merging discrete topological structures and differential equations in a manner compatible with various static and autoregressive GNN models. The study \citep{zhuang2019ordinary} introduces GODE, which enables the modeling of continuous diffusion processes on graphs. It also suggests that the oversmoothing issue in GNNs may be associated with the asymptotic stability of ODEs.
 Recently, GraphCON \citep{rusch2022graph} adopts the coupled oscillator model that preserves the graph's Dirichlet energy over time and mitigates the oversmoothing problem. 
In \citep{chamrowgor:grand2021}, the authors modeled information propagation as a diffusion process of a substance from regions of higher to lower concentration. The Beltrami diffusion model is utilized in \citep{charoweyn:blend2021,SonKanWan:C22} to enhance rewiring and improve the robustness of the graph. 
The study by \citep{crifraben:sheaf2022} introduces general sheaf diffusion operators to regulate the diffusion process and maintain non-smoothness in heterophilic graphs, leading to improved node classification performance. Meanwhile, ACMP \citep{wang2022acmp} is inspired by particle reaction-diffusion processes, taking into account repulsive and attractive force interactions between particles. Concurrently, the graph CDE model \citep{ZhaKanSon:C23} is crafted to handle heterophilic graphs and is inspired by the convection-diffusion process. GRAND++ \citep{thorpe2022grand++} leverages heat diffusion with sources to train models effectively with a limited amount of labeled training data.
Concurrently, GREAD \citep{choi2023gread} articulates a GNN approach, which is premised on reaction-diffusion equations, aiming to negotiate heterophilic datasets effectively. In another development, the continuous GNN as an ODE \citep{maskey2023fractional} encapsulates a graph spatial domain rewiring, leveraging the fractional order of the graph Laplacian matrix, presenting a substantial advancement in understanding graph structures. 
We also recommend that interested readers refer to the recent survey \citep{han2023continuous} on continuous GNNs for a more thorough summarization.

\emph{Our FROND extends the above integer-order continuous GNNs by incorporating the Caputo fractional derivative. The models mentioned can be reduced from our unified mathematical framework, with variations manifesting from the choice of the dynamic operator $\calF(\bW,\bX(t))$ in \cref{eq.main_object} and as $\beta$ equals 1 in the fractional derivative operator $D_t^\beta$.}

\tb{Skip Connections in GNNs}

The incorporation of skip or dense connections within network layers has been a transformative approach within deep learning literature. Initially popularized through the ResNet architecture \citep{HeCVPR2016}, this strategy introduces shortcut pathways for gradient flow during backpropagation, thereby simplifying the training of more profound networks. While this architectural design has been instrumental in improving Convolutional Neural Networks (CNNs), it has also been employed in GNNs to bolster their representational capacity and mitigate the vanishing gradient problem.
For example, the Graph U-Net \citep{gao2019graph} employs skip connections to enable efficient information propagation across layers. Similarly, the Jump Knowledge Network \citep{xu2018representation} implements a layer-aggregation mechanism that amalgamates outputs from all preceding layers, a strategy reminiscent of the dense connections found in DenseNet \citep{huang2017densely}. 
Furthermore, the work \citep{chen2020simple} introduces GCNII, an extension of the standard GCN model that incorporates two simple techniques, initial residual and identity mapping, to tackle the oversmoothing problem. Expanding on the idea of depth in GNNs, \citep{li2019deepgcns,li2020deepergcn} propose DeepGCNs, an innovative architecture that employs residual/dense connections along with dilated convolutions. 
The work \citep{di2023understanding} suggests that gradient-flow message passing neural networks may be able to deal with heterophilic graphs provided that a residual connection is available.
The paper \citep{gutteridge2023drew} proposes a spatial domain rewiring and focuses on long-range interactions. DRew in \citep{gutteridge2023drew} does not adhere to any ODE evolutionary structure. Additionally, the skip connection in the vDRew from \citep{gutteridge2023drew} specifically links an $n-k$-th layer to the $n$-th layer. This design is fundamentally different from our FDE approach.

By incorporating fractional calculus and memory effects into our framework, we not only offer a new perspective on understanding the structural design of skip connections in GNNs as a discretized fractional dynamical system, but we also establish a foundation for the development of more versatile and powerful mechanisms for graph representation learning.

\section{Review of Caputo Time-Fractional Derivative} \label{app.review}
We appreciate the need for a more accessible explanation of the Caputo time-fractional derivative and its derivation, as the mathematical intricacies may be challenging for some readers in the GNN community. To address this, we are providing a more comprehensive background in this section. In the main paper, we briefly touched upon fractional calculus, with a particular focus on the \emph{Caputo} fractional derivative that has been employed in our work. In this appendix, we aim to provide a more detailed overview of it and explain why it is widely employed in applications. We have based our FROND framework on the assumption that the solution to the fractional differential equation exists and is unique. The appendix provides explicit conditions for this, which are automatically satisfied in most neural network designs exhibiting local Lipschitz continuity. To simplify, these conditions are akin to those for ordinary differential equations, a common assumption implicitly made in integer-order continuous GNNs such as GRAND \citep{chamberlain2021grand}, GraphCON \citep{rusch2022graph}, GRAND++ \citep{thorpe2022grand++}, GREAD \citep{choi2023gread} and CDE \citep{ZhaKanSon:C23}.

\subsection{Caputo Fractional Derivative and Its Compatibility of Integer-order Derivative}
In the main paper, our focus is predominantly on the order $\beta \in (0,1]$ for the sake of simplification. The Caputo fractional derivative of a function $f(t)$ over an interval $[0,b]$, of a general positive order $\beta \in (0,\infty)$, is defined as follows:
\begin{align}
\label{Cap_Frac}
   D_t^\beta f(t)=\frac{1}{\Gamma(\lceil \beta \rceil-\beta)} \int_0^t(t-\tau)^{\lceil \beta \rceil-\beta-1} f^{[\lceil \beta \rceil]}(\tau) \mathrm{d} \tau,
\end{align}
Here, $\lceil \beta \rceil$ is the smallest integer greater than or equal to $\beta$, $\Gamma(\cdot)$ denotes the gamma function, and $f^{[\lceil \beta \rceil]}(\cdot)$ denotes the $\lceil \beta \rceil$-order derivative of $f(\cdot)$. Within this definition, it is presumed that $f^{[\lceil \beta \rceil]} \in L^1[0,b]$, i.e., $f^{[\lceil \beta \rceil]}$ is Lebesgue integrable, to ensure the well-defined nature of $D_t^\beta f(t)$ as per \eqref{Cap_Frac} \citep{diethelm2010analysis}. For a vector-valued function, the Caputo fractional derivative is defined on a component-by-component basis for each dimension, similar to the integer-order derivative. For ease of exposition, we discuss only the scalar case here, although all the following results can be generalized to vector-valued functions.
The Laplace transform for a general order $\beta \in (0,\infty)$ is presented in Theorem 7.1 \citep{diethelm2010analysis} as:
\begin{align}
\label{L_Cap_Frac}
\mathcal{L} D_t^\beta f(s)=s^\beta \mathcal{L} f(s)-\sum_{k=1}^{\lceil \beta \rceil} s^{\beta -k} f^{[k-1]}(0).
\end{align}
where we assume that the Laplace transform $\mathcal{L}f$ exists on $[s_0,\infty)$ for some $s_0\in\mathbb{R}$. In contrast, for the integer-order derivative $f^{[\beta]}$ where $\beta$ is a positive integer, we also have the formulation \eqref{L_Cap_Frac}, with the only difference being the range of $\beta$. Therefore, as $\beta$ approaches some integer, the Laplace transform of the Caputo fractional derivative converges to the Laplace transform of the traditional integer-order derivative.  \emph{As a result, we can conclude that the Caputo fractional derivative operator generalizes the traditional integer-order derivative since their Laplace transforms coincide when $\beta$ takes an integer value.} The inverse Laplace transform specifies the uniquely determined $D_t^\beta f=f^{[\beta]}$ when $\beta$ is an integer (in the sense of almost everywhere \citep{Cohen2007}).

Under specific reasonable conditions, we can directly present this generalization as follows. Suppose $f^{[\lceil \beta \rceil]}(t)$ \eqref{Cap_Frac} is continuously differentiable. In this context, integration by parts can be utilized to demonstrate that
\begin{align}
D_t^\beta f(t) & =\frac{1}{\Gamma(\lceil\beta\rceil-\beta)}\Bigg(-\left[f^{[\lceil\beta\rceil]}(\tau) \frac{(t-\tau)^{\lceil\beta\rceil-\beta}}{\lceil\beta\rceil-\beta}\right]\bigg|_0^t+\int_0^t f^{[\lceil\beta\rceil+1]}(\tau) \frac{(t-\tau)^{\lceil\beta\rceil-\beta}}{\lceil\beta\rceil-\beta} \ud \tau\Bigg) \nn
& =\frac{t^{\lceil\beta\rceil-\beta} f^{[\lceil\beta\rceil]}(0)}{\Gamma(\lceil \beta \rceil-\beta+1)}+\frac{1}{\Gamma(\lceil \beta \rceil-\beta+1)} \times\int_{0}^{t}(t-\tau)^{\lceil \beta \rceil-\beta} f^{[\lceil\beta\rceil+1]}(\tau) \ud \tau.
\end{align}
As $\beta\rightarrow \lceil \beta \rceil$, we have 
\begin{equation}
\begin{split}
\lim_{\beta\rightarrow \lceil \beta \rceil} D_t^\beta f(t) 
&= f^{[\lceil \beta \rceil]}(0)+\int_0^t f^{[\lceil\beta\rceil+1]}(\tau) \mathrm{d} \tau\\
&= f^{[\lceil \beta \rceil]} (0)+ f^{[\lceil \beta \rceil]}(t) -  f^{[\lceil \beta \rceil]}(0) \\
&= f^{[\lceil \beta \rceil]}(t).
\end{split}
\end{equation}
In parallel to the integer-order derivative, \emph{given certain conditions} (\citep{diethelm2010analysis}[Lemma 3.13]), the Caputo fractional derivative possesses the semigroup property as illustrated in \citep{diethelm2010analysis}[Lemma 3.13]:
\begin{align}
D_t^{\varepsilon} D_t^n f= D_t^{n+\varepsilon} f. 
\end{align}
Nonetheless, it is crucial to recognize that, in general, the Caputo fractional derivative does not exhibit the semigroup property, a characteristic inherent to integer-order derivatives, as detailed in \citep{diethelm2010analysis}[Section 3.1].
The Caputo fractional derivative also exhibits \emph{linearity}, but does not adhere to the same Leibniz and chain rules as its integer counterpart. As such properties are not utilized in our work, we refer interested readers to \citep{diethelm2010analysis}[Theorem 3.17 and Remark 3.5.].

\subsection{Comparison between Riemann–Liouville and Caputo Derivative}
Another well-known fractional derivative is the Riemann–Liouville derivative, which, however, sees less use in practical applications (see \cref{ssec.reason} for more insights). In this section, we offer a succinct introduction to the Riemann–Liouville derivative and compare it with Caputo's definition. The Riemann–Liouville fractional derivative is given as
\begin{align}
\widehat{D}_t^\beta f(t):=\frac{1}{\Gamma(\lceil \beta \rceil-\beta)} \frac{\mathrm{d}^{\lceil \beta \rceil}}{\mathrm{d}t^{\lceil \beta \rceil}} \int_0^t(t-\tau)^{\lceil \beta \rceil-\beta-1} f(\tau) \mathrm{d} \tau  
\end{align}
Here again, we make the assumption that sufficient conditions are satisfied to ensure well-definedness (refer to \citep{diethelm2010analysis}[section 2.2] for details).

We compare the Taylor expansion for the two definitions of fractional derivatives, namely the Riemann-Liouville and Caputo derivatives, with the conventional integer-order derivative. This comparison allows us to clearly highlight the distinctions among the differential equations defined under these three different approaches.

$\bullet$ \tb{Classical Integer-order Taylor Expansion:} \citep{diethelm2010analysis}[Theorem 2.C] Assuming that $f$ has absolutely continuous $(m-1)$-st derivative, we have that for $t\in[0, b]$,
\begin{align}
f(t)=\sum_{k=0}^{m-1} \frac{t^k}{k !} \ddfrac{^k f(0)}{t^k} + J^m\frac{\mathrm{d}^m}{\mathrm{d}t^m} f(t)  \label{eq.R7}
\end{align}
where $J^n f(t):=\frac{1}{\Gamma(n)} \int_0^t(t-\tau)^{n-1} f(\tau) \ud \tau$. Note that here, \emph{$k$ is an integer}.

$\bullet$ \tb{Riemann-Liouville Fractional Taylor Expansion:}  \citep{diethelm2010analysis}[Theorem 2.24] Let $n>0$ and $m=\lfloor n\rfloor+1$. Assume that $f$ is such that $J^{m-n} f$ has absolutely continuous $(m-1)$-st derivative. Then,
\begin{align}
f(t)=\frac{t^{n-m}}{\Gamma(n-m+1)} J^{m-n} f(0)+\sum_{k=1}^{m-1} \frac{t^{k+n-m}}{\Gamma(k+n-m+1)} \widehat{D}_t^{k+n-m} f(0)+J^n \widehat{D}_t^n f(t) . \label{eq.R8}
\end{align}
Note that in the case $n \in \mathbb{N}$ we have $m=n+1$ and $\Gamma(n-m+1)=\Gamma(0)=\infty$, and the first term before the sum vanishes. Hence, we recover the classical result. For general $n$, \emph{the order in $\widehat{D}_t^{k+n-m}$ is not a integer.}

$\bullet$ \tb{Caputo Fractional Taylor Expansion:} \citep{diethelm2010analysis}[Theorem 3.8.] Assume that $n \geq 0, m=\lceil n\rceil$, and $f$ has absolutely continuous $(m-1)$-st derivative. Then
\begin{align}
f(t)=\sum_{k=0}^{m-1} \frac{t^k}{k !}D_t^k f(0) + J^n  D_t^n f(t).  \label{eq.R9}
\end{align}
Note \emph{the order in $D_t^k$ is an integer.} If we compare \cref{eq.R7,eq.R8,eq.R9}, it becomes evident that the Caputo derivative closely resembles the classical integer-order derivative in terms of Taylor expansion. This fact influences the initial conditions for the differential equations introduced in the following section.

\subsection{(Caputo) Fractional Differential Equation}

In this section, we first compare the initial conditions for FDEs under the Riemann-Liouville and Caputo definitions. Following this, we present the precise conditions for the existence and uniqueness of the solution to the fractional differential equation. These conditions closely align with those of ordinary differential equations, which are widely assumed by integer-order continuous GNNs \citep{chamberlain2021grand,rusch2022graph,thorpe2022grand++,choi2023gread,ZhaKanSon:C23}.

\subsubsection{Riemann-Liouville Case} Drawing from the Riemann-Liouville fractional Taylor expansion, let us assume that $e$ is a given function with the property that there exists some function $g$ such that $g=\widehat{D}_t^\beta  e$. The solution of the Riemann-Liouville differential equation of the form
\begin{align}
\widehat{D}_t^\beta f=g   \label{eq.R10}
\end{align}
is given by
\begin{align}
f(t)=e(t)+\sum_{j=1}^{\lceil \beta \rceil} c_j t^{n-j}   \label{eq.R11}
\end{align}
where $c_j$ are arbitrary constants. In other words, to uniquely determine the solution from \cref{eq.R8}, we need to know the value of $\widehat{D}_t^{k+n-m} f(0)$. This is akin to a $k$ order ordinary differential equation where the initial conditions are assumed as $\frac{\mathrm{d}^k}{\mathrm{d}t^k} f(0)$, \emph{with the distinction that the order in $\widehat{D}_t^{k+n-m}$ is not an integer.}

\subsubsection{Caputo Case}\label{ssec.supp_capudfe} Similarly, if $e$ is a given function with the property that $e=D_t^\beta g$ and if we intend to solve
\begin{align}
D_t^\beta f =g \label{eq.R12}
\end{align}
then we find
\begin{align}
f(t)=e(t)+\sum_{j=1}^{\lceil \beta\rceil} c_j t^{\lceil \beta \rceil-j} \label{eq.R13}
\end{align}
once more, with $c_j$ as arbitrary constants. Thus, to obtain a unique solution, it is natural to prescribe the values of \emph{integer order derivatives} $f(0), D_t^1 f(0), \ldots, D_t^{\lceil \beta\rceil-1} f(0)$ in the Caputo setting, \emph{mirroring traditional ordinary differential equations.}

\subsubsection{Existence and Uniqueness of the (Caputo) Solution}
Next, we delve into a general Caputo fractional differential equation, presented as follows:
\begin{align}
D_t^\beta y(t)=g(t, y(t)) \label{eq.R14}
\end{align}
conjoined with suitable initial conditions. As hinted in \cref{eq.R12,eq.R13}, the initial conditions take the form:
\begin{align}
D_t^k y(0)=y_0^{(k)}, \quad k=0,1, \ldots, \lceil \beta\rceil-1. \label{eq.R15}
\end{align}
The following theorem addresses the existence and uniqueness of solutions:
\begin{itemize}[topsep=0pt, itemsep=0pt, partopsep=0pt, parsep=0pt,leftmargin=18pt]
    \item \tb{Caputo existence and uniqueness theorem:} \citep{diethelm2010analysis}[Theorem 6.8] Let $y_0^{(0)}, \ldots, y_0^{(m-1)} \in \mathbb{R}$ and $h^*>0$. Define the set $G:=\left[0, h^*\right] \times \mathbb{R}$
and let the function $g: G \rightarrow \mathbb{R}$ be continuous and fulfill a \emph{Lipschitz condition} with respect to the second variable, i.e.,
\begin{align*}
\left|g\left(x, y_1\right)-g\left(x, y_2\right)\right| \leq L\left|y_1-y_2\right|
\end{align*}
for some constant $L>0$ independent of $x, y_1$, and $y_2$. Then there \emph{uniquely exists} function $y \in C\left[0, h^*\right]$ solving the initial value problem \cref{eq.R14,eq.R15}.
\end{itemize}
For a point of reference, we also provide the well-known Picard–Lindelöf uniqueness theorem for first-order ordinary differential equations.

\begin{itemize}[topsep=0pt, itemsep=0pt, partopsep=0pt, parsep=0pt,leftmargin=18pt]
\item \tb{Picard–Lindelöf theorem} \citep{hartman2002ordinary}[Page 8] Let $D \subseteq \mathbb{R} \times \mathbb{R}^n$ be a closed rectangle with $\left(t_0, y_0\right) \in \operatorname{int} D$, the interior of $D$. Let $g: D \rightarrow \mathbb{R}^n$ be a function that is continuous in $t$ and \emph{Lipschitz continuous} in $y$. Then, there exists some $\varepsilon>0$ such that the initial value problem
\begin{align*}
y^{\prime}(t)=g(t, y(t)), \quad y\left(t_0\right)=y_0 .
\end{align*}
has a \emph{unique solution} $y(t)$ on the interval $\left[t_0, t_0+\varepsilon\right]$.
\end{itemize}

This allows us to draw parallels between the existence and uniqueness theorem of the Caputo fractional differential equation and its integer-order ordinary differential equation equivalent. We also remind readers that standard neural networks, as compositions of linear maps and pointwise non-linear activation functions with bounded derivatives (such as fully-connected and convolutional networks), satisfy global Lipschitz continuity with respect to the input. For attention neural networks, which are compositions of softmax and matrix multiplication, we observe local Lipschitz continuity. 
To see this, suppose $\mathbf{v}=\operatorname{softmax}(\mathbf{u}) \in \mathbb{R}^{n \times 1}$. Then
\begin{align*}
\frac{\ud \mathbf{v}}{\partial \mathbf{u}}=\operatorname{diag}(\mathbf{v})-\mathbf{v} \mathbf{v}^{\top}=\left[\begin{array}{cccc}
v_1\left(1-v_1\right) & -v_1 v_2 & \ldots & -v_1 v_n \\
-v_2 v_1 & v_2\left(1-v_2\right) & \ldots & -v_2 v_n \\
\vdots & \vdots & \ddots & \vdots \\
-v_n v_1 & -v_n v_2 & \ldots & v_n\left(1-v_n\right)
\end{array}\right].
\end{align*}
For bounded input, we have a bounded Jacobian. All the integer-order continuous GNN works, such as recent contributions like \citep{chamberlain2021grand,rusch2022graph,thorpe2022grand++,choi2023gread,ZhaKanSon:C23} assume the uniqueness of the ODE solutions. \emph{This means that all the integer-order continuous GNNs can be extended by our FROND framework with fractional dynamics.} 

\subsection{Reasons for Choosing Caputo Derivative} \label{ssec.reason}

We now explain the reasons behind our preference for the Caputo fractional derivative:

\begin{enumerate}
    \item As previously discussed, Caputo fractional differential equations align with integer-order differential equations concerning initial conditions.
\item The Caputo fractional derivative maintains a more intuitive resemblance to the integer-order derivative and satisfies the significant property of equating to zero when applied to a constant. This property is not satisfied by the Riemann-Liouville fractional derivative. Refer to \citep{diethelm2010analysis}[Example 2.4. and Example 3.1.] for further clarification.
\item Given its widespread application in the literature for practical use cases, numerical methods for solving Caputo fractional differential equations have been meticulously developed and exhaustively analyzed \citep{diethelm2010analysis,diethelm2004detailed,deng2007short}.
\end{enumerate}

\section{Numerical Solvers for FROND} \label{sec.supp_solvers_frond}

We remind readers that numerous methods for training neural ODEs, and consequently updating the weights $\theta$ in the neural network have been proposed. These include the autodifferentiation technique in PyTorch \citep{yan2019robustness,paszke2017automatic}, the adjoint sensitivity method \citep{chen2018neural}, and Snode \citep{quaglino2019snode}.
In our work, we employ the most straightforward autodifferentiation technique for training FROND with fractional neural differential equations, leveraging the numerical solvers outlined in \citep{diethelm2010analysis,diethelm2004detailed,deng2007short}.
While we plan to investigate more sophisticated techniques for training FROND in future work, we have open-sourced our current solver implementations in \url{https://github.com/zknus/torchfde}. We believe these will serve as valuable tools for the GNN community, encouraging the advancement of a unique class of GNNs that incorporate memory effects.

In traditional integer-order continuous GNNs \citep{chamberlain2021grand,thorpe2022grand++,rusch2022graph,SonKanWan:C22,choi2023gread,ZhaKanSon:C23}, the time parameter $t$ serves as a continuous analog to GNN layers, resembling the concept of neural ODEs \citep{chen2018neural} as continuous residual networks. 
Time discretization plays a crucial role in many numerical solvers for neural ODEs. For example, the explicit Euler scheme reduces neural ODEs to residual networks with shared hidden layers \citep{chen2018neural}. More sophisticated discretization methods, such as adaptive step size solvers \citep{atkinson2011numerical}, provide accurate solutions but require additional computational resources.

Unlike prior studies, our work involves fractional-order ODEs, which are more complex than ODEs when the derivative order $\beta$ takes non-integer values.
We present the \emph{fractional Adams–Bashforth–Moulton method} with three variants utilized in this work, demonstrating how the time parameter continues to serve as a continuous analog to the layer index and how the non-local nature of fractional derivatives leads to nontrivial dense or skip connections between layers. Additionally, we also present one implicit L1 solver for solving FROND when $\beta$ is not an integer. It is worth noting that various neural ODE solvers remain applicable for FROND when $\beta$ is an integer.

We first recall the FROND framework
\begin{align*}
D_t^\beta \bX(t)= \calF(\bW,\bX(t)), \quad \beta>0,
\end{align*}
where $\beta$ denotes the fractional order of the derivative, and $\calF$ is a dynamic operator on the graph like the models presented in \cref{sec.diff}. The initial condition is set as $\bX^{[\lceil\beta\rceil-1]}(0) =\ldots = \bX(0)= \bX$ consisting of the preliminary node features, akin to the initial conditions seen in ODEs.

\subsection{Basic predictor} \label{subsec:supp_predictor}

Referencing \citep{diethelm2004detailed}, we first employ a preliminary numerical solver  called ``predictor'' through time discretisation $t_j=jh$, where the discretisation parameter $h$ is a small positive value: 
\begin{align}
{}_{\mathrm{P}}\bX^{(k)}= \sum_{j=0}^{\lceil\beta\rceil-1} \frac{t_{k}^j}{j!} \bX^{[j]}(0)+\frac{1}{\Gamma(\beta)} \sum_{j=0}^{k-1} \mu_{j, k} \calF(\bW,\bX^{(j)}),
\end{align}
where $\mu_{j, n}= \frac{h^\beta}{\beta}\left((n-j)^\beta-(n-1-j)^\beta\right)$, $k$ denotes the discrete time index (iteration), and $t_k=kh$ represents the discretized time steps. $\bX^{(k)}$ is the numerical approximation of  $\bX(t_k)$.
When $\beta=1$, this method simplifies to the Euler solver in \citep{chen2018neural,chamberlain2021grand} as $\mu_{j, n}\equiv h$, yielding ${}_{\mathrm{P}}\bX^{(k)} = \bX^{(k-1)}  + h\calF(\bW,\bX^{(k-1)})$. 
Thus, our basic predictor can be considered as the fractional Euler method or fractional Adams–Bashforth method, which is a generalization of the Euler method used in \citep{chen2018neural,chamberlain2021grand}. However, when $\beta<1$, we need to utilize the full memory $\{\calF(\bW,\bX^{(j))}\}_{j=0}^{k-1}$. 

\begin{figure}[H]
    \centering
    \adjustbox{scale=0.8,center}{

\tikzset{every picture/.style={line width=0.75pt}} 

\begin{tikzpicture}[x=0.75pt,y=0.75pt,yscale=-1,xscale=1]

\draw  [color={rgb, 255:red, 155; green, 155; blue, 155 }  ,draw opacity=1 ][fill={rgb, 255:red, 255; green, 255; blue, 255 }  ,fill opacity=1 ] (190.39,67.6) -- (316.6,45.3) -- (315.25,98.07) -- (189.03,120.37) -- cycle ;
\draw  [fill={rgb, 255:red, 74; green, 144; blue, 226 }  ,fill opacity=1 ] (287.1,76.9) .. controls (286.37,75.4) and (287.66,73.85) .. (289.98,73.44) .. controls (292.31,73.02) and (294.79,73.89) .. (295.52,75.39) .. controls (296.25,76.88) and (294.96,78.43) .. (292.64,78.85) .. controls (290.32,79.27) and (287.84,78.39) .. (287.1,76.9) -- cycle ;
\draw   (239.42,85.71) .. controls (238.69,84.22) and (239.97,82.66) .. (242.3,82.25) .. controls (244.62,81.83) and (247.1,82.7) .. (247.84,84.2) .. controls (248.57,85.69) and (247.28,87.24) .. (244.96,87.66) .. controls (242.63,88.08) and (240.15,87.2) .. (239.42,85.71) -- cycle ;
\draw    (240.13,86.9) -- (219.11,100.76) ;
\draw    (210.62,81.14) -- (238.8,83.77) ;
\draw [color={rgb, 255:red, 2; green, 122; blue, 250 }  ,draw opacity=0.4 ]   (249.8,83.83) -- (262.59,81.47) -- (264.6,81.09) -- (287.23,76.91) ;
\draw [shift={(247.84,84.2)}, rotate = 349.51] [fill={rgb, 255:red, 2; green, 122; blue, 250 }  ,fill opacity=0.4 ][line width=0.08]  [draw opacity=0] (7.2,-1.8) -- (0,0) -- (7.2,1.8) -- cycle    ;
\draw   (202.21,82.65) .. controls (201.47,81.16) and (202.76,79.61) .. (205.09,79.19) .. controls (207.41,78.77) and (209.89,79.64) .. (210.62,81.14) .. controls (211.36,82.63) and (210.07,84.18) .. (207.74,84.6) .. controls (205.42,85.02) and (202.94,84.15) .. (202.21,82.65) -- cycle ;
\draw   (211.79,102.98) .. controls (211.05,101.48) and (212.34,99.93) .. (214.67,99.51) .. controls (216.99,99.1) and (219.47,99.97) .. (220.2,101.46) .. controls (220.94,102.96) and (219.65,104.51) .. (217.32,104.93) .. controls (215,105.34) and (212.52,104.47) .. (211.79,102.98) -- cycle ;
\draw [color={rgb, 255:red, 0; green, 0; blue, 0 }  ,draw opacity=1 ]   (192,149.5) -- (349.34,254.39) ;
\draw [shift={(351,255.5)}, rotate = 213.69] [color={rgb, 255:red, 0; green, 0; blue, 0 }  ,draw opacity=1 ][line width=0.75]    (10.93,-3.29) .. controls (6.95,-1.4) and (3.31,-0.3) .. (0,0) .. controls (3.31,0.3) and (6.95,1.4) .. (10.93,3.29)   ;
\draw  [color={rgb, 255:red, 155; green, 155; blue, 155 }  ,draw opacity=1 ][fill={rgb, 255:red, 255; green, 255; blue, 255 }  ,fill opacity=1 ] (233.99,97.3) -- (360.2,75) -- (358.85,127.77) -- (232.63,150.07) -- cycle ;
\draw  [fill={rgb, 255:red, 74; green, 144; blue, 226 }  ,fill opacity=1 ] (330.7,106.1) .. controls (329.97,104.6) and (331.26,103.05) .. (333.58,102.64) .. controls (335.91,102.22) and (338.39,103.09) .. (339.12,104.59) .. controls (339.85,106.08) and (338.56,107.63) .. (336.24,108.05) .. controls (333.92,108.47) and (331.44,107.59) .. (330.7,106.1) -- cycle ;
\draw   (283.02,114.91) .. controls (282.29,113.42) and (283.57,111.86) .. (285.9,111.45) .. controls (288.22,111.03) and (290.7,111.9) .. (291.44,113.4) .. controls (292.17,114.89) and (290.88,116.44) .. (288.56,116.86) .. controls (286.23,117.28) and (283.75,116.4) .. (283.02,114.91) -- cycle ;
\draw    (283.73,116.1) -- (262.71,129.96) ;
\draw    (254.22,110.34) -- (282.4,112.97) ;
\draw [color={rgb, 255:red, 2; green, 122; blue, 250 }  ,draw opacity=0.4 ]   (293.4,113.03) -- (306.19,110.67) -- (308.2,110.29) -- (330.83,106.11) ;
\draw [shift={(291.44,113.4)}, rotate = 349.51] [fill={rgb, 255:red, 2; green, 122; blue, 250 }  ,fill opacity=0.4 ][line width=0.08]  [draw opacity=0] (7.2,-1.8) -- (0,0) -- (7.2,1.8) -- cycle    ;
\draw   (245.81,111.85) .. controls (245.07,110.36) and (246.36,108.81) .. (248.69,108.39) .. controls (251.01,107.97) and (253.49,108.84) .. (254.22,110.34) .. controls (254.96,111.83) and (253.67,113.38) .. (251.34,113.8) .. controls (249.02,114.22) and (246.54,113.35) .. (245.81,111.85) -- cycle ;
\draw   (255.39,132.18) .. controls (254.65,130.68) and (255.94,129.13) .. (258.27,128.71) .. controls (260.59,128.3) and (263.07,129.17) .. (263.8,130.66) .. controls (264.54,132.16) and (263.25,133.71) .. (260.92,134.13) .. controls (258.6,134.54) and (256.12,133.67) .. (255.39,132.18) -- cycle ;
\draw  [color={rgb, 255:red, 155; green, 155; blue, 155 }  ,draw opacity=1 ][fill={rgb, 255:red, 255; green, 255; blue, 255 }  ,fill opacity=1 ] (279.59,126.1) -- (405.8,103.8) -- (404.45,156.57) -- (278.23,178.87) -- cycle ;
\draw  [fill={rgb, 255:red, 74; green, 144; blue, 226 }  ,fill opacity=1 ] (376.7,134.5) .. controls (375.97,133) and (377.26,131.45) .. (379.58,131.04) .. controls (381.91,130.62) and (384.39,131.49) .. (385.12,132.99) .. controls (385.85,134.48) and (384.56,136.03) .. (382.24,136.45) .. controls (379.92,136.87) and (377.44,135.99) .. (376.7,134.5) -- cycle ;
\draw   (329.02,143.31) .. controls (328.29,141.82) and (329.57,140.26) .. (331.9,139.85) .. controls (334.22,139.43) and (336.7,140.3) .. (337.44,141.8) .. controls (338.17,143.29) and (336.88,144.84) .. (334.56,145.26) .. controls (332.23,145.68) and (329.75,144.8) .. (329.02,143.31) -- cycle ;
\draw    (329.73,144.5) -- (308.71,158.36) ;
\draw    (300.22,138.74) -- (328.4,141.37) ;
\draw [color={rgb, 255:red, 2; green, 122; blue, 250 }  ,draw opacity=0.4 ]   (339.4,141.43) -- (352.19,139.07) -- (354.2,138.69) -- (376.83,134.51) ;
\draw [shift={(337.44,141.8)}, rotate = 349.51] [fill={rgb, 255:red, 2; green, 122; blue, 250 }  ,fill opacity=0.4 ][line width=0.08]  [draw opacity=0] (7.2,-1.8) -- (0,0) -- (7.2,1.8) -- cycle    ;
\draw   (291.81,140.25) .. controls (291.07,138.76) and (292.36,137.21) .. (294.69,136.79) .. controls (297.01,136.37) and (299.49,137.24) .. (300.22,138.74) .. controls (300.96,140.23) and (299.67,141.78) .. (297.34,142.2) .. controls (295.02,142.62) and (292.54,141.75) .. (291.81,140.25) -- cycle ;
\draw   (301.39,160.58) .. controls (300.65,159.08) and (301.94,157.53) .. (304.27,157.11) .. controls (306.59,156.7) and (309.07,157.57) .. (309.8,159.06) .. controls (310.54,160.56) and (309.25,162.11) .. (306.92,162.53) .. controls (304.6,162.94) and (302.12,162.07) .. (301.39,160.58) -- cycle ;
\draw [color={rgb, 255:red, 4; green, 116; blue, 248 }  ,draw opacity=0.4 ]   (339.12,104.59) .. controls (378.42,112.07) and (376.4,117.04) .. (382.32,132.12) ;
\draw [shift={(383,133.8)}, rotate = 247.48] [fill={rgb, 255:red, 4; green, 116; blue, 248 }  ,fill opacity=0.4 ][line width=0.08]  [draw opacity=0] (7.2,-1.8) -- (0,0) -- (7.2,1.8) -- cycle    ;
\draw [color={rgb, 255:red, 4; green, 116; blue, 248 }  ,draw opacity=0.4 ]   (295.52,75.39) .. controls (332.84,84.19) and (330.87,88.96) .. (336.19,101.22) ;
\draw [shift={(337,103)}, rotate = 244.8] [fill={rgb, 255:red, 4; green, 116; blue, 248 }  ,fill opacity=0.4 ][line width=0.08]  [draw opacity=0] (7.2,-1.8) -- (0,0) -- (7.2,1.8) -- cycle    ;
\draw  [color={rgb, 255:red, 155; green, 155; blue, 155 }  ,draw opacity=1 ][fill={rgb, 255:red, 255; green, 255; blue, 255 }  ,fill opacity=1 ] (324.39,154.6) -- (450.6,132.3) -- (449.25,185.07) -- (323.03,207.37) -- cycle ;
\draw  [color={rgb, 255:red, 155; green, 155; blue, 155 }  ,draw opacity=1 ][fill={rgb, 255:red, 255; green, 255; blue, 255 }  ,fill opacity=1 ] (328.39,157.6) -- (454.6,135.3) -- (453.25,188.07) -- (327.03,210.37) -- cycle ;
\draw  [color={rgb, 255:red, 155; green, 155; blue, 155 }  ,draw opacity=1 ][fill={rgb, 255:red, 255; green, 255; blue, 255 }  ,fill opacity=1 ] (333.39,160.1) -- (459.6,137.8) -- (458.25,190.57) -- (332.03,212.87) -- cycle ;
\draw  [color={rgb, 255:red, 155; green, 155; blue, 155 }  ,draw opacity=1 ][fill={rgb, 255:red, 255; green, 255; blue, 255 }  ,fill opacity=1 ] (337.77,163.45) -- (463.99,141.15) -- (462.63,193.91) -- (336.42,216.21) -- cycle ;
\draw  [color={rgb, 255:red, 155; green, 155; blue, 155 }  ,draw opacity=1 ][fill={rgb, 255:red, 255; green, 255; blue, 255 }  ,fill opacity=1 ] (342.27,165.95) -- (468.49,143.65) -- (467.13,196.41) -- (340.92,218.71) -- cycle ;
\draw  [color={rgb, 255:red, 155; green, 155; blue, 155 }  ,draw opacity=1 ][fill={rgb, 255:red, 255; green, 255; blue, 255 }  ,fill opacity=1 ] (347.27,169.45) -- (473.49,147.15) -- (472.13,199.91) -- (345.92,222.21) -- cycle ;
\draw  [color={rgb, 255:red, 155; green, 155; blue, 155 }  ,draw opacity=1 ][fill={rgb, 255:red, 255; green, 255; blue, 255 }  ,fill opacity=1 ] (353.29,175.06) -- (479.5,152.76) -- (478.15,205.53) -- (351.93,227.83) -- cycle ;
\draw  [fill={rgb, 255:red, 74; green, 144; blue, 226 }  ,fill opacity=1 ] (450,183.86) .. controls (449.27,182.36) and (450.56,180.81) .. (452.88,180.4) .. controls (455.21,179.98) and (457.69,180.85) .. (458.42,182.35) .. controls (459.15,183.84) and (457.86,185.39) .. (455.54,185.81) .. controls (453.22,186.23) and (450.74,185.35) .. (450,183.86) -- cycle ;
\draw   (402.32,192.67) .. controls (401.59,191.17) and (402.87,189.62) .. (405.2,189.21) .. controls (407.52,188.79) and (410,189.66) .. (410.74,191.16) .. controls (411.47,192.65) and (410.18,194.2) .. (407.86,194.62) .. controls (405.53,195.04) and (403.05,194.16) .. (402.32,192.67) -- cycle ;
\draw    (403.03,193.86) -- (382.01,207.72) ;
\draw    (373.52,188.1) -- (401.7,190.73) ;
\draw [color={rgb, 255:red, 2; green, 122; blue, 250 }  ,draw opacity=0.4 ]   (412.7,190.79) -- (425.49,188.43) -- (427.5,188.05) -- (450.13,183.86) ;
\draw [shift={(410.74,191.16)}, rotate = 349.51] [fill={rgb, 255:red, 2; green, 122; blue, 250 }  ,fill opacity=0.4 ][line width=0.08]  [draw opacity=0] (7.2,-1.8) -- (0,0) -- (7.2,1.8) -- cycle    ;
\draw   (365.11,189.61) .. controls (364.37,188.12) and (365.66,186.57) .. (367.99,186.15) .. controls (370.31,185.73) and (372.79,186.6) .. (373.52,188.1) .. controls (374.26,189.59) and (372.97,191.14) .. (370.64,191.56) .. controls (368.32,191.98) and (365.84,191.11) .. (365.11,189.61) -- cycle ;
\draw   (374.69,209.94) .. controls (373.95,208.44) and (375.24,206.89) .. (377.57,206.47) .. controls (379.89,206.06) and (382.37,206.93) .. (383.1,208.42) .. controls (383.84,209.92) and (382.55,211.47) .. (380.22,211.89) .. controls (377.9,212.3) and (375.42,211.43) .. (374.69,209.94) -- cycle ;
\draw  [color={rgb, 255:red, 155; green, 155; blue, 155 }  ,draw opacity=1 ][fill={rgb, 255:red, 255; green, 255; blue, 255 }  ,fill opacity=1 ] (398.29,205.1) -- (524.5,182.8) -- (523.15,235.57) -- (396.93,257.87) -- cycle ;
\draw  [fill={rgb, 255:red, 74; green, 144; blue, 226 }  ,fill opacity=1 ] (495,213.9) .. controls (494.27,212.4) and (495.56,210.85) .. (497.88,210.44) .. controls (500.21,210.02) and (502.69,210.89) .. (503.42,212.39) .. controls (504.15,213.88) and (502.86,215.43) .. (500.54,215.85) .. controls (498.22,216.27) and (495.74,215.39) .. (495,213.9) -- cycle ;
\draw   (447.32,222.71) .. controls (446.59,221.22) and (447.87,219.66) .. (450.2,219.25) .. controls (452.52,218.83) and (455,219.7) .. (455.74,221.2) .. controls (456.47,222.69) and (455.18,224.24) .. (452.86,224.66) .. controls (450.53,225.08) and (448.05,224.2) .. (447.32,222.71) -- cycle ;
\draw    (448.03,223.9) -- (427.01,237.76) ;
\draw    (418.52,218.14) -- (446.7,220.77) ;
\draw [color={rgb, 255:red, 2; green, 122; blue, 250 }  ,draw opacity=0.4 ]   (457.7,220.83) -- (470.49,218.47) -- (472.5,218.09) -- (495.13,213.91) ;
\draw [shift={(455.74,221.2)}, rotate = 349.51] [fill={rgb, 255:red, 2; green, 122; blue, 250 }  ,fill opacity=0.4 ][line width=0.08]  [draw opacity=0] (7.2,-1.8) -- (0,0) -- (7.2,1.8) -- cycle    ;
\draw   (410.11,219.65) .. controls (409.37,218.16) and (410.66,216.61) .. (412.99,216.19) .. controls (415.31,215.77) and (417.79,216.64) .. (418.52,218.14) .. controls (419.26,219.63) and (417.97,221.18) .. (415.64,221.6) .. controls (413.32,222.02) and (410.84,221.15) .. (410.11,219.65) -- cycle ;
\draw   (419.69,239.98) .. controls (418.95,238.48) and (420.24,236.93) .. (422.57,236.51) .. controls (424.89,236.1) and (427.37,236.97) .. (428.1,238.46) .. controls (428.84,239.96) and (427.55,241.51) .. (425.22,241.93) .. controls (422.9,242.34) and (420.42,241.47) .. (419.69,239.98) -- cycle ;
\draw [color={rgb, 255:red, 4; green, 116; blue, 248 }  ,draw opacity=1 ]   (293.02,73.39) .. controls (497.47,7.08) and (513.84,135.95) .. (501.69,209.64) ;
\draw [shift={(501.5,210.75)}, rotate = 279.65] [fill={rgb, 255:red, 4; green, 116; blue, 248 }  ,fill opacity=1 ][line width=0.08]  [draw opacity=0] (7.2,-1.8) -- (0,0) -- (7.2,1.8) -- cycle    ;
\draw [color={rgb, 255:red, 4; green, 116; blue, 248 }  ,draw opacity=0.4 ]   (458.42,182.35) .. controls (494.12,191.28) and (495.16,197.09) .. (497.5,208.57) ;
\draw [shift={(497.88,210.44)}, rotate = 257.98] [fill={rgb, 255:red, 4; green, 116; blue, 248 }  ,fill opacity=0.4 ][line width=0.08]  [draw opacity=0] (7.2,-1.8) -- (0,0) -- (7.2,1.8) -- cycle    ;
\draw [color={rgb, 255:red, 4; green, 116; blue, 248 }  ,draw opacity=1 ]   (334.58,102.64) .. controls (474.62,75.43) and (504.7,118.32) .. (499.96,209.06) ;
\draw [shift={(499.88,210.44)}, rotate = 273.19] [fill={rgb, 255:red, 4; green, 116; blue, 248 }  ,fill opacity=1 ][line width=0.08]  [draw opacity=0] (7.2,-1.8) -- (0,0) -- (7.2,1.8) -- cycle    ;
\draw [color={rgb, 255:red, 4; green, 116; blue, 248 }  ,draw opacity=1 ]   (383.58,130.64) .. controls (492.9,114.91) and (495.95,152.49) .. (497.83,208.73) ;
\draw [shift={(497.88,210.44)}, rotate = 268.11] [fill={rgb, 255:red, 4; green, 116; blue, 248 }  ,fill opacity=1 ][line width=0.08]  [draw opacity=0] (7.2,-1.8) -- (0,0) -- (7.2,1.8) -- cycle    ;
\draw [color={rgb, 255:red, 4; green, 116; blue, 248 }  ,draw opacity=1 ]   (290.73,75.78) .. controls (349.79,58.6) and (386.51,101.48) .. (383.25,131.95) ;
\draw [shift={(383,133.8)}, rotate = 279.29] [fill={rgb, 255:red, 4; green, 116; blue, 248 }  ,fill opacity=1 ][line width=0.08]  [draw opacity=0] (7.2,-1.8) -- (0,0) -- (7.2,1.8) -- cycle    ;

\draw (184,220) node [anchor=north west][inner sep=0.75pt]  [color={rgb, 255:red, 0; green, 0; blue, 0 }  ,opacity=1 ] [align=left] {time discretization};
\draw (190.77,103.94) node [anchor=north west][inner sep=0.75pt]  [font=\tiny]  {$\bX(0)$};
\draw (234.37,134.14) node [anchor=north west][inner sep=0.75pt]  [font=\tiny]  {$\bX(t_1)$};
\draw (280.37,161.94) node [anchor=north west][inner sep=0.75pt]  [font=\tiny]  {$\bX(t_2)$};
\draw (353.67,211.9) node [anchor=north west][inner sep=0.75pt]  [font=\tiny]  {$\bX(t_{n-1})$};
\draw (398.67,242.94) node [anchor=north west][inner sep=0.75pt]  [font=\tiny]  {$\bX(t_{n})$};

\end{tikzpicture}

}
    \caption{\small Model discretization in FROND with the basic predictor solver. Unlike the Euler discretization in ODEs, FDEs incorporate connections to historical times, introducing memory effects. Specifically, the dark blue connections observed in FDEs are absent in ODEs. The weight of these skip connections correlates with $\mu_{j, k}(\beta)$ as detailed in \cref{eq.pre}.} 
    \label{fig.block_inf}
\end{figure}
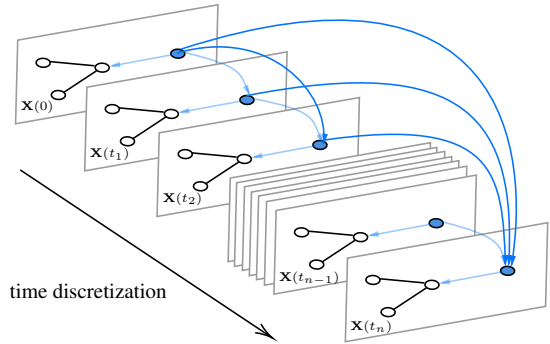

The block diagram of this basic predictor, shown in \cref{fig.block}, reveals that our framework introduces nontrivial dense or skip connections between layers.
A more refined visualization is conveyed in \cref{fig.block_inf}, elucidating the manner in which information propagates through layers and the graph's spatial domain.

\subsection{Predictor-corrector} \label{subsec:supp_corrector}
The corrector formula from \citep{diethelm2004detailed}, a fractional variant of the one-step Adams-Moulton method, refines the initial approximation using the predictor ${}_{\mathrm{P}}\bX^{(k)}$ as follows:
\begin{align}
\bX^{(k)}= \sum_{j=0}^{\lceil\beta\rceil-1} \frac{t_{k}^j}{j !} \bX^{[j]}(0) +\frac{1}{\Gamma(\beta)} \sum_{j=0}^{k-1} \eta_{j, k} \calF(\bW,\bX^{(j)}) +\frac{1}{\Gamma(\beta)} \eta_{k, k} \calF(\bW,{}_{\mathrm{P}}\bX^{(k)}). \label{eq.cor}
\end{align}
Here we show the coefficients $\eta_{j, n}$ in the predictor-corrector variant \cref{eq.cor} from \citep{diethelm2004detailed}:
\begin{align}
\eta_{j, k}(\beta)=\frac{h^\beta}{\beta(\beta+1)} \times 
\begin{cases}
(k-1)^{\beta+1}-(k-1-\beta)k^\beta & \text { if } j=0, \\
(k-j+1)^{\beta+1}+(k-1-j)^{\beta+1}-2(k-j)^{\beta+1} & \text { if } 1 \le j \le k-1, \\ 
1 & \text { if } j=k.
\end{cases} \label{eq.solver_coeffi}
\end{align}

\subsection{Short memory principle} \label{subsec:supp_Short_memory}
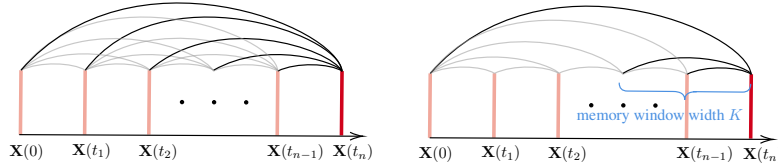
\begin{figure}[H]
    \centering
    \adjustbox{scale=0.6,center}{

\tikzset{every picture/.style={line width=0.75pt}} 

\begin{tikzpicture}[x=0.75pt,y=0.75pt,yscale=-1,xscale=1]

\draw [color={rgb, 255:red, 241; green, 168; blue, 156 }  ,draw opacity=1 ][line width=2.25]    (83.13,100.77) -- (82.64,154.97) ;
\draw [color={rgb, 255:red, 241; green, 168; blue, 156 }  ,draw opacity=1 ][line width=2.25]    (137.09,100.77) -- (136.6,154.97) ;
\draw [color={rgb, 255:red, 241; green, 168; blue, 156 }  ,draw opacity=1 ][line width=2.25]    (245.03,101.47) -- (244.54,155.67) ;
\draw [color={rgb, 255:red, 208; green, 2; blue, 27 }  ,draw opacity=1 ][line width=2.25]    (299,101.47) -- (298.51,155.67) ;
\draw [color={rgb, 255:red, 155; green, 155; blue, 155 }  ,draw opacity=0.56 ][line width=0.75]    (83.13,100.77) .. controls (98.83,93.15) and (126.71,94.31) .. (137.09,100.77) ;
\draw [color={rgb, 255:red, 155; green, 155; blue, 155 }  ,draw opacity=0.56 ][line width=0.75]    (83.13,100.77) .. controls (128.51,75.05) and (184.28,93.73) .. (191.06,101.01) ;
\draw [color={rgb, 255:red, 155; green, 155; blue, 155 }  ,draw opacity=0.56 ][line width=0.75]    (137.09,100.77) .. controls (152.79,93.15) and (180.68,94.31) .. (191.06,100.77) ;
\draw [color={rgb, 255:red, 155; green, 155; blue, 155 }  ,draw opacity=0.56 ][line width=0.75]    (191.06,101.47) .. controls (206.76,93.85) and (234.65,95.02) .. (245.03,101.47) ;
\draw [color={rgb, 255:red, 0; green, 0; blue, 0 }  ,draw opacity=1 ][line width=0.75]    (245.03,101.47) .. controls (260.73,93.85) and (288.62,95.02) .. (299,101.47) ;
\draw [color={rgb, 255:red, 155; green, 155; blue, 155 }  ,draw opacity=0.56 ][line width=0.75]    (137.09,100.77) .. controls (182.48,75.05) and (238.24,93.73) .. (245.03,101.01) ;
\draw [color={rgb, 255:red, 0; green, 0; blue, 0 }  ,draw opacity=1 ][line width=0.75]    (191.06,100.77) .. controls (236.45,75.05) and (292.21,93.73) .. (299,101.01) ;
\draw [color={rgb, 255:red, 155; green, 155; blue, 155 }  ,draw opacity=0.56 ][line width=0.75]    (83.13,100.54) .. controls (140.2,51.12) and (234.65,82.06) .. (245.03,101.47) ;
\draw [color={rgb, 255:red, 0; green, 0; blue, 0 }  ,draw opacity=1 ][line width=0.75]    (137.09,100.77) .. controls (194.17,51.35) and (288.62,82.29) .. (299,101.71) ;
\draw [color={rgb, 255:red, 0; green, 0; blue, 0 }  ,draw opacity=1 ][line width=0.75]    (83.13,100.77) .. controls (146.5,17.84) and (296.71,69.8) .. (299,101.47) ;
\draw    (218.45,127.82) ;
\draw [shift={(218.45,127.82)}, rotate = 0] [color={rgb, 255:red, 0; green, 0; blue, 0 }  ][fill={rgb, 255:red, 0; green, 0; blue, 0 }  ][line width=0.75]      (0, 0) circle [x radius= 1.34, y radius= 1.34]   ;
\draw    (191.47,127.71) ;
\draw [shift={(191.47,127.71)}, rotate = 0] [color={rgb, 255:red, 0; green, 0; blue, 0 }  ][fill={rgb, 255:red, 0; green, 0; blue, 0 }  ][line width=0.75]      (0, 0) circle [x radius= 1.34, y radius= 1.34]   ;
\draw    (164.5,127.59) ;
\draw [shift={(164.5,127.59)}, rotate = 0] [color={rgb, 255:red, 0; green, 0; blue, 0 }  ][fill={rgb, 255:red, 0; green, 0; blue, 0 }  ][line width=0.75]      (0, 0) circle [x radius= 1.34, y radius= 1.34]   ;
\draw [color={rgb, 255:red, 241; green, 168; blue, 156 }  ,draw opacity=1 ][line width=2.25]    (29.16,100.77) -- (28.67,154.97) ;
\draw [color={rgb, 255:red, 155; green, 155; blue, 155 }  ,draw opacity=0.56 ][line width=0.75]    (29.16,100.77) .. controls (44.86,93.15) and (72.74,94.31) .. (83.13,100.77) ;
\draw [color={rgb, 255:red, 155; green, 155; blue, 155 }  ,draw opacity=0.56 ][line width=0.75]    (29.16,100.19) .. controls (74.54,74.47) and (130.31,93.15) .. (137.09,100.42) ;
\draw [color={rgb, 255:red, 155; green, 155; blue, 155 }  ,draw opacity=0.56 ][line width=0.75]    (29.16,100.77) .. controls (86.23,51.35) and (180.68,82.29) .. (191.06,101.71) ;
\draw [color={rgb, 255:red, 155; green, 155; blue, 155 }  ,draw opacity=0.55 ][line width=0.75]    (29.16,100.77) .. controls (92.53,17.84) and (242.74,69.8) .. (245.03,101.47) ;
\draw [color={rgb, 255:red, 0; green, 0; blue, 0 }  ,draw opacity=1 ][line width=0.75]    (29.16,100.77) .. controls (122.21,-5.62) and (294.91,63.84) .. (299,101.94) ;
\draw [color={rgb, 255:red, 241; green, 168; blue, 156 }  ,draw opacity=1 ][line width=2.25]    (427.13,103.44) -- (426.64,157.63) ;
\draw [color={rgb, 255:red, 241; green, 168; blue, 156 }  ,draw opacity=1 ][line width=2.25]    (481.09,103.44) -- (480.6,157.63) ;
\draw [color={rgb, 255:red, 241; green, 168; blue, 156 }  ,draw opacity=1 ][line width=2.25]    (589.03,104.14) -- (588.54,158.33) ;
\draw [color={rgb, 255:red, 208; green, 2; blue, 27 }  ,draw opacity=1 ][line width=2.25]    (643,104.14) -- (642.51,158.33) ;
\draw [color={rgb, 255:red, 155; green, 155; blue, 155 }  ,draw opacity=0.56 ][line width=0.75]    (427.13,103.44) .. controls (442.83,95.81) and (470.71,96.98) .. (481.09,103.44) ;
\draw [color={rgb, 255:red, 155; green, 155; blue, 155 }  ,draw opacity=0.56 ][line width=0.75]    (481.09,103.44) .. controls (496.79,95.81) and (524.68,96.98) .. (535.06,103.44) ;
\draw [color={rgb, 255:red, 155; green, 155; blue, 155 }  ,draw opacity=0.56 ][line width=0.75]    (535.06,104.14) .. controls (550.76,96.51) and (578.65,97.68) .. (589.03,104.14) ;
\draw [color={rgb, 255:red, 0; green, 0; blue, 0 }  ,draw opacity=1 ][line width=0.75]    (589.03,104.14) .. controls (604.73,96.51) and (632.62,97.68) .. (643,104.14) ;
\draw [color={rgb, 255:red, 0; green, 0; blue, 0 }  ,draw opacity=1 ][line width=0.75]    (535.06,103.44) .. controls (580.45,77.72) and (636.21,96.4) .. (643,103.67) ;
\draw    (562.45,130.49) ;
\draw [shift={(562.45,130.49)}, rotate = 0] [color={rgb, 255:red, 0; green, 0; blue, 0 }  ][fill={rgb, 255:red, 0; green, 0; blue, 0 }  ][line width=0.75]      (0, 0) circle [x radius= 1.34, y radius= 1.34]   ;
\draw    (535.47,130.37) ;
\draw [shift={(535.47,130.37)}, rotate = 0] [color={rgb, 255:red, 0; green, 0; blue, 0 }  ][fill={rgb, 255:red, 0; green, 0; blue, 0 }  ][line width=0.75]      (0, 0) circle [x radius= 1.34, y radius= 1.34]   ;
\draw    (508.5,130.26) ;
\draw [shift={(508.5,130.26)}, rotate = 0] [color={rgb, 255:red, 0; green, 0; blue, 0 }  ][fill={rgb, 255:red, 0; green, 0; blue, 0 }  ][line width=0.75]      (0, 0) circle [x radius= 1.34, y radius= 1.34]   ;
\draw [color={rgb, 255:red, 241; green, 168; blue, 156 }  ,draw opacity=1 ][line width=2.25]    (373.16,103.44) -- (372.67,157.63) ;
\draw [color={rgb, 255:red, 155; green, 155; blue, 155 }  ,draw opacity=0.56 ][line width=0.75]    (373.16,103.44) .. controls (388.86,95.81) and (416.74,96.98) .. (427.13,103.44) ;
\draw [color={rgb, 255:red, 155; green, 155; blue, 155 }  ,draw opacity=0.56 ][line width=0.75]    (373.16,102.86) .. controls (418.54,77.13) and (474.31,95.81) .. (481.09,103.09) ;
\draw [color={rgb, 255:red, 155; green, 155; blue, 155 }  ,draw opacity=0.56 ][line width=0.75]    (373.16,103.44) .. controls (430.23,54.02) and (524.68,84.96) .. (535.06,104.37) ;
\draw  [color={rgb, 255:red, 74; green, 144; blue, 226 }  ,draw opacity=1 ] (532,111.25) .. controls (531.96,115.92) and (534.27,118.27) .. (538.94,118.31) -- (577.13,118.66) .. controls (583.8,118.72) and (587.11,121.08) .. (587.06,125.75) .. controls (587.11,121.08) and (590.46,118.78) .. (597.13,118.84)(594.13,118.81) -- (635.44,119.19) .. controls (640.11,119.23) and (642.46,116.92) .. (642.5,112.25) ;
\draw [color={rgb, 255:red, 0; green, 0; blue, 0 }  ,draw opacity=1 ]   (27.67,154.97) -- (314.67,156.32) ;
\draw [shift={(316.67,156.33)}, rotate = 180.27] [color={rgb, 255:red, 0; green, 0; blue, 0 }  ,draw opacity=1 ][line width=0.75]    (10.93,-3.29) .. controls (6.95,-1.4) and (3.31,-0.3) .. (0,0) .. controls (3.31,0.3) and (6.95,1.4) .. (10.93,3.29)   ;
\draw [color={rgb, 255:red, 0; green, 0; blue, 0 }  ,draw opacity=1 ]   (372.67,157.63) -- (659.67,158.99) ;
\draw [shift={(661.67,159)}, rotate = 180.27] [color={rgb, 255:red, 0; green, 0; blue, 0 }  ,draw opacity=1 ][line width=0.75]    (10.93,-3.29) .. controls (6.95,-1.4) and (3.31,-0.3) .. (0,0) .. controls (3.31,0.3) and (6.95,1.4) .. (10.93,3.29)   ;
\draw [color={rgb, 255:red, 0; green, 0; blue, 0 }  ,draw opacity=1 ][line width=0.75]    (373.16,103.44) .. controls (466.21,-2.96) and (638.91,66.51) .. (643,104.61) ;
\draw [color={rgb, 255:red, 155; green, 155; blue, 155 }  ,draw opacity=0.55 ][line width=0.75]    (373.16,103.44) .. controls (436.53,20.51) and (586.74,72.46) .. (589.03,104.14) ;

\draw (18,161.4) node [anchor=north west][inner sep=0.75pt]  [font=\normalsize]  {$\mathbf{X}( 0)$};
\draw (70.67,160.4) node [anchor=north west][inner sep=0.75pt]  [font=\normalsize]  {$\mathbf{X}( t_{1})$};
\draw (126.67,161.07) node [anchor=north west][inner sep=0.75pt]  [font=\normalsize]  {$\mathbf{X}( t_{2})$};
\draw (234.67,161.07) node [anchor=north west][inner sep=0.75pt]  [font=\normalsize]  {$\mathbf{X}( t_{n-1})$};
\draw (289.67,161.73) node [anchor=north west][inner sep=0.75pt]  [font=\normalsize]  {$\mathbf{X}( t_{n})$};
\draw (362,163.4) node [anchor=north west][inner sep=0.75pt]  [font=\normalsize]  {$\mathbf{X}( 0)$};
\draw (414.67,162.4) node [anchor=north west][inner sep=0.75pt]  [font=\normalsize]  {$\mathbf{X}( t_{1})$};
\draw (470.67,163.07) node [anchor=north west][inner sep=0.75pt]  [font=\normalsize]  {$\mathbf{X}( t_{2})$};
\draw (578.67,163.07) node [anchor=north west][inner sep=0.75pt]  [font=\normalsize]  {$\mathbf{X}( t_{n-1})$};
\draw (634.67,163.73) node [anchor=north west][inner sep=0.75pt]  [font=\normalsize]  {$\mathbf{X}( t_{n})$};
\draw (495,131.67) node [anchor=north west][inner sep=0.75pt]  [color={rgb, 255:red, 74; green, 144; blue, 226 }  ,opacity=1 ] [align=left] {memory window width $\displaystyle K$ };

\end{tikzpicture}

    }\vspace{-0.3cm}
    \caption{\small Diagrams of fractional Adams–Bashforth–Moulton method with full (left) and short (right) memory. 
    }\label{fig.block}
\end{figure}

When $T$ is large, computational time complexity becomes a challenge due to the non-local nature of fractional derivatives. To mitigate this, \citep{deng2007short,podlubny1999fractional} suggest leveraging the short memory principle to modify the summation in \cref{eq.pre,eq.cor} to $\sum_{j=n-K}^{n-1}$. This corresponds to employing a shifting memory window with a fixed width $K$. The block diagram is depicted in \cref{fig.block}. 

\subsection{L1 Solver}\label{subs:l1solver}
 The L1 scheme is one of the most popular methods to approximate the Caputo fractional derivative in time. It utilizes a backward differencing method for effective approximation of derivatives. Referencing \citep{gao2011compact,sun2006fully}, we have the L1 approximation of Caputo fractional derivative as follows:
\begin{align*}
D^{\beta}_t \bX^{(k)} \approx \mu \sum_{j=0}^{k-1} R_{k,j}^{\beta}(\bX^{(j+1)}-\bX^{(j)})
\end{align*}
where $h$ is the temporal step size,
\begin{align*}
\mu=\frac{1}{h^{\beta}\Gamma(2-\beta)},\qquad R_{k,j}^{\beta}=(k-j)^{1-\beta}-(k-j-1)^{1-\beta}, \qquad 0\leq j \leq k-1.
\end{align*}

Applying the L1 solver for our problem, we obtain
\begin{align*}
\mu \sum_{j=0}^{k-1} R_{k,j}^{\beta}(\bX^{({j+1})}-\bX^{(j)}) = (\bA(\bX^{(k)})-\bI)\bX^{(k)}.
\end{align*}
 Manipulating the above equation, we have
\begin{align*}
\bX^{(k)}-\frac{1}{\mu}(\bA(\bX^{(k)})-\bI)\bX^{(k)})= \bX^{(k-1)}-\sum_{j=0}^{k-2} R_{k,j}^{\beta}(\bX^{(j+1)}-\bX^{(j)}) 
\end{align*}

The above formula is an implicit nonlinear scheme. To solve it without calculating the inversion of a matrix, we propose the following iteration method:
\begin{enumerate}[label=(\arabic*),topsep=0pt, itemsep=2pt, partopsep=0pt, parsep=2pt,leftmargin=10pt]
    \item Compute a basic approximation of $\bX(t_k)$ with the following formula:
\begin{align*}
{}_{\mathrm{P}}\bX^{(k)}-\frac{1}{\mu}(\bA(\bX^{(k-1)})-\bI)\bX^{(k-1)}= \bX^{(k-1)}-\sum_{j=0}^{k-2} R_{k,j}^{\beta}(\bX^{(j+1)}-\bX^{(j)}). 
\end{align*}
\item Substitute the above ${}_{\mathrm{P}}\bX^{(k)}$ into the implicit scheme to update $\bX^{(k)}$:
\begin{align}
\label{eq.l1solver}
\bX^{(k)}-\frac{1}{\mu}(\bA({}_{\mathrm{P}}\bX^{(k)})-\bI){}_{\mathrm{P}}\bX^{(k)}= \bX^{(k-1)}-\sum_{j=0}^{k-2} R_{k,j}^{\beta}(\bX^{(j+1)}-\bX^{(j)}). 
\end{align}
\end{enumerate}
The step (2) can be repeated multiple times to obtain an accurate approximation of $\bX(t_k)$.

\section{Datasets, Settings and More Experiments for \textbf{F-GRAND} model }\label{sec:supp_more_exp}

\subsection{Datasets}\label{subsec:supp_dataset}
The dataset statistics used in \cref{tab:noderesults} are provided in \cref{tab:nod_dat_sta}. Following the experimental framework in \citep{chamberlain2021grand}, we select the largest connected component from each dataset, except for the tree-like graph datasets (Airport and Disease). However, for the study of oversmoothing, we use a fixed data splitting approach over the entire datasets, as described in \citep{chami2019hyperbolic_GCNN}.

\begin{table}[h]
    \centering
    \caption{Dataset Statistics used in \cref{tab:noderesults}}
    \label{tab:nod_dat_sta}
\begin{tabular}{ccccccc} 
\toprule
Dataset & Type & Classes & Features & Nodes & Edges \\
\hline Cora & citation & 7 & 1433 & 2485 & 5069  \\
Citeseer & citation & 6 & 3703 & 2120 & 3679\\
PubMed & citation & 3 & 500 & 19717 & 44324 \\
Coauthor CS & co-author & 15 & 6805 & 18333 & 81894 \\
Computers & co-purchase & 10 & 767 & 13381 & 245778 \\
Photos & co-purchase & 8 & 745 & 7487 & 119043  \\
CoauthorPhy & co-author  & 5 & 8415 & 34493 & 247962 \\
OGB-Arxiv & citation & 40 & 128 & 169343 & 1166243 \\
Airport & tree-like & 4 & 4 & 3188 & 3188 \\
Disease & tree-like &2 &1000 &1044 &1043 \\
\bottomrule
\end{tabular}
    
\end{table}

\begin{table}[h]
\centering
\caption{Dataset and graph statistics used in \cref{tab.graphclas}}
\label{table:dataset_graph_statistics}
\begin{tabular}{l|c|c|c|c}
\toprule
Dataset & Graphs (Fake) & Total Nodes & Total Edges & Avg. Nodes per Graph \\
Politifact (POL) & 314 (157) & 41,054 & 40,740 & 131 \\
Gossipcop (GOS) & 5464 (2732) & 314,262 & 308,798 & 58 \\
\bottomrule
\end{tabular}

\end{table}

\subsection{Graph Classification Details}
\label{subsec.graph_class_set}
We use the Fake-NewsNet datasets from \citep{dou2021upfd}, constructed based on fact-checking information obtained from Politifact and Gossipcop.
The dataset incorporates four distinct node feature categories, including 768-dimensional BERT features and 300-dimensional spaCy features, which are derived using pre-trained BERT and spaCy word2vec models, respectively. Additionally, a 10-dimensional profile feature is extracted from individual Twitter accounts' profiles. Each graph within the dataset is characterized by a hierarchical tree structure, with the root node representing the news item and the leaf nodes representing Twitter users who have retweeted said news.
An edge exists between a user node and the news node if the user retweeted the original news tweet, while an edge between two user nodes is established when one user retweets the news tweet from another user. This hierarchical organization facilitates the analysis of the spread and influence of both genuine and fabricated news within the Twitter ecosystem. The datasets statistics are summarized in \cref{table:dataset_graph_statistics}.

\subsection{Implementation Details}\label{subsec:supp_imp_det}
Our FROND framework adheres to the experimental settings of the foundational integer-order continuous GNNs, diverging only in the introduction of fractional derivatives in place of integer derivatives. In implementing FROND, we employ one fully-connected (FC) layer on the raw input features to obtain the initial node representations $\bX(0)$. Subsequently, we utilize another FC layer as the decoder function to process the FDE output, $\bX(T)$, for executing downstream tasks.
For more detailed information regarding the hyperparameter settings, we kindly direct the readers to the accompanying supplementary material, which includes the provided code for reproducibility. Our experiments were conducted using NVIDIA RTX A5000 graphics cards.

\subsection{Large scale Ogbn-products dataset}
In this section, we extend our evaluation to include another large-scale dataset, Ogbn-products, adhering to the experimental settings outlined in \citep{hu2021ogbdataset}. For effective handling of this large dataset, we employ a mini-batch training approach, which involves sampling nodes and constructing subgraphs, as proposed by GraphSAINT \citep{zeng2020graphsaint}. 
Upon examination, we observe that F-GRAND-l demonstrates superior performance compared to both GRAND-l and the GCN model, although it falls slightly short of the performance exhibited by GraphSAGE.
This outcome could potentially be attributed to the insufficient dynamic setting in \cref{eq.frac_gra_dif_L}. As such, the more advanced dynamic $\calF(\bW,\bX(t))$ in \cref{eq.main_object} may require additional refinement. 
\begin{table*}[!htb]
    \caption{Node classification accuracy(\%) on Ogbn-products dataset}
    \centering
    \resizebox{1.0\textwidth}{!}{
    \begin{tabular}{c|ccccccc}
    \toprule
        Model &  MLP & Node2vec & Full-batch GCN & GraphSAGE & GRAND-l & F-GRAND-l \\
        \midrule
        Acc &   61.06$\pm$0.08 & 72.49$\pm$0.10 & 75.64$\pm$0.21 & 78.29$\pm$0.16 & 75.56$\pm$0.67 & 77.25$\pm$0.62 \\
    \bottomrule     
    \end{tabular}}
    \label{tab:ogbn-products}
\end{table*}

\subsection{Performance of Different Solver Variants} 
\label{subsec:supp_solver_exp}

In this work, we introduce two types of solvers with distinct variants. We evaluate the performance of these variants in \cref{tab:res_solver}. Specifically, we run F-GRAND on the Cora and Airport datasets with $h=1$ and $T=64$. The solver variants perform comparably. For the Cora dataset, the fractional Adams–Bashforth–Moulton method with a short memory parameter of $K=10$ performs slightly worse than the other variants. However, it demonstrates comparable performance to other solver variants on the Airport dataset.

\label{sec:supp_ablation}
\begin{table}[!htb]
     \caption{Node classification accuracy(\%) under different solver when time $T=64$}
    \centering
    
    \begin{tabular}{c|cccc}
    \toprule
             & Predictor\cref{eq.pre} & Predictor-Corrector \cref{eq.cor} & Short Memory & Implicit L1\\
    \midrule
        Cora($\beta=0.6$) &  83.44$\pm$0.91  &  83.45$\pm$1.09 &  81.51$\pm$1.07 & 82.85$\pm$1.08 \\
         Airport($\beta=0.1$) &  97.41$\pm$0.42   &  96.85$\pm$0.36  &  97.23$\pm$0.59  & 96.06$\pm$1.59 \\

    \bottomrule
    \end{tabular}
   
    \label{tab:res_solver}
\end{table}

\begin{table}[ht]
\caption{Node classification accuracy based on memory $K$ on the Cora dataset when time $T=40$.}
\centering
\resizebox{1.0\textwidth}{!}{
\begin{tabular}{c|c|c|c|c|c|c|c|c|c}
\hline
memory  $K$ & 1 & 5 & 10 & 15 & 20 & 25 & 30  & 35 & 40\\
\hline
Accuracy (\%)  & 74.9$\pm$0.8  & 80.8$\pm$0.8 & 83.3$\pm$1.1 & 83.9$\pm$1.2  & 84.2$\pm$1.1 & 84.1$\pm$1.2  & 84.5$\pm$1.1 &  84.1$\pm$1.1 & 84.8$\pm$1.1 \\
\hline
Inference (ms)  & 9.81   & 17.53  &  24.97  &  32.03 & 38.79  & 42.99  & 45.27 & 48.70 & 48.35\\ 
\hline  
\end{tabular}
}
\label{table:memory_accuracy}
\end{table}

\subsubsection{Further Clarification On Two Accuracies}
This section aims to clarify potential ambiguities surrounding the term ``accuracy'' by distinguishing between ``task accuracy'' and ``numerical accuracy.'' Task accuracy pertains to the performance of GNNs on tasks such as node classification. In contrast, numerical accuracy relates to the precision of numerical solutions to FDEs, a critical concern in mathematics.

For example, generally, a larger $K$ value in the Short Memory solver might enhance both numerical and GNN task accuracy. However, it comes with the trade-off of demanding more computational resources. Furthermore, the two accuracies are related, but not equivalent to each other.
For added clarity, we conducted an ablation study on the Cora dataset, keeping all parameters constant except for the memory parameter $K$. The outcomes of this study are detailed in \cref{table:memory_accuracy}. Our observations indicate that while increasing the value of $K$ can improve numerical accuracy and potentially GNN task accuracy, the computational cost also rises. Notably, the gains in task accuracy plateau beyond a $K$ value of 15.

We also remind the readers that in the literature, to solve FDEs, there exist other more numerically accurate solvers like \citep{jin2017correction,tian2015class,lv2016error} that use higher convergence order. In general, these kinds of solvers can theoretically reduce computation cost and memory storage, as we can obtain the same numerical accuracy using larger step sizes compared to lower-order solvers. It does not aim to improve GNN task accuracy as we can take smaller step sizes to achieve this, but it may be helpful for other performances like computation cost and memory storage reduction. In our paper, we focus on task accuracy. Therefore, classical solvers are used in our work. 
Nonetheless, more numerically accurate solvers could potentially benefit other applications of fractional dynamics, particularly when GNNs are utilized to simulate and forecast real physical systems.

\subsection{Computation Time}
\label{subsec:supp_computetime}
It should be emphasized that our FROND framework \emph{does not introduce any additional training parameters} to the backbone integer-order continuous GNNs. Instead, we simply modify the integration method from standard integration to fractional integration.

In this section, we report the inference time of the different solver variants in \cref{tab:time_solverT4,tab:time_solverT64,tab:time_correctorT4,tab:time_correctorT64}. For comparison, we consider the neural ODE solver for $\beta=1$, which includes Euler, RK4, Implicit Adams, and dopri5 methods as per in the paper \citep{chen2018neural}. We observe that when $T=4$, the inference time required by the FROND solver variants is similar to that of the ODE Euler solver. However, for larger $T=64$, the basic Predictor \cref{eq.pre} solver requires more inference time than Euler and is comparable to RK4. For more accurate approximation solver variants \cref{eq.cor,eq.l1solver} incorporating the corrector formula, \cref{tab:time_correctorT4,tab:time_correctorT64} show that these methods require more computational time as the number of iterations increases. While the advantages of these solvers might not be pronounced for GNN node classification tasks, they could provide benefits for other applications of fractional dynamics, such as when GNNs are used to simulate and forecast real physical systems.

\begin{table}[!htb]
     \caption{Average time under different solvers when time $T=4$ and hidden dimension is 64 on Cora dataset}
    \centering
\resizebox{1\textwidth}{!}{
    \begin{tabular}{c|cccccccc}
    \toprule
             & Predictor\cref{eq.pre} & Predictor-Corrector\cref{eq.cor} & Short Memory & Implicit L1 & Euler & RK4 & Implicit Adams & dopri5 \\
    \midrule
       Inference time (ms) & 0.98 & 1.67 & 0.98 & 0.62 & 0.96 & 2.06 & 3.20 &  11.91\\
    \bottomrule
    \end{tabular}}
   
    \label{tab:time_solverT4}
\end{table}

\begin{table}[!htb]
     \caption{Average time under different solvers when time $T=64$ and hidden dimension is 64 on Cora dataset}
    \centering
    \tiny
\resizebox{1\textwidth}{!}{
    \begin{tabular}{c|cccccccc}
    \toprule
             & Predictor\cref{eq.pre} & Predictor-Corrector\cref{eq.cor} & Short Memory & Implicit L1 & Euler & RK4 & Implicit Adams & dopri5 \\
    \midrule
       Inference time (ms) & 44.46 & 160.92  & 30.26  & 221.74  & 12.16 & 42.66  & 103.46  & 66.15 \\

    \bottomrule
    \end{tabular}}
   
    \label{tab:time_solverT64}
\end{table}

\begin{table}[!htb]
     \caption{Average time of \cref{eq.cor,eq.l1solver} with correctors, used to refine the approximation, when time $T=4$ and hidden dimension is 64 on the Cora dataset.}
    \centering
\makebox[\textwidth][c]{
    
    \begin{tabular}{c|cccc}
    \toprule
          Predictor-Corrector \cref{eq.cor}   & 1  & 3  & 5 & 10  \\
    \midrule
       Inference time (ms) & 1.67 & 3.31  & 4.74  & 8.34  \\

    \bottomrule
    \end{tabular}}

    \begin{tabular}{c|cccc}
    \toprule
          Implicit-L1 \cref{eq.l1solver}   & 1  & 3  & 5 & 10  \\
    \midrule
       Inference time (ms) & 0.62 & 1.04 & 1.48  & 2.55  \\

    \bottomrule
    \end{tabular}
   
    \label{tab:time_correctorT4}
\end{table}

\begin{table}[!htb]
     \caption{Average time of \cref{eq.cor,eq.l1solver} with correctors, used to refine the approximation, when time $T=64$ and hidden dimension is 64 on the Cora dataset.}
    \centering
\makebox[\textwidth][c]{
    
    \begin{tabular}{c|cc}
    \toprule
          Predictor-Corrector \cref{eq.cor}   & 1  & 3    \\
    \midrule
       Inference time (ms) & 160.92 & 442.88    \\

    \bottomrule
    \end{tabular}}

    \begin{tabular}{c|cc}
    \toprule
          Implicit-L1 \cref{eq.l1solver}   & 1  & 3    \\
    \midrule
       Inference time (ms) & 221.74 &   441.60  \\

    \bottomrule
    \end{tabular}
   
    \label{tab:time_correctorT64}
\end{table}

\subsection{Continued Study of Oversmoothing}
\label{subsec.over-smooth_set}

To corroborate that FROND mitigates the issue of oversmoothing and performs well with an increasing number of layers, we conducted an experiment employing the basic predictor with up to 128 layers in the main paper. The results are presented in \cref{fig:over-smooth}. For this experiment, we utilized the fixed data splitting approach for the Cora and Citeseer dataset without using the Largest Connected Component (LCC)  as described in \citep{chami2019hyperbolic_GCNN}.

In the supplementary material, we further probe oversmoothing by conducting experiments with an increased number of layers, reaching up to 256. The results of these experiments are illustrated in \cref{tab:over-smooth1}. From our observations, F-GRAND-l maintains a consistent performance level even as the number of layers escalates.
This contrasts with GRAND-l, where there is a notable performance decrease with the increase in layers. For instance, on the Cora datasets, the accuracy of GRAND-l drops from 81.29\% with 4 layers to 73.37\% with 256 layers. In stark contrast, our F-GRAND-l model exhibits minimal performance decrease on this dataset. On the Airport dataset, F-GRAND-l registers a slight decrease to 94.91\% with 256 layers from 97.0\% with 4 layers. However, the performance of GRAND-l significantly drops to 53.0\%. These observations align with our expectations, as \cref{thm.rate} predicts a slow algebraic convergence rate, while GRAND exhibits a more rapid performance degradation.

Additionally, we note that the optimal number of layers for F-GRAND is 64 on the Cora and Airport datasets, whereas on the Cirtesser dataset, the best performance is achieved with 16 layers.

\begin{table*}[!htb]
\centering
\caption{oversmoothing mitigation under fixed data splitting without LCC} \label{tab:over-smooth1} 
\tiny
\makebox[\textwidth][c]{
\begin{tabular}{cccccccccccc} 
\toprule
Dataset & Model  & 4 & 8 & 16 & 32 & 64  & 80 & 128 &256  \\

\midrule
\multirow{4}{*}{Cora} 
& GCN &  81.35$\pm$1.27 & 15.3$\pm$3.63 & 19.70$\pm$7.06 & 21.86$\pm$6.09 & 13.0$\pm$0.0 & 13.0$\pm$0.0 & 13.0$\pm$0.0  & 13.0$\pm$0.0 \\

& GAT &  80.95$\pm$2.28 & 31.90$\pm$0.0 & 31.90$\pm$0.0 & 31.90$\pm$0.0 & 31.90$\pm$0.0 &31.90$\pm$0.0 & 31.90$\pm$0.0  & 31.90$\pm$0.0 \\

& GRAND-l &  81.29$\pm$0.43 & 82.95$\pm$0.52 & 82.48$\pm$0.46 & 81.72$\pm$0.35 & 81.33$\pm$0.22 & 81.07$\pm$0.44 & 80.09$\pm$0.43 & 73.37$\pm$0.59 \\

& F-GRAND-l & 81.17$\pm$0.75 & 82.68$\pm$0.64 & 83.05$\pm$0.81 & 82.90$\pm$0.81 & 83.44$\pm$0.91 & 82.85$\pm$0.89 & 82.34$\pm$0.83  & 
 81.74$\pm$0.53 \\

\midrule
\multirow{4}{*}{Citeseer} 
& GCN &  68.84$\pm$2.46 & 61.58$\pm$2.09 & 10.64$\pm$1.79 & 7.7$\pm$0.0 & 7.7$\pm$0.0 & 7.7$\pm$0.0 & 7.7$\pm$0.0 & 7.7$\pm$0.0 \\ 

& GAT &  65.20$\pm$0.57 & 18.10$\pm$0.0 &  18.10$\pm$0.0 & 18.10$\pm$0.0 &18.10$\pm$0.0 & 18.10$\pm$0.0 & 18.10$\pm$0.0 & 18.10$\pm$0.0\\

& GRAND-l &    70.68$\pm$1.23 & 70.39$\pm$0.68 & 70.18$\pm$0.56 & 68.90$\pm$1.50 & 68.01$\pm$1.47 &  67.44$\pm$1.25  & 63.45$\pm$2.86 & 56.98$\pm$1.26  \\

& F-GRAND-l &   70.68$\pm$1.23 & 71.04$\pm$0.68 & 71.08$\pm$1.12 & 70.83$\pm$0.90 & 70.27$\pm$0.86 & 70.50$\pm$0.76 & 70.32$\pm$1.67  & 71.0$\pm$0.45\\

\midrule
\multirow{4}{*}{Airport} 
& GCN &  84.77$\pm$1.45 & 74.43$\pm$8.19 & 62.56$\pm$2.16 & 15.27$\pm$0.0 & 15.27$\pm$0.0 & 15.27$\pm$0.0 &  15.27$\pm$0.0 &  15.27$\pm$0.0 \\ 

& GAT &  83.59$\pm$1.51 & 67.02$\pm$4.70 & 46.56$\pm$0.0 & 46.56$\pm$0.0 & 46.56$\pm$0.0 & 46.56$\pm$0.0 & 46.56$\pm$0.0 & 46.56$\pm$0.0 \\

& GRAND-l & 80.53$\pm$9.59  & 79.88$\pm$9.67 & 76.24$\pm$3.80 & 68.67$\pm$4.02 & 62.28$\pm$10.83 &  50.38$\pm$2.98 & 57.96$\pm$11.63 & 53.0$\pm$14.85\\

& F-GRAND-l & 97.0$\pm$0.79 & 97.09$\pm$0.87  & 96.97$\pm$0.84 & 96.50$\pm$0.60 & 97.41$\pm$0.42 & 96.53$\pm$0.74 & 97.03$\pm$0.55 & 94.91$\pm$3.72 \\

\bottomrule
\end{tabular}}

\end{table*}

\subsection{Ablation Study: Selection of $\beta$ Continued} \label{sec.betafull}
In the main paper, we explore the impact of the fractional order parameter $\beta$ across a variety of graph datasets, with the results of these investigations presented in \cref{tab:nodeclasbeta}. More comprehensive details concerning the variations in $\beta$ can be found in \cref{tab:betat8}.

\begin{table*}
    \tiny
    \centering
    \caption{Node classification accuracy(\%) under different value of $\beta$ when time $T=8$.}
    \resizebox{1.0\textwidth}{!}{
    \begin{tabular}{c|cccccccccc}
    \toprule
      $\beta$  & 0.1 & 0.2 & 0.3 & 0.4 & 0.5 & 0.6 & 0.7 & 0.8 & 0.9 & 1.0  \\
      \midrule
      Cora & 74.80$\pm$0.42 & 76.10$\pm$0.34 & 77.0$\pm$0.98  & 77.80$\pm$0.75  & 79.60$\pm$0.91  & 80.79$\pm$0.58 &   81.56$\pm$0.30  &   82.44$\pm$0.51 &   82.68$\pm$0.64 &  82.37$\pm$0.59    \\
      \midrule
      Airport & 97.09$\pm$0.87  &  96.67$\pm$0.91 & 95.80$\pm$2.03 & 94.04$\pm$3.62 & 91.66$\pm$6.34 & 89.24$\pm$7.87 & 84.36$\pm$8.04 & 79.29$\pm$6.01 &  78.73$\pm$6.33 &  78.88$\pm$9.67 \\
    \bottomrule
    \end{tabular}}

    \label{tab:betat8}
\end{table*}

\subsection{Robustness Against Adversarial Attacks}
\label{subsec:supp_robust_exp}
Despite the significant advancements GNNs have made in inference tasks on graph-structured data, they are recognized as being susceptible to adversarial attacks \citep{zugnerKDD2018}. Adversaries, aiming to deceive a trained GNN, can either introduce new nodes into the graph during the inference phase, known as an injection attack \citep{Wang2020ScalableAO,speit_attack,zouKDD2021,hussain2022adversarial}, or manipulate the graph's topology by adding or removing edges, termed as a modification attack \citep{Chen2018FastGA,WaniekNHB2018,Du2017TopologyAG}.
In this section, we present preliminary experiments assessing the robustness of our model against adversarial attacks. Specifically, we carry out graph modification adversarial attacks using the Metattack method \citep{zugner_adversarial_2019}. Our approach adheres to the attack setting described in Pro-GNN \citep{jin2020prognn}, and we utilize the perturbed graph provided by the DeepRobust library \citep{li2020deeprobust} to ensure a fair comparison. The perturbation rate, indicating the proportion of altered edges, is incrementally adjusted in 5\% steps from 0\% to 25\%.

The results of these experiments are presented in \cref{tab:metattack}. 
It should be noted that the impact of Meta-attacks with higher strengths detrimentally affects the performance of all models under test. However, our FROND-nl model consistently demonstrates enhanced resilience against adversarial attacks compared to the baselines, including GRAND-nl. 
For instance, at a perturbation rate of 25\%, F-GRAND-nl outshines the baselines by an estimated margin of 10-15\% on the Cora dataset.

Comprehensive testing against various adversarial attack methods and a theoretical understanding are detailed in our recent work \citep{ZhaKanSon:C24}.

\begin{table}[H]
\centering
\caption{Node classification accuracy (\%) under {\bf modification, poisoning, non-targeted}  attack (Metattack) in {\bf transductive} learning.  
}
\small
\makebox[\textwidth][c]{
\begin{tabular}{cccccc} 
\toprule
Dataset & Ptb Rate(\%)   & GGN & GAT & GRAND-nl  & F-GRAND-nl   \\

\midrule
\multirow{6}{*}{Cora} 
& 0 &  83.50$\pm$0.44 &  83.97$\pm$0.65 & 83.14$\pm$1.06 & 83.48$\pm$1.08    \\

& 5 &  76.55$\pm$0.79 &  80.44$\pm$0.74 & 80.54$\pm$1.17 &  80.25$\pm$0.90    \\

& 10 &  70.39$\pm$1.28 &  75.61$\pm$0.59 & 76.59$\pm$1.21 & 77.94$\pm$0.48    \\

& 15 &  65.10$\pm$0.71 & 69.78$\pm$1.28 & 71.62$\pm$1.39 &  75.14$\pm$1.16    \\

& 20 &  59.56$\pm$2.72 & 59.94$\pm$0.92 & 57.52$\pm$1.20 &  69.04$\pm$1.13    \\

& 25 &  47.53$\pm$1.96 & 54.78$\pm$0.74 & 53.70$\pm$1.91 &  63.40$\pm$1.44    \\

\midrule
\multirow{3}{*}{Citeseer} 
& 0 &  71.96$\pm$0.55 & 73.26$\pm$0.83 & 71.40$\pm$1.08 &  70.14$\pm$0.83    \\

& 5 &  70.88$\pm$0.62 & 72.89$\pm$0.83 & 70.99$\pm$1.12 &  70.0$\pm$1.72    \\

& 10 &  67.55$\pm$0.89 & 70.63$\pm$0.48 & 68.83$\pm$1.31 &  68.64$\pm$1.11    \\

& 15 &  64.52$\pm$1.11 &  69.02$\pm$1.09 & 66.78$\pm$0.92 &  67.90$\pm$0.41    \\

& 20 &  62.03$\pm$3.49 & 61.04$\pm$1.52 & 58.95$\pm$1.33 &  65.84$\pm$0.75    \\

& 25 &  56.94$\pm$2.09 & 61.85$\pm$1.12 &  60.52$\pm$1.29 &  66.50$\pm$1.16    \\

\bottomrule
\end{tabular}}
\vspace{-0.1cm}
 \label{tab:metattack} 
\vspace{-0.45cm}
\end{table}

\subsection{Comparison between Riemann-Liouville (RL) derivative and Caputo Derivative}
The underlying rationale for opting for the Caputo derivative over the Riemann-Liouville (RL) derivative is extensively delineated in \cref{ssec.reason}. However, a supplementary experiment was conducted utilizing the RL derivative in lieu of the Caputo derivative, the results of which are documented in  \cref{tab:rl_cap}.
It can be observed that the task accuracies for both approaches are very similar. 
Further investigations on the use of different fractional derivatives and how to optimize the whole model architecture to adapt to a particular choice will be explored in future work.

\begin{table*}[!ht]
\caption{Comparison between RL-GRAND-l (using Riemann-Liouville derivative) and the original F-GRAND-l (using Caputo derivative).}
\centering
\resizebox{1.0\textwidth}{!}{
\begin{tabular}{c|ccccccccc}
\toprule
Method & Cora & Citeseer & Pubmed  & CoauthorCS  & Computer   & Photo   & CoauthorPhy  & Airport & Disease \\
\midrule
GRAND-l & 83.6$\pm$1.0 & 73.4$\pm$0.5 & 78.8$\pm$1.7  & 92.9$\pm$0.4     & 83.7$\pm$1.2  & 92.3$\pm$0.9   &  93.5$\pm$0.9 & 80.5$\pm$9.6 & 74.5$\pm$3.4 \\
RL-GRAND-l & 84.6$\pm$1.2 & 74.2$\pm$1.0 & 80.1$\pm$1.2 & 92.8$\pm$0.3  & 87.4$\pm$1.1  & 93.3$\pm$0.7  & 94.1$\pm$0.3 & 96.2$\pm$0.2 & 90.7$\pm$1.3 \\
F-GRAND-l & 84.8$\pm$1.1 &  74.0$\pm$1.5  & 79.4$\pm$1.5 & 93.0$\pm$0.3    & 84.4$\pm$1.5  & 92.8$\pm$0.6   &  94.5$\pm$0.4 & 98.1$\pm$0.2 &  92.4$\pm$3.9  \\
\bottomrule
\end{tabular}
}
\label{tab:rl_cap}
\end{table*}

\subsection{Fractal Dimension of Graph Datasets} \label{sec.app_fractal}
\begin{table}[ht]
\centering
\caption{Comparison between  the estimated fractal dimension, the best order $\beta$ and the $\delta$-hyperbolicity} \label{table:fractal}
\begin{tabular}{cccccccc}
Dataset & Disease & Airport & Pubmed & Citeseer & Cora \\
\midrule
fractal dimension  & 2.47  &2.17  &2.25  &0.62 & 1.22 \\
\midrule
best $\beta$ (F-GRAND-l) &0.6  &0.5  &0.9  & 0.9 &0.9  \\
best $\beta$ (F-GRAND-nl)  &0.7 &0.1  &0.4  & 0.9 &0.9  \\
$\delta$-hyperbolicity  & {0.0} &{1.0} &{3.5} &{4.5} &{11.0}\\
\end{tabular} 
\end{table}

\begin{figure}
    \centering
    \includegraphics[width=0.7\textwidth]{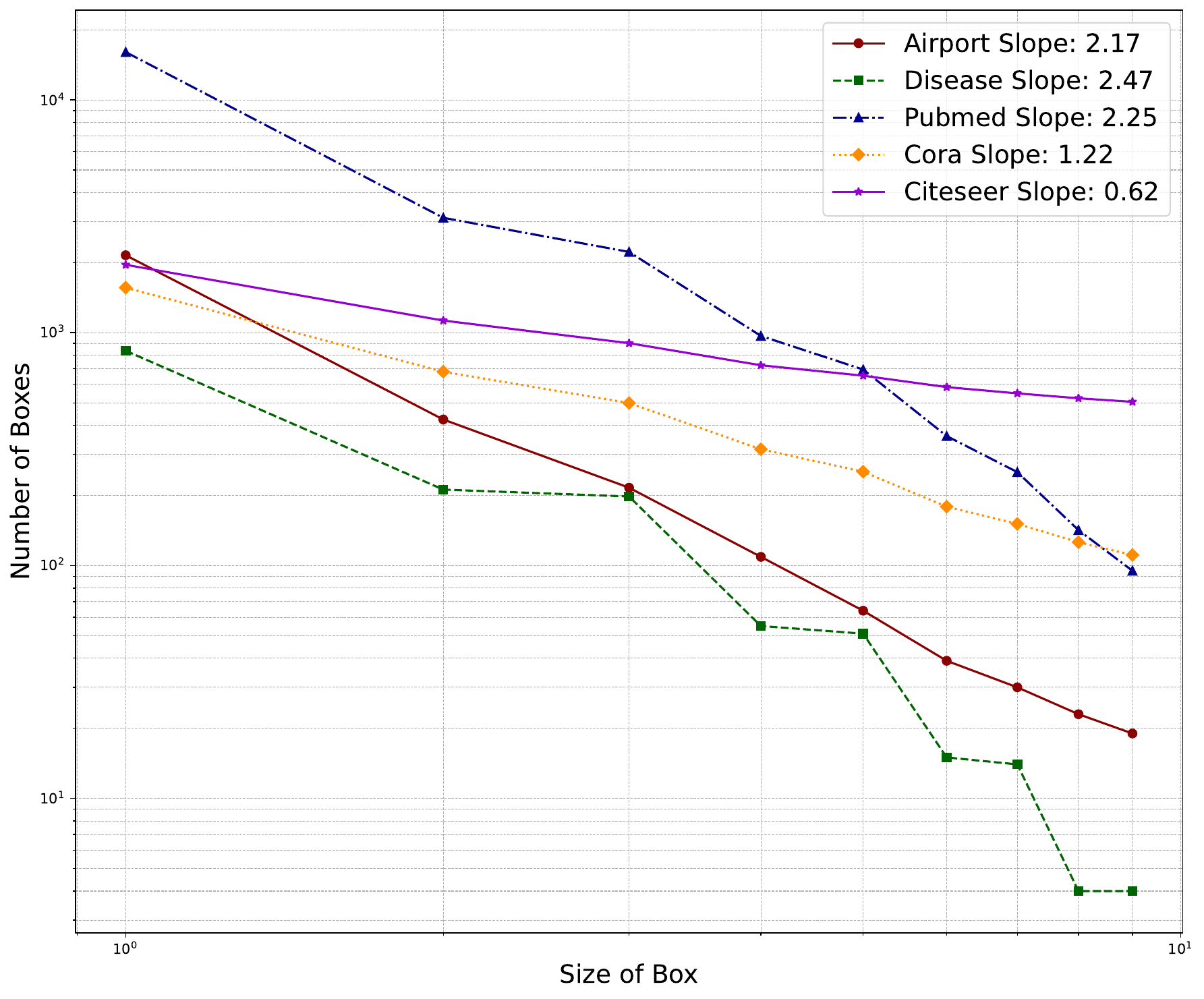}
    \caption{\small The fractal dim of datasets. We use the Compact-Box-Burning algorithm in \citep{song2007calculate} to compute the log-log slope (fractal dim) of the box size and the minimum number of boxes needed to cover the graph.}
    \label{fig:fracdim} 
\end{figure}

In \cref{fig:fracdim}, using the Compact-Box-Burning algorithm from \citep{song2007calculate}, we compute the fractal dimension for some datasets that have moderate sizes. As noted in \cref{tab:noderesults}, there is a clear trend between $\delta$-hyperbolicity (as referenced in \citep{chami2019hyperbolic_GCNN} for assessing tree-like structures—with lower values suggesting more tree-like graphs) and the fractal dimension of datasets. Specifically, a lower $\delta$-hyperbolicity corresponds to a larger fractal dimension. As discussed in \cref{sec.intro,sec.exp}, we believe that our fractional derivative $D_t^\beta$ effectively captures the fractal geometry in the datasets. 
Notably, we discerned a trend: a larger fractal dimension typically corresponds to a smaller optimal $\beta$. 

\section{More Dynamics in FROND Framework}\label{sec.app_moredynamic}

\subsection{Review of Graph ODE Models} \label{ssec.app_odereview}
\tb{GRAND++:} The work by \citep{thorpe2022grand++} introduces graph neural diffusion with a source term, aimed at graph learning in scenarios with a limited quantity of labeled nodes. This approach leverages a subset of feature vectors, those associated with labeled nodes, indexed by $\calI$, and considered ``trustworthy'' to act as a source term. It adheres to \cref{eq.grand_main,eq.gra_dif_L}, incorporating an additional source term, facilitating the propagation of information from nodes in $\calI$ to node $i$.
\begin{align}
    \frac{\ud \bX(t)}{\ud  t} =F(\bX(t))+s(\{\bx_{i}\}_{i\in\calI})
\end{align}
Here, $\calI$ denotes the set of source nodes, $s(\cdot)$ represents a source function, and $F(\cdot)$ embodies the function depicting the right-hand side of \cref{eq.grand_main,eq.gra_dif_L}. The model is manifested in two variations, respectively denoted as GRAND++-nl and GRAND++-l. 

\tb{GraphCON:} Inspired by oscillator dynamical systems, GraphCON \citep{rusch2022graph} is defined through the employment of second-order ODEs. It is crucial to highlight that, for computation, the second-order ODE is decomposed into two first-order ODEs: 
\begin{align}
\begin{aligned}
& \frac{{\ud} \mathbf{Y}(t)}{{\ud} t} = \sigma ({\mathbf{F}_{\theta}(\mathbf{X}(t), t)) - \gamma\mathbf{X}(t) -\tilde{\alpha}\mathbf{Y}(t)}, \
& \frac{{\ud} \mathbf{X}(t)}{{\ud} t} = \mathbf{Y}(t), \label{eq.graphcon_diff}
\end{aligned}
\end{align}
where $\sigma(\cdot)$ is the activation function, $\mathbf{F}_{\theta}(\mathbf{X}(t), t)$ is the neural network function with parameters $\theta$, $\gamma$ and $\tilde{\alpha}$ are learnable coefficients, and $\mathbf{Y}(t)$ is the velocity term converting the second-order ODE to two first-order ODEs.\\
Analogous to the GRAND model, the GraphCON model is also available in both linear (GraphCON-l) and non-linear (GraphCON-nl) versions concerning time. The differentiation between these versions is determined by whether the function $\mathbf{F}_{\theta}$ undergoes updates based on time $t$.

\tb{CDE:} With the objective of addressing heterophilic graphs, the paper \citep{ZhaKanSon:C23} integrates the concept of convection-diffusion equations (CDE) into GNNs, leading to the proposition of the neural CDE model: This innovative model incorporates a convection term and introduces a unique velocity for each node, aiming to preserve diversity in heterophilic graphs. The corresponding formula is illustrated in \cref{eq.cde_diff}.
\begin{align}
\frac{\ud \bX(t)}{\ud t} =(\mathbf{A}(\mathbf{X}(t))-\mathbf{I}) \mathbf{X}(t) + \operatorname{div}(\bV(t) \circ \bX(t))
\label{eq.cde_diff}
\end{align}
In this equation, $\bV(t)$ represents the velocity field of the graph at time $t$, $\operatorname{div}(\cdot)$ denotes the divergence operator as defined in the paper \citep{chamberlain2021grand,SonKanWan:C22}, and $\circ$ symbolizes the element-wise (Hadamard) product. \\
\tb{GREAD:}
To address the challenges posed by heterophilic graphs, the authors in \citep{choi2023gread} present the GREAD model. This model enhances the GRAND model by incorporating a reaction term, thereby formulating a diffusion-reaction equation within GNNs. The respective formula is depicted in \cref{eq.gread_diff}, and the paper offers various alternatives for the reaction term.
\begin{align}
\frac{\ud \bX(t)}{\ud t} = -\alpha \bL(\bX(t)) + \alpha r(\bX(t))
\label{eq.gread_diff}
\end{align}
In this equation, $r(\bX(t))$ represents the reaction term, and $\alpha$ is a trainable parameter used to balance the impact of each term.

\subsection{F-GRAND++}
Building upon the GRAND++ model \citep{thorpe2022grand++}, we define F-GRAND++ as follows:
\begin{align}
    D_t^\beta \bX(t) =F(\bX(t))+s(\{\bx_{i}\}_{i\in\calI})\label{eq.rl_grand++}
\end{align}
We follow the same experimental settings as delineated in the GRAND++ paper. Given that the primary focus of GRAND++ is the model’s performance under limited-label scenarios, our experiments also align with this setting. The sole distinction lies in the incorporation of fractional dynamics.
Within this framework, we substitute the ordinary differential equation $\frac{\ud \bX(t)}{\ud t}$ used in GRAND++ with our FROND fractional derivative $D_t^\beta \bX(t)$. The optimal $\beta$ is determined through hyperparameter tuning. When $\beta=1$, F-GRAND++ seamlessly reverts to GRAND++, and the results from the original paper are reported.
Our observations distinctly indicate that the Fractional-GRAND++ consistently surpasses the performance of the original GRAND++ in nearly all scenarios. We also present the complete comparison results in \cref{tab:res_fgrand++_full}, where it is evident that F-GRAND++ demonstrates greater effectiveness in learning with low labeling rates compared to GRAND++, GRAND, and other baseline methods.

\begin{table}[!htb]    
\caption{Node classification results (\%)  under limited-label scenarios}

    \label{tab:res_fgrand++}
    \small
    \centering
    \resizebox{1.0\textwidth}{!}{
    \begin{tabular}{c|ccccccc}
    \toprule
        Model & pre class & Cora & Citeseer & Pubmed & CoauthorCS & Computer & Photo \\

        \midrule
        GRAND++   & 1 &  54.94$\pm$16.09 &  58.95$\pm$9.59 &   65.94$\pm$4.87  & 60.30$\pm$1.50 &  67.65$\pm$0.37    &    83.12$\pm$0.78    \\
    F-GRAND++   & 1 &  \tb{57.31$\pm$8.89} &  \tb{59.11$\pm$6.73} &   \tb{65.98$\pm$2.72}  & \tb{67.71$\pm$1.91} &  67.65$\pm$0.37 &  83.12$\pm$0.78   \\
    & $\beta$ & 0.95  & 0.95 & 0.85 & 0.7 & 1.0 & 1.0 \\

    \midrule
        GRAND++   & 2 &  66.92$\pm$10.04 &  64.98$\pm$8.31 &  69.31$\pm$4.87  & 76.53$\pm$1.85 &  74.47$\pm$1.48   &  83.71$\pm$0.90    \\
    F-GRAND++    & 2 &  \tb{70.09$\pm$8.36} &  64.98$\pm$8.31 &  \tb{69.37$\pm$5.36}  & \tb{77.97$\pm$2.35}  &  \tb{78.85$\pm$0.96}  &  83.71$\pm$0.90    \\

    & $\beta$ & 0.9 & 1.0 & 0.95  & 0.5  & 0.8  &  1.0 \\

    \midrule
        GRAND++ &  5&  77.80$\pm$4.46 &  70.03$\pm$3.63 &   71.99$\pm$1.91  & 84.83$\pm$0.84 &  82.64$\pm$0.56    &  88.33$\pm$1.21    \\
    F-GRAND++   & 5 &  \tb{78.79$\pm$1.66} &  \tb{70.26$\pm$2.36} &   \tb{73.38$\pm$5.67}  &   \tb{86.09$\pm$2.09}     &  82.64$\pm$0.56   &   \tb{88.56$\pm$0.67}    \\
    & $\beta$ &0.9  & 0.8 & 0.9 & 0.8 &  1.0 & 0.75 \\
    \midrule

    GRAND++   & 10 &  80.86$\pm$2.99 &  72.34$\pm$2.42 &   75.13$\pm$3.88  &  86.94$\pm$0.46 &  82.99$\pm$0.81  &  90.65$\pm$1.19    \\
    F-GRAND++  & 10 &  \tb{82.73$\pm$0.81} &  \tb{73.52$\pm$1.44} &   \tb{77.15$\pm$2.87}   & \tb{87.85$\pm$1.44}  &  \tb{83.26$\pm$0.41}  &  \tb{91.15$\pm$0.52}     \\
    & $\beta$ & 0.95 & 0.9 & 0.95 &0.6 & 0.7 & 0.95 \\

    \midrule

     GRAND++ &  20 &  82.95$\pm$1.37 &  73.53$\pm$3.31  &   79.16$\pm$1.37  & 90.80$\pm$0.34 &  85.73$\pm$0.50  &  93.55$\pm$0.38   \\
    F-GRAND++  & 20&  \tb{84.57$\pm$1.07} &  \tb{74.81$\pm$1.78}  &   \tb{79.96$\pm$1.68}  & \tb{91.03$\pm$0.72} &  \tb{85.78$\pm$0.43}     &   93.55$\pm$0.38    \\
    & $\beta$ & 0.9 & 0.85 & 0.95 & 0.9 & 0.9 & 1.0\\

    \bottomrule
    \end{tabular}
    }

\end{table}

\begin{table}[!htb]    
\caption{Full table: Classification accuracy of different GNNs trained with different number of labeled data per class (\#per class) on six benchmark graph node classification tasks. The highest accuracy is highlighted in bold for each number of labeled data per class. These results show that F-GRAND++ is more effective in learning with low-labeling rates than GRAND++ and GRAND. Where available, baseline results are cited from \citep{thorpe2022grand++}.}
 \label{tab:res_fgrand++_full}
    \small
    \centering
    \resizebox{1.0\textwidth}{!}{
\begin{tabular}{|c|c|c|c|c|c|c|c|}
\hline Model & \#per class & CORA & CiteSeer & PubMed & CoauthorCS & Computer & Photo \\
\hline \begin{tabular}{c} 
F-GRAND++ \\
\end{tabular} & \begin{tabular}{c}
1 \\
2 \\
5 \\
10 \\
20
\end{tabular} & \begin{tabular}{c}
\tb{57.31 $\pm$ 8.89} \\
\tb{70.09 $\pm$ 8.36} \\
\tb{78.79 $\pm$ 1.66} \\
\tb{82.73 $\pm$ 0.81} \\
\tb{84.57 $\pm$ 1.07}
\end{tabular} & \begin{tabular}{c}
\tb{59.11 $\pm$ 6.73}\\
 \tb{64.98 $\pm$ 8.31} \\
\tb{70.26 $\pm$ 2.36} \\
\tb{73.52 $\pm$ 1.44 }\\
\tb{74.81 $\pm$ 1.78 }\\
\end{tabular} & \begin{tabular}{c}
\tb{65.98 $\pm$ 2.72}\\
\tb{69.37 $\pm$ 5.36} \\
73.38 $\pm$ 5.67\\
\tb{77.15 $\pm$ 2.87} \\
\tb{79.96 $\pm$ 1.68}
\end{tabular} & \begin{tabular}{l}
\tb{67.71 $\pm$ 1.91} \\
77.97 $\pm$ 2.35 \\
{86.09 $\pm$ 2.09}\\
{87.85 $\pm$ 1.44} \\
{91.03 $\pm$ 0.72} 
\end{tabular} & \begin{tabular}{l}
\tb{67.65 $\pm$ 0.37}\\
\tb{78.85 $\pm$ 0.96}\\
\tb{82.64 $\pm$ 0.56}\\
\tb{83.26 $\pm$ 0.41}\\
\tb{85.78 $\pm$ 0.43}
\end{tabular} & \begin{tabular}{l}
\tb{83.12 $\pm$ 0.78} \\
\tb{83.71 $\pm$ 0.90} \\
88.56 $\pm$ 0.67\\
\tb{91.15 $\pm$ 0.52}\\
\tb{93.55 $\pm$ 0.38}
\end{tabular} \\
\hline \begin{tabular}{c} 
GRAND++ \\
\end{tabular} & \begin{tabular}{c}
1 \\
2 \\
5 \\
10 \\
20
\end{tabular} & \begin{tabular}{c}
54.94 $\pm$ 16.09\\
66.92 $\pm$ 10.04\\
77.80 $\pm$ 4.46\\
80.86 $\pm$ 2.99\\
82.95 $\pm$ 1.37\\
\end{tabular} & \begin{tabular}{l}
58.95 $\pm$ 9.59 \\ 
\tb{64.98 $\pm$ 8.31} \\ 
70.03 $\pm$ 3.63 \\ 
72.34 $\pm$ 2.42 \\ 
73.53 $\pm$ 3.31
\end{tabular} & \begin{tabular}{l}
65.94 $\pm$ 4.87 \\
69.31 $\pm$ 4.87 \\
71.99 $\pm$ 1.91 \\
75.13 $\pm$ 3.88 \\
79.16 $\pm$ 1.37
\end{tabular} & \begin{tabular}{l}
60.30 $\pm$ 1.50 \\
76.53 $\pm$ 1.85 \\
84.83 $\pm$ 0.84 \\
86.94 $\pm$ 0.46 \\
90.80 $\pm$ 0.34
\end{tabular} & \begin{tabular}{l}
\tb{67.65 $\pm$ 0.37} \\
76.47 $\pm$ 1.48 \\
82.64 $\pm$ 0.56 \\
82.99 $\pm$ 0.81 \\
85.73 $\pm$ 0.50
\end{tabular} & \begin{tabular}{l}
\tb{83.12 $\pm$ 0.78} \\
\tb{83.71 $\pm$ 0.90}\\
88.33 $\pm$ 1.21 \\
90.65 $\pm$ 1.19 \\
\tb{93.55 $\pm$ 0.38}
\end{tabular} \\
\hline \begin{tabular}{c} 
GRAND \\
\end{tabular} & \begin{tabular}{c}
1 \\
2 \\
5 \\
10 \\
20
\end{tabular} & \begin{tabular}{c}
52.53 $\pm$ 16.40 \\
64.82 $\pm$ 11.16 \\
76.07 $\pm$ 5.08 \\
80.25 $\pm$ 3.40 \\
82.86 $\pm$ 2.39
\end{tabular} & \begin{tabular}{c}
50.06 $\pm$ 17.98 \\
59.55 $\pm$ 10.89 \\
68.37 $\pm$ 5.00 \\
71.90 $\pm$ 7.66 \\
73.02 $\pm$ 5.89
\end{tabular} & \begin{tabular}{c}
62.11 $\pm$ 10.58 \\
69.00 $\pm$ 7.55 \\
\tb{73.98 $\pm$ 5.08} \\
76.33 $\pm$ 3.41 \\
78.76 $\pm$ 1.69
\end{tabular} & \begin{tabular}{l}
59.15 $\pm$ 5.73 \\
73.83 $\pm$ 5.58 \\
85.29 $\pm$ 2.19 \\
87.81 $\pm$ 1.36 \\
91.03 $\pm$ 0.47
\end{tabular} & \begin{tabular}{l}
48.67 $\pm$ 1.66 \\
74.77 $\pm$ 1.85 \\
80.72 $\pm$ 1.09 \\
82.42 $\pm$ 1.10 \\
84.54 $\pm$ 0.90
\end{tabular} & \begin{tabular}{l}
81.25 $\pm$ 2.50 \\
82.13 $\pm$ 3.27 \\
88.27 $\pm$ 1.94 \\
90.98 $\pm$ 0.93 \\
93.53 $\pm$ 0.47
\end{tabular} \\
\hline GCN & \begin{tabular}{c}
1 \\
2 \\
5 \\
10 \\
20
\end{tabular} & \begin{tabular}{c}
47.72 $\pm$ 15.33 \\
60.85 $\pm$ 14.01 \\
73.86 $\pm$ 7.97 \\
78.82 $\pm$ 5.38 \\
82.07 $\pm$ 2.03
\end{tabular} & \begin{tabular}{c}
48.94 $\pm$ 10.24 \\
58.06 $\pm$ 9.76 \\
67.24 $\pm$ 4.19 \\
72.18 $\pm$ 3.47 \\
74.21 $\pm$ 2.90
\end{tabular} & \begin{tabular}{c}
58.61 $\pm$ 12.83 \\
60.45 $\pm$ 16.20 \\
68.69 $\pm$ 7.93 \\
72.59 $\pm$ 3.19 \\
76.89 $\pm$ 3.27
\end{tabular} & \begin{tabular}{l}
65.22 $\pm$ 2.25 \\
\tb{83.61 $\pm$ 1.49} \\
86.66 $\pm$ 0.43 \\
88.60 $\pm$ 0.50 \\
91.09 $\pm$ 0.35
\end{tabular} & \begin{tabular}{l}
49.46 $\pm$ 1.65 \\
76.90 $\pm$ 1.49 \\
82.47 $\pm$ 0.97 \\
82.53 $\pm$ 0.74 \\
82.94 $\pm$ 1.54
\end{tabular} & \begin{tabular}{l}
82.94 $\pm$ 2.17 \\
83.61 $\pm$ 0.71 \\
\tb{88.86 $\pm$ 1.56} \\
90.41 $\pm$ 0.35 \\
91.95 $\pm$ 0.11
\end{tabular} \\
\hline GAT & \begin{tabular}{c}
1 \\
2 \\
5 \\
10 \\
20
\end{tabular} & \begin{tabular}{c}
47.86 $\pm$ 15.38 \\
58.30 $\pm$ 13.55 \\
71.04 $\pm$ 5.74 \\
76.31 $\pm$ 4.87 \\
79.92 $\pm$ 2.28
\end{tabular} & \begin{tabular}{c}
50.31 $\pm$ 14.27 \\
55.55 $\pm$ 9.19 \\
67.37 $\pm$ 5.08 \\
71.35 $\pm$ 4.92 \\
73.22 $\pm$ 2.90
\end{tabular} & \begin{tabular}{c}
58.84 $\pm$ 12.81 \\
60.24 $\pm$ 14.44 \\
68.54 $\pm$ 5.75 \\
72.44 $\pm$ 3.50 \\
75.55 $\pm$ 4.11
\end{tabular} & \begin{tabular}{l}
51.13 $\pm$ 5.24 \\
63.12 $\pm$ 6.09 \\
71.65 $\pm$ 4.53 \\
74.71 $\pm$ 3.35 \\
79.95 $\pm$ 2.88
\end{tabular} & \begin{tabular}{l}
37.14 $\pm$ 7.81 \\
65.07 $\pm$ 8.86 \\
71.43 $\pm$ 7.34 \\
76.04 $\pm$ 0.35 \\
80.05 $\pm$ 1.81
\end{tabular} & \begin{tabular}{l}
73.58 $\pm$ 8.15 \\
76.89 $\pm$ 4.89 \\
83.01 $\pm$ 3.64 \\
87.42 $\pm$ 2.38 \\
89.38 $\pm$ 2.48
\end{tabular} \\
\hline \begin{tabular}{c} 
GraphSage \\
\end{tabular} & \begin{tabular}{c}
1 \\
2 \\
5 \\
10 \\
20
\end{tabular} & \begin{tabular}{c}
43.04 $\pm$ 14.01 \\
53.96 $\pm$ 12.18 \\
68.14 $\pm$ 6.95 \\
75.04 $\pm$ 5.03 \\
80.04 $\pm$ 2.54
\end{tabular} & \begin{tabular}{c}
48.81 $\pm$ 11.45 \\
54.39 $\pm$ 11.37 \\
64.79 $\pm$ 5.16 \\
68.90 $\pm$ 5.08 \\
72.02 $\pm$ 2.82
\end{tabular} & \begin{tabular}{c}
55.53 $\pm$ 12.71 \\
58.97 $\pm$ 12.65 \\
66.07 $\pm$ 6.16 \\
70.74 $\pm$ 3.11 \\
74.55 $\pm$ 3.09
\end{tabular} & \begin{tabular}{l}
61.35 $\pm$ 1.35 \\
76.51 $\pm$ 1.31 \\
\tb{89.06 $\pm$ 0.69} \\
\tb{89.68 $\pm$ 0.39} \\
\tb{91.33 $\pm$ 0.36}
\end{tabular} & \begin{tabular}{l}
27.65 $\pm$ 2.39 \\
42.63 $\pm$ 4.29 \\
64.83 $\pm$ 1.62 \\
74.66 $\pm$ 1.29 \\
79.98 $\pm$ 0.96
\end{tabular} & \begin{tabular}{l}
45.36 $\pm$ 7.13 \\
51.93 $\pm$ 4.21 \\
78.26 $\pm$ 1.93 \\
84.38 $\pm$ 1.75 \\
91.29 $\pm$ 0.67
\end{tabular} \\
\hline \begin{tabular}{c} 
MoNet \citep{monti2017geometric} 
\end{tabular} & \begin{tabular}{c}
1 \\
2 \\
5 \\
10 \\
20
\end{tabular} & \begin{tabular}{c}
47.72 $\pm$ 15.53 \\
60.85 $\pm$ 14.01 \\
73.86 $\pm$ 7.97 \\
78.82 $\pm$ 5.38 \\
82.07 $\pm$ 2.03
\end{tabular} & \begin{tabular}{c}
39.13 $\pm$ 11.37 \\
48.52 $\pm$ 9.52 \\
61.66 $\pm$ 6.61 \\
68.08 $\pm$ 6.29 \\
71.52 $\pm$ 4.11
\end{tabular} & \begin{tabular}{l}
56.47 $\pm$ 4.67 \\
61.03 $\pm$ 6.93 \\
67.92 $\pm$ 2.50 \\
71.24 $\pm$ 1.54 \\
76.49 $\pm$ 1.75
\end{tabular} & \begin{tabular}{l}
58.99 $\pm$ 5.17 \\
76.57 $\pm$ 4.06 \\
87.02 $\pm$ 1.67 \\
88.76 $\pm$ 0.49 \\
90.31 $\pm$ 0.41
\end{tabular} & \begin{tabular}{l}
23.78 $\pm$ 7.57 \\
38.19 $\pm$ 3.72 \\
59.38 $\pm$ 4.73 \\
68.66 $\pm$ 3.30 \\
73.66 $\pm$ 2.87
\end{tabular} & \begin{tabular}{l}
34.72 $\pm$ 8.18 \\
43.03 $\pm$ 8.22 \\
71.80 $\pm$ 5.02 \\
78.66 $\pm$ 3.17 \\
88.61 $\pm$ 1.18
\end{tabular} \\
\hline
\end{tabular}
}
\end{table}

\subsection{F-CDE} \label{subsec.supp_f_cde_formular}
Drawing inspiration from the graph neural CDE model  \citep{ZhaKanSon:C23}, we further define the F-CDE model as follows:
\begin{align}
D_t^\beta \bX(t) =(\mathbf{A}(\mathbf{X}(t))-\mathbf{I}) \mathbf{X}(t) + \operatorname{div}(\bV(t) \circ \bX(t))
\label{eq.rl_cde}
\end{align}
In this expression, $\bV(t)$ represents the velocity field of the graph at time $t$. The divergence operator, $\operatorname{div}(\cdot)$, is defined as per the formulation given in \citep{SonKanWan:C22}, and $\circ$ symbolizes the element-wise (Hadamard) product.

We follow the same experimental setting as in the CDE paper\citep{ZhaKanSon:C23}. Given that the primary focus of CDE is on evaluating model performance on large heterophilic datasets, our experiments are also conducted under similar conditions. The statistics for the dataset are available in \cref{tab:data_hetero}.
The sole distinction in our approach lies in incorporating fractional dynamics; we achieve this by replacing the ODE used in CDE with our FROND fractional derivative. The complete comparison results in \cref{tab:res_cde_hetero_full} conspicuously reveal that Fractional CDE exhibits superior performance compared to the conventional CDE and other baselines across various datasets.

\begin{table*}[!ht]
\centering
\caption{Dataset statistics used in \cref{tab:res_cde_hetero}}
\begin{tabular}{ccccccc}

\toprule
 Dataset & Nodes & Edges & Classes & Node Features   \\
 \midrule
 Roman-empire &  22662     &  32927     &      18   &  300      \\
 Wiki-cooc &  10000     &  2243042     &      5   &  100     \\
 Minesweeper &  10000     &  39402     &     2   &  7     \\
 Questions &  48921     &  153540     &     2   &  301    \\
 Workers &  11758     &  519000     &     2   &  10     \\
 Amaon-ratings &  24492     &  93050     &     5   &  300      \\

 \bottomrule
\end{tabular}
\label{tab:data_hetero}
\end{table*}

\begin{table}[H]
    \caption{Full table: Node classification accuracy(\%) of large heterophilic datasets.}
    \label{tab:res_cde_hetero_full}
    \small
    \centering
    \resizebox{0.9\textwidth}{!}{
    \begin{tabular}{cccccccc}
    \toprule
        Model & Roman-empire & Wiki-cooc & Minesweeper & Questions & Workers & Amazon-ratings \\

    \midrule
    ResNet  &  65.71$\pm$0.44     &  89.36$\pm$0.71     &  50.95$\pm$1.12   & 70.10$\pm$0.75      &  73.08$\pm$1.28     &  45.70$\pm$0.69 \\ 
    \midrule
    H2GCN\citep{zhu2020h2gcn}   &  68.09$\pm$0.29       & 89.24$\pm$0.32    &  89.95$\pm$0.38     &66.66$\pm$1.84           & 81.76$\pm$0.68     & 41.36$\pm$0.47  \\

    CPGNN\citep{zhu2021cpgnn}  &  63.78$\pm$0.50       & 84.84$\pm$0.66    &  71.27$\pm$1.14     & 67.09$\pm$2.63          & 72.44$\pm$0.80     & 44.36$\pm$0.35 \\
    GPR-GNN\citep{chien2020gprgnn}  &  73.37$\pm$0.68      & 91.90$\pm$0.78     &  81.79$\pm$0.98    & 73.41$\pm$1.24         &  70.59$\pm$1.15    &  43.90$\pm$0.48 \\
    GloGNN\citep{li2022Glognn}  &  63.85$\pm$0.49       &  88.49$\pm$0.45     &  62.53$\pm$1.34   &  67.15$\pm$1.92        &  73.90$\pm$0.95    &  37.28$\pm$0.66  \\
    FAGCN\citep{bo2021fagcn}  &  70.53$\pm$0.99          &  91.88$\pm$0.37   &  89.69$\pm$0.60     & \tb{77.04$\pm$1.56}    &  81.87$\pm$0.94    &  46.32$\pm$2.50 \\
    GBK-GNN\citep{du2022gbkgnn}   &  75.87$\pm$0.43       &  97.81$\pm$0.32   &  83.56$\pm$0.84    & 72.98$\pm$1.05        &  78.06$\pm$0.91   &  43.47$\pm$0.51  \\
    ACM-GCN\citep{luan2022acmgcn}   &  68.35$\pm$1.95      &  87.48$\pm$1.06    &  90.47$\pm$0.57     & OOM                  &  78.25$\pm$0.78   &  38.51$\pm$3.38   \\
    \midrule
    GRAND\citep{chamrowgor:grand2021}  &  71.60$\pm$0.58         & 92.03$\pm$0.46     & 76.67$\pm$0.98        & 70.67$\pm$1.28   &  75.33$\pm$0.84    &  45.05$\pm$0.65  \\
    GraphBel\citep{SonKanWan:C22} &  69.47$\pm$0.37     &  90.30$\pm$0.50     &  76.51$\pm$1.03     & 70.79$\pm$0.99         &  73.02$\pm$0.92  &43.63$\pm$0.42  \\
    
    Diag-NSD\citep{crifraben:sheaf2022}  &  77.50$\pm$0.67     &  92.06$\pm$0.40    &  89.59$\pm$0.61     & 69.25$\pm$1.15         &  79.81$\pm$0.99   &  37.96$\pm$0.20 \\
    ACMP\citep{wang2022acmp} &  71.27$\pm$0.59 & 92.68$\pm$0.37  & 76.15$\pm$1.12  & 71.18$\pm$1.03 & 75.03$\pm$0.92  &44.76$\pm$0.52 \\
    \midrule
        CDE &   91.64$\pm$0.28 &  97.99$\pm$0.38  &   95.50$\pm$5.23  &  75.17$\pm$0.99 &  80.70$\pm$1.04 &  47.63$\pm$0.43    \\
    F-CDE  &   \tb{93.06$\pm$0.55} &  \tb{98.73$\pm$0.68}  &   \tb{96.04$\pm$0.25}  & 75.17$\pm$0.99 &  \tb{82.68$\pm$0.86} &  \tb{49.01$\pm$0.56}    \\

    \gray{$\beta$ for F-CDE} &  \gray{0.9}& \gray{0.6 } &\gray{0.6} & \gray{1.0 }& \gray{0.4}&\gray{0.1 } \\

    \bottomrule
    \end{tabular}

    }
\end{table}

\subsection{F-GREAD}
Our FROND framework is also extendable to the GREAD model \citep{choi2023gread}, as defined in \cref{eq.rl_gread}.
\begin{align}
D_t^\beta \bX(t) = -\alpha \bL(\bX(t)) + \alpha r(\bX(t))
\label{eq.rl_gread}
\end{align}
where $r(\bX(t))$ represents a reaction term, and $\alpha$ is a trainable parameter used to emphasize each term. 

We adhere to the same experimental setting outlined in the GREAD paper \citep{choi2023gread}, concentrating exclusively on heterophilic datasets. We choose the Blurring-Sharpening (BS) as the reaction term to formulate both GREAD-BS and F-GREAD-BS, as GREAD-BS exhibits strong performance according to Table 4 in the GREAD paper \citep{choi2023gread}. The results presented in \cref{tab:res_f_gread} (refer to \cref{tab:res_f_gread_full} for comprehensive comparisons with other baselines) demonstrate that our FROND framework enhances the performance of GREAD across all examined datasets.

\begin{table}[!htb]
   \caption{Node classification accuracy(\%) of heterophilic datasets}
    \label{tab:res_f_gread}
    \small
    \centering
    \begin{tabular}{cccccccc}
    \toprule
        Model & Chameleon & Squirrel & Film &  Texas & Wisconsin \\

    \midrule
        GREAD-BS &   71.38$\pm$1.31 &  59.22$\pm$1.44  &   37.90$\pm$1.17  &  88.92$\pm$3.72 &  89.41$\pm$3.30     \\
    F-GREAD-BS  &   \tb{71.45$\pm$1.98} &  \tb{60.86$\pm$1.05}  &   \tb{38.28$\pm$0.74}  &  \tb{92.97$\pm$4.39} &  \tb{90.59$\pm$3.80}     \\
    $\beta$  & 0.9 & 0.9 & 0.8& 0.9 & 0.9 \\
    \bottomrule
    \end{tabular}
 
\end{table}

\begin{table}[!htb]
   \caption{Full table: Node classification accuracy(\%) of heterophilic datasets}
    \label{tab:res_f_gread_full}
    \small
    \centering
    \begin{tabular}{cccccccc} 
    \toprule
        Model & Chameleon & Squirrel & Film &  Texas & Wisconsin \\
    \midrule
    Geom-GCN\citep{pei2020geomgcn} &   60.00$\pm$2.81 &  38.15$\pm$0.92  &   31.59$\pm$1.15  &  66.76$\pm$2.72 &  64.51$\pm$3.66     \\
    H2GCN\citep{zhu2020h2gcn} & 60.11$\pm$2.15 &  36.48$\pm$1.86  &   35.70$\pm$1.00  &  84.86$\pm$7.23 &  87.65$\pm$4.98     \\
    GGCN\citep{yan2022ggcn} & 71.14$\pm$1.84 &  55.17$\pm$1.58  &   37.54$\pm$1.56  &  84.86$\pm$4.55 &  86.86$\pm$3.29     \\
    LINKX\citep{lim2021LINKX} &  68.42$\pm$1.38 &  \tb{61.81$\pm$1.80}  &   36.10$\pm$1.55  &  74.60$\pm$8.37 &  75.49$\pm$5.72     \\
    GloGNN\citep{li2022Glognn} & 69.78$\pm$2.42 &  57.54$\pm$1.39  &   37.35$\pm$1.30  &  84.32$\pm$4.15 &  87.06$\pm$3.53     \\
    ACM-GCN\citep{luan2022acmgcn} & 66.93$\pm$1.85 &  54.40$\pm$1.88  &   36.28$\pm$1.09  &  87.84$\pm$4.40 &  88.43$\pm$3.22     \\
    GCNII\citep{chen2020simple} & 63.86$\pm$3.04 &  38.47$\pm$1.58  &   37.44$\pm$1.30  &  77.57$\pm$3.83 &  80.39$\pm$3.40     \\

    \midrule
    
    CGNN\citep{xhonneux2020continuous} &  46.89$\pm$1.66 &  29.24$\pm$1.09  &   35.95$\pm$0.86  &  71.35$\pm$4.05 &  74.31$\pm$7.26     \\

    GRAND\citep{chamrowgor:grand2021} &  54.67$\pm$2.54 &  40.05$\pm$1.50  &   35.62$\pm$1.01  &  75.68$\pm$7.25 &  79.41$\pm$3.64     \\

    BLEND\citep{charoweyn:blend2021} & 60.11$\pm$2.09 &  43.06$\pm$1.39  &   35.63$\pm$1.01  &  83.24$\pm$4.65 &  84.12$\pm$3.56     \\
    Sheaf\citep{crifraben:sheaf2022} & 68.04$\pm$1.58 &  56.34$\pm$1.32  &   37.81$\pm$1.15  &  85.05$\pm$5.51 &  89.41$\pm$4.74     \\
    GRAFF\citep{di2022graff} & 71.08$\pm$1.75 &  54.52$\pm$1.37  &   36.09$\pm$0.81  &  88.38$\pm$4.53 &  87.45$\pm$2.94     \\

    \midrule
        GREAD-BS &   71.38$\pm$1.31 &  59.22$\pm$1.44  &   37.90$\pm$1.17  &  88.92$\pm$3.72 &  89.41$\pm$3.30     \\
    F-GREAD-BS  &   \tb{71.45$\pm$1.98} &  60.86$\pm$1.05  &   \tb{38.28$\pm$0.74}  &  \tb{92.97$\pm$4.39} &  \tb{90.59$\pm$3.80}     \\
    $\beta$  & 0.9 & 0.9 & 0.8& 0.9 & 0.9 \\
    \bottomrule
    \end{tabular}
\end{table}

\subsection{F-GraphCON}
We also incorporate the following fractional-order oscillators dynamics, inspired by \citep{radwan2008fractional,rusch2022graph}:
\begin{align}
    \label{eq.graphcon2}
    \begin{aligned} 
    & D_t^\beta \mathbf{Y}=\sigma\left(\mathbf{F}_\theta(\mathbf{X}, t)\right)-\gamma \mathbf{X}-\alpha \mathbf{Y} \\ 
    & D_t^\beta \mathbf{X}=\mathbf{Y}\end{aligned}
\end{align}
which represent the fractional dynamics version of GraphCON \citep{rusch2022graph}. We denote this as F-GraphCON, with two variants, F-GraphCON-GCN and F-GraphCON-GAT. Here, $\mathbf{F}_\theta$ is set as GCN and GAT, as in the setting described in \citep{rusch2022graph}. We refer readers to \citep{rusch2022graph} for further details. Notably, when $\beta=1$, F-GraphCON simplifies to GraphCON, devoid of memory functionality.

\begin{table}[!htb]
    \centering
    \caption{Node classification accuracy(\%) based on GraphCON model}
    \begin{tabular}{c|ccccc}
    \toprule
         & Cora & Citeseer & Pubmed  & Airport & Disease  \\
        GraphCON-GCN &  81.9$\pm$1.7  &  72.9$\pm$2.1 &  78.8$\pm$2.6  & 68.6$\pm$2.1 &  87.5$\pm$4.1 \\
        GraphCON-GAT &  83.2$\pm$1.4  &  73.2$\pm$1.8  &  79.4$\pm$1.3  & 74.1$\pm$2.7 &  65.7$\pm$5.9  \\
    \midrule
        F-GraphCON-GCN &  \tb{84.6$\pm$1.4}  &  \tb{75.3$\pm$1.1}
          &  \tb{80.3$\pm$1.3}  & \tb{97.3$\pm$0.5}  & \tb{92.1$\pm$2.8 } \\
        $\beta$   & 0.9  & 0.8  & 0.9 &  0.1 & 0.1\\
        F-GraphCON-GAT &  83.9$\pm$1.2  &   73.4$\pm$1.5  &  79.4$\pm$1.3   & 97.3$\pm$0.8  & 86.9$\pm$4.0  \\
         $\beta$   & 0.7  & 0.9 & 1.0 &  0.1 & 0.1 \\
    \bottomrule
    \end{tabular}
    \label{tab:res_GraphCON}
\end{table}

\begin{table}[!htb]
    \centering
    \caption{Full table: Node classification accuracy(\%) based on GraphCON model.}
    \begin{tabular}{c|ccccc}
    \toprule
         & Cora & Citeseer & Pubmed  & Airport & Disease  \\

    \midrule
    GCN    &  81.5$\pm$1.3        &  71.9$\pm$1.9   &  77.8$\pm$2.9  &  81.6$\pm$0.6  &  69.8$\pm$0.5   \\
    GAT  &  81.8$\pm$1.3       &  71.4$\pm$1.9   &  78.7$\pm$2.3   &  81.6$\pm$0.4  &  70.4$\pm$0.5   \\
    HGCN &  78.7$\pm$1.0 &  65.8$\pm$2.0 &  76.4$\pm$0.8 & 85.4$\pm$0.7  &  89.9$\pm$1.1\\
    GIL  &   82.1$\pm$1.1  &  71.1$\pm$1.2  &   77.8$\pm$0.6 &  91.5$\pm$1.7 & 90.8$\pm$0.5 \\

    \midrule
    GRAND-l    &  83.6$\pm$1.0        &  73.4$\pm$0.5   &  78.8$\pm$1.7 &  80.5$\pm$9.6  &  74.5$\pm$3.4   \\
    GRAND-nl    &  82.3$\pm$1.6        &  70.9$\pm$1.0   &  77.5$\pm$1.8   &  90.9$\pm$1.6  &  81.0$\pm$6.7   \\

        GraphCON-GCN &  81.9$\pm$1.7  &  72.9$\pm$2.1 &  78.8$\pm$2.6  & 68.6$\pm$2.1 &  87.5$\pm$4.1 \\
        GraphCON-GAT &  83.2$\pm$1.4  &  73.2$\pm$1.8  &  79.4$\pm$1.3  & 74.1$\pm$2.7 &  65.7$\pm$5.9  \\
    \midrule
        F-GraphCON-GCN &  \tb{84.6$\pm$1.4}  &  \tb{75.3$\pm$1.1}
          &  \tb{80.3$\pm$1.3}  & \tb{97.3$\pm$0.5}  & \tb{92.1$\pm$2.8 } \\
        $\beta$   & 0.9  & 0.8  & 0.9 &  0.1 & 0.1\\
        F-GraphCON-GAT &  83.9$\pm$1.2  &   73.4$\pm$1.5  &  79.4$\pm$1.3   & 97.3$\pm$0.8  & 86.9$\pm$4.0  \\
         $\beta$   & 0.7  & 0.9 & 1.0 &  0.1 & 0.1 \\
    \bottomrule
    \end{tabular}
    \label{tab:res_GraphCON_full}
\end{table}

\subsection{F-FLODE}

\begin{table}[!htb]
    \centering
  \caption{Node classification accuracy(\%) of undirected graphs based on F-FLODE model  \label{tab:F-FLODE_undirect}}
    \begin{tabular}{c|ccccc}
    \toprule
         & Film & Squirrel & Chameleon    \\
        FLODE &  37.16$\pm$1.42  &  64.23$\pm$1.84 &  73.60$\pm$1.55   \\
     \midrule
       F-FLODE &  \tb{37.95$\pm$1.27}  &  \tb{65.53$\pm$1.83}  &  \tb{74.17$\pm$1.59}    \\
    $\beta$  & 0.8 & 0.9 & 0.9  \\
    \bottomrule
    \end{tabular}
\end{table}

\begin{table}[!htb]
    \centering
  \caption{Node classification accuracy(\%) of directed graphs based on F-FLODE model \label{tab:F-FLODE_direct}}
    \begin{tabular}{c|ccccc}
    \toprule
         & Film & Squirrel & Chameleon    \\
        FLODE &  37.41$\pm$1.06  &  74.03$\pm$1.58 &  77.98$\pm$1.05   \\
     \midrule
       F-FLODE &  \tb{37.97$\pm$1.15}  &  \tb{75.03$\pm$1.42}  &  \tb{78.51$\pm$1.09}    \\
  $\beta$  & 0.9 & 0.9 & 0.9  \\
    \bottomrule
    \end{tabular}
\end{table}

In the work of \citep{maskey2023fractional}, the authors introduce the FLODE model, which incorporates fractional graph shift operators within integer-order continuous GNNs. 
Specifically, instead of utilizing a Laplacian matrix $\mathbf{L}$, they employ the fractional power of $\mathbf{L}$, denoted as $\mathbf{L}^\alpha$ (see \cref{eq.f-svd}).
Our research diverges from this approach, focusing on the incorporation of time-fractional derivative $D_t^\beta$ for updating graph node features in a memory-inclusive dynamical process.
It is pivotal to differentiate the term ``fractional'' as used in our work from that in \citep{maskey2023fractional}, as they signify fundamentally distinct concepts in the literature. Fundamentally, FLODE differs from our work in key aspects:
\begin{itemize}
    \item FLODE employs the fractional (real-valued) power of $\mathbf{L}$, namely $\mathbf{L}^\alpha$. 
The feature evolution model used by FLODE, specifically in its first heat diffusion-type variant, is given by:
\begin{align}\label{FLODE}\tag{FLODE}
\frac{\mathrm{d}\mathbf{X}(t)}{\mathrm{d}t} = -\mathbf{L}^{\alpha}\bX(t)\bPhi.
\end{align}
This is a graph spatial domain rewiring technique, as $\mathbf{L}^\alpha$ introduces dense connections compared to $\mathbf{L}$. As a result, FLODE introduces space-based long-range interactions during the feature updating process.
    
\item In contrast, our FROND model incorporates the time-fractional derivative $D_t^\beta$ to update graph node features in a memory-inclusive dynamical process. In this context, time acts as a continuous counterpart to the layer index, leading to significant dense skip connections between layers due to memory dependence. Thus, FROND induces time/layer-based long-range interactions in the feature update process. Note that FLODE does not utilize time-fractional derivatives.
Our method is not only \emph{compatible with various integer-order continuous GNNs, including FLODE (see \cref{F-FLODE}),} but also extends them to graph FDE models. 

\end{itemize}
We next introduce the F-FLODE model, which utilizes time-fractional derivatives for updating graph node features in FLODE:
\begin{align} \label{F-FLODE}\tag{F-FLODE}
D_t^\beta \mathbf{X}(t)=-\mathbf{L}^\alpha \mathbf{X}(t) \mathbf{\Phi},
\end{align}
where $\mathbf{L}$ denotes the symmetrically normalized adjacency matrix. The $\alpha$-fractional power of the graph Laplacian, $\mathbf{L}^\alpha$, is given by:
\begin{align}
\mathbf{L}^\alpha:=\mathbf{U} \boldsymbol{\Sigma}^\alpha \mathbf{V}^{\mathrm{H}} . \label{eq.f-svd}
\end{align}
In this formulation, $\mathbf{U}$, $\boldsymbol{\Sigma}$, and $\mathbf{V}$ are obtained from the SVD decomposition of $\mathbf{L}=\mathbf{U} \boldsymbol{\Sigma} \mathbf{V}^{\mathrm{H}}$, and $\alpha \in \mathbb{R}$ represents the order. The channel mixing matrix $\mathbf{\Phi}$, a symmetric matrix, follows the setting in \citep{maskey2023fractional}.

Following the experimental setup outlined in \citep{maskey2023fractional}, we present our results in \cref{tab:F-FLODE_undirect,tab:F-FLODE_direct}, demonstrating that our FROND framework enhances the performance of FLODE across all evaluated datasets.  
Note the difference in the equations in \cref{FLODE} and \cref{F-FLODE}, where the two are equivalent only when $\beta=1$.  
This example illustrates that the FROND framework encompasses the FLODE model as a special case when $\beta=1$. Our experimental results indicate that F-FLODE outperforms FLODE with the optimal $\beta\ne 1$ in general.

\section{Proofs of Results}\label{sec:supp_proof}

In this section, we provide detailed proofs of the results stated in the main paper.

\subsection{Proof of \cref{thm.rand_walk_int}}
\label{sect:proof_thm:rand_walk_int}

\begin{proof}
We observe that for $0<\beta<1$ they possess the properties, the coefficients $c_k$, $b_m$ defined in \cref{eq.ckbm}  satisfying the following properties \citep{gorenflo2002time}.
\begin{align*}
& \sum_{k=1}^{\infty} c_k=1, \quad 1>\beta=c_1>c_2>c_3>\ldots \rightarrow 0,\\
& b_0=1, \quad b_m=1-\sum_{k=1}^m c_k=\sum_{k=m+1}^{\infty} c_k,
1=b_0>b_1>b_2>b_3>\ldots \rightarrow 0.
\end{align*}

From the definition of the transition probability \cref{eq.random_pro}, we have

\begin{align}
&\P(\bR(t_{n+1}) =  \bx_{h})\nn
&=  b_n \P(\bR(t_{0}) = \bx_{h}) + c_n  \P(\bR(t_{1}) = \bx_{h})+ \ldots + c_2  \P(\bR(t_{n-1}) = \bx_{h}) + \nn
& \quad\quad +(c_1-\sigma^\beta) \P(\bR(t_{n}) = \bx_{h}) + \sum_{j=1}^{n} \sigma^\beta\frac{ W_{j h}}{d_j} \P(\bR(t_{n}) = \bx_{j}) \nn
& = b_n \P(\bR(t_{0}) = \bx_{h}) + c_n  \P(\bR(t_{1}) = \bx_{h})+ \ldots + c_2  \P(\bR(t_{n-1}) = \bx_{h}) + \nn
& \quad\quad + c_1 \P(\bR(t_{n}) = \bx_{h})  -\sigma^\beta \P(\bR(t_{n}) = \bx_{h})  + \sum_{j=1}^{N} \sigma^\beta\frac{ W_{j h}}{d_j} \P(\bR(t_{n}) = \bx_{j}). \label{eq.dadfa}
\end{align}
By rearranging, we have
\begin{align*}
&\P(\bR(t_{n+1}) =  \bx_{h})  -\sum_{k=1}^{n} c_k  \P(\bR(t_{n+1-k}) = \bx_{h}) - b_n \P(\bR(t_{0}) = \bx_{h})\\
&=  (-1)^{0}\displaystyle\binom{\beta}{0} \P(\bR(t_{n+1}) =  \bx_{h}) - \sum_{k=1}^{n} (-1)^{k+1}\displaystyle\binom{\beta}{k}  \P(\bR(t_{n+1-k}) = \bx_{h})  \\
&\qquad\qquad - \sum_{k=0}^n(-1)^k\displaystyle\binom{\beta}{k}\P(\bR = \bx_{h})  \nn
&=  \sum_{k=0}^{n} (-1)^{k}\displaystyle\binom{\beta}{k}  \P(\bR(t_{n+1-k}) = \bx_{h})  - \sum_{k=0}^n(-1)^k\displaystyle\binom{\beta}{k}\P(\bR = \bx_{h})  \nn
&=  \sum_{k=0}^{n} (-1)^{k}\displaystyle\binom{\beta}{k} \left[ \P(\bR(t_{n+1-k}) = \bx_{h})  - \P(\bR = \bx_{h})\right]  \nn
& =  -\sigma^\beta  \P(\bR(t_{n}) = \bx_{h})  + \sum_{j=1}^{n} \sigma^\beta\frac{ W_{j h}}{d_j} \P(\bR(t_{n}) = \bx_{j}). 
\end{align*} 

Dividing both sides of the final equality by $\sigma^\beta$, it follows that
\begin{align}  
 &\sum_{k=0}^{n} (-1)^{k}\displaystyle\binom{\beta}{k} \frac{\P(\bR(t_{n+1-k}) = \bx_{h})  - \P(\bR = \bx_{h})}{\sigma^\beta} \nn
  & =  - \P(\bR(t_{n}) = \bx_{h})  + \sum_{j=1}^{N} \frac{ W_{j h}}{d_j} \P(\bR(t_{n}) = \bx_{j}).  \label{eq.tafa}
\end{align}
From the Griinwald-Letnikov fractional derivatives formulation \citep{podlubny1999fractional}[eq. (2.54)], the limit of LHS of \cref{eq.tafa} is
\begin{align}
  \lim_{\substack{\sigma\rightarrow 0\\ n\sigma= t}} \sum_{k=0}^{n}(-1)^k\binom{\beta}{k} \frac{\P(\bR(t_{n+1-k}) = \bx_{h})  - \P(\bR = \bx_{h})}{\sigma^\beta}  = D_t^\beta \P(\bR(t) = \bx_{h}) \equiv [D_t^\beta \bP(t)]_h.
\end{align}
where $\bP(t)\coloneqq\lim_{n \to \infty}\P(\bR(t_{n}))$ and $[D_t^\beta \bP(t)]_h$ denotes the $h$-th element of the vector.
On the other hand, the RHS of \cref{eq.tafa} is
\begin{align}
    - \P(\bR(t_{n}) = \bx_{h})  + \sum_{j=1}^{N} \frac{ W_{j h}}{d_j} \P(\bR(t_{n}) = \bx_{j}) = [-\bL  \P(\bR(t_{n}))]_h
\end{align}
where $\P(\bR(t_{n}))$ is the probability (column) vector with $j$-th element being $\P(\bR(t_{n}) = \bx_{j})$, and $[-\bL  \P(\bR(t_{n}))]_h$ denotes the $h$-th element of the vector $-\bL  \P(\bR(t_n))$.

Putting them together, we have 
\begin{align}
    D_t^\beta \bP(t) = -\bL  \bP(t) \label{eq.pro_diff}
\end{align}
since we assume $t_n=t$ in the limit.

The proof of \cref{thm.rand_walk_int} is now complete.
\end{proof}

\subsection{Proof of \cref{cor.rand_walk_int}}

It directly follows from the the linearity of FDEs and $\bX(0)=\bX = \sum_i {}_i\bP(0)\bx_i$ where recall that the initial probability vector $_i\P(\bR(0))\equiv{}_i\bP(0)$ is represented as a one-hot vector with the $i$-th entry marked as 1. 

\subsection{Proof of \cref{thm.rate}}

Before presenting the formal proof, we aim to provide additional insights and intuition regarding the algebraic convergence from two perspectives.
\begin{itemize}
    \item  Fractional Random Walk Perspective: In a standard random walk, a walker moves to a new position at each time step without delay. However, in a fractional random walk, which is more reflective of our model's behavior, the walker has a probability of revisiting past positions. This revisitation is not arbitrary; it is governed by a waiting time that follows a power-law distribution with a long tail. This characteristic fundamentally changes the walk's dynamics, introducing a memory component and leading to a slower, algebraic rate of convergence. This behavior is intrinsically different from normal random walks, where the absence of waiting times facilitates a quicker, exponential, convergence.

\item  Analytic Perspective: From an analytic perspective, the essential slow algebraic rate primarily stems from the slow convergence of the Mittag-Leffler function towards zero. 
To elucidate this, let us consider the scalar scenario. Recall that the Mittag-Leffler function $E_\beta$ is defined as:
\begin{align*}
E_\beta(z):=\sum_{j=0}^{\infty} \frac{z^j}{\Gamma(j \beta+1)}
\end{align*}
for values of $z$ where the series converges. Specifically, when $\beta=1$,
\begin{align*}
E_1(z)=\sum_{j=0}^{\infty} \frac{z^j}{\Gamma(j+1)}=\sum_{j=0}^{\infty} \frac{z^j}{j !}=\exp (z)
\end{align*}
corresponds to the well-known exponential function. According to [A1, Theorem 4.3.], the eigenfunctions of the Caputo derivative are expressed through the Mittag-Leffler function. In more precise terms, if we define $y(t)$ as
\begin{align*}
y(t):=E_\beta\left(-\lambda t^n\right), \quad t \geq 0,
\end{align*}
it follows that
\begin{align*}
D_t^\beta y(t)=-\lambda y(t).
\end{align*}
Notably, when $\beta=1$, this reduces to $\frac{\mathrm{d} \exp (-\lambda t)}{\mathrm{d} t}=-\lambda \exp (-\lambda t)$.
We examine the behavior of $E_\beta\left(-\lambda t^n\right)$. From \citep{diethelm2010analysis}[Theorem 7.3.], when $0<\beta<1$, it is noted that:

(a) The function $y(t)$ is completely monotonic on $(0, \infty)$.

(b) As $x \rightarrow \infty$,
\begin{align*}
y(t)=\frac{t^{-\beta}}{\lambda \Gamma(1-\beta)}(1+o(1)) .
\end{align*}

Thus, the function $E_\beta\left(-\lambda t^\beta\right)$ converges to zero at a rate of $\Theta\left(t^{-\beta}\right)$. Our paper extends this to the general high-dimensional case by replacing the scalar $\lambda$ with the Laplacian matrix $\mathbf{L}$, wherein the eigenvalues of $\mathbf{L}$ play a critical role analogous to $\lambda$ in the scalar case.

For a diagonalizable Laplacian matrix $\mathbf{L}$, the proof essentially reverts to the scalar case as outlined above (refer to \cref{eq.fdsfdfff} in our paper). However, in scenarios where $\mathbf{L}$ is non-diagonalizable and has a general Jordan normal form, it becomes necessary to employ the Laplace transform technique to demonstrate that the algebraic rate remains valid (refer to the context between \cref{eq.fdsfdfff} and \cref{eq.trfsed}).

\end{itemize}

\begin{proof}
We first prove the stationary probability $\bpi=\left( \frac{d_1}{\sum_{j=1}^{N}d_j}, \ldots, \frac{d_N}{\sum_{j=1}^{N}d_j}\right)$ by induction. Assume that for $i=1,\ldots,n$, the probability distribution $\P(\bR(t_{n}))$ always equals $\bpi\T$. For $i=n+1$, from \cref{eq.dadfa}, it follows that
\begin{align*}
[\P(\bR(t_{n+1}))]_h & = \P(\bR(t_{n+1})=\bx_{h}) \\
& = b_n \P(\bR(t_{0}) = \bx_{i}) + \sum_{k}c_k  \P(\bR(t_{n+1-k}) = \bx_{h}) \\
&\quad  \quad -\sigma^\beta \P(\bR(t_{n}) = \bx_{h})  + \sum_{j=1}^{N} \sigma^\beta\frac{ W_{j h}}{d_j} \P(\bR(t_{n}) = \bx_{j}) \\
&=  \bpi_{h} b_n +\sum_{k=1}^{n} \bpi_{h}c_k - \bpi_{h} \sigma^\beta  +  \sum_{j=1}^{N}\bpi_{j} \sigma^\beta\frac{ W_{jh}}{d_{j}} \\
& = \bpi_{h} (b_n +\sum_{k=1}^{n} c_k) - \bpi_{h} \sigma^\beta  +  \sum_{j=1}^{N} \frac{d_j}{\sum_{j=1}^{N}d_j} \sigma^\beta\frac{ W_{jh}}{d_{j}} \\
& = \bpi_{h} - \bpi_{h} \sigma^\beta  + \sigma^\beta \sum_{j=1}^{N} \frac{W_{jh}}{\sum_{j=1}^{N}d_j} \\
& = \bpi_{h} - \bpi_{h} \sigma^\beta  + \sigma^\beta  \frac{d_h}{\sum_{j=1}^{N}d_j} \\
& = \bpi_{h}.
\end{align*}
This proves the existence of stationary probability. The uniqueness follows from this observation: if $\P(\bR(t_{1}))=\bpi'\ne \bpi$, we do not have $\P(\bR(t_{2}))= \P(\bR(t_{1}))$ since otherwise it indicates that the Markov chain defined by 
\begin{align}
&\P(\bR(t_{n+1})=\bx_{j_{n+1}} \given \bR(t_0)=\ldots,\bR(t_1)=\ldots,\ldots,\bR(t_n)=\bx_{j_{n}}) \\
&=\P(\bR(t_{n+1})=\bx_{j} \given \bR(t_n)=\bx_{i} ) \\
&=\P(\bR(t_{2})=\bx_{j} \given \bR(t_1)=\bx_{i} )  \\
& =
\begin{cases}
c_1-\sigma^\beta + b_1 & \text { if staying at current location with } j= i \\
\sigma^\beta\frac{ W_{ij}}{d_{i}}& \text { if jumping to neighboring nodes with } j\ne j
\end{cases}
\end{align}
has stationary distribution other than $\bpi$, which contradicts the assumption of a strongly connected and aperiodic graph.

We next establish the algebraic convergence as $0<\beta<1$. 

It is evident that for the matrix $\bW \bD^{-1}$, given that it is column stochastic and the graph is strongly connected and aperiodic, the Perron-Frobenius theorem \citep{horn2012matrix}[Lemma 8.4.3., Theorem 8.4.4] confirms that the value $1$ is the unique eigenvalue of this matrix that equals its spectral radius, which is also $1$. Consequently, it follows that the matrix $\bL = \bI - \bW \bD^{-1}$ has an eigenvalue of 0, with all other eigenvalues possessing positive real parts.
Considering the Jordan canonical form of $\bL$, denoted as $\bL=\bS\bJ\bS^{-1}$, it is observed that $\bJ$ contains a block that consists solely of a single $0$, while the other blocks are characterized by eigenvalues $\lambda_k$ possessing positive real parts.

We rewrite \cref{eq.zdfva} as 
\begin{align}
    D_t^\beta \bY(t) = -\bJ \bY(t) \label{eq.pro_diff_trans}
\end{align}
where $\bS^{-1}\bP(t) = \bY(t)\in \Real^N$ representing a transformation of the feature space, and the transformed initial condition is defined as $\bS^{-1}\bP(0)= \bY(0)$. 

If the matrix $\bL$ is diagonalizable, then its Jordan canonical form $\bJ$ becomes a diagonal matrix, with the diagonal elements representing the eigenvalues of $\bL$. In this scenario, the differential equation can be decoupled into a set of independent equations, each described by 
\begin{align}
    D_t^\beta \bY_k(t) = -\lambda_k \bY_k(t) \label{eq.fdsfdfff}. 
\end{align}
Here, $\bY_k$ signifies the $k$-th component of the vector $\bY$.
According to  \citep{diethelm2010analysis}[Theorem 4.3.], the solution to each differential equation in the given context is represented as:
\begin{align}
  \bY_k(t) =  \bY_k(0) E_\beta(-\lambda_kt^{\beta}) 
\end{align}
where is $E_\beta(\cdot)$ is the Mittag-Leffler function define as  $E_\beta(z) = \sum_{j=0}^{\infty} \frac{z^j}{\Gamma(\beta j+1)}$  and $\Gamma(\cdot)$ is the gamma function. This formulation leads to two important observations:
\begin{enumerate}[topsep=0pt, itemsep=0pt, partopsep=0pt, parsep=0pt,leftmargin=10pt]
    \item For the index $j$ such that the eigenvalue $\lambda_j = 0$, the solution simplifies to $\bY_j(t) \equiv \bY_j(0)$. This corresponds to a \emph{stationary vector} $\mathbf{v}$ in the original space when transformed back to $\bP(t)$. \label{gadsf}
    \item According to \citep{podlubny1999fractional}[Theorem 1.4.], for indices $k \neq j$, since $\lambda_k$ has a positive real part, the convergence to zero is characterized by the following order:
$$\bY_k(t)= \Theta(t^{-\beta}).$$
\end{enumerate}
Asymptotically, this indicates that all components $\bY_k(t)$, except $\bY_j(t)$, will converge to zero at an algebraic rate. In terms of $\bP(t)$, this translates into a convergence towards a stationary vector in the eigenspace corresponding to the eigenvalue $0$, while components associated with other eigenspaces diminish at an algebraic rate.
 
If the matrix $\bJ$ is not diagonal, the entries of $\bY(t)$ corresponding to distinct Jordan blocks in $\bJ$ remain uncoupled. Therefore, it suffices to consider a single Jordan block corresponding to a non-zero eigenvalue $\lambda_k$. In this case, employing the Laplace transform technique becomes useful for demonstrating that the algebraic rate of convergence remains valid.
We assume the Jordan block $\bJ(\lambda_k)$, associated with $\lambda_k$, is of size $m$. For simplicity, we assume without loss of generality that it is the first block, which simplifies the subscript $\bY_i$. 
It follows that for this Jordan block we have 
\begin{align*}
\begin{aligned}
 D_t^\beta \bY_1(t) & =-\lambda_k \bY_1(t)-\bY_2(t), \\
&  \vdots  \quad \vdots \\
 D_t^\beta \bY_{m-1}(t) & =-\lambda_k \bY_{m-1}(t)-\bY_m(t), \\
 D_t^\beta \bY_m(t) & =-\lambda_k \bY_m(t),
\end{aligned}
\end{align*}
which can be solved from the bottom up. Beginning with the last equation, we obtain:
\begin{align*}
\bY_m(t)=\bY_m(0) E_\beta(-\lambda_k t^{\beta}) = \Theta(t^{-\beta}).
\end{align*}
Further, the differential equation for $\bY_{m-1}(t)$ is given by:
\begin{align*}
D_t^\beta \bY_{m-1}(t)=-\lambda_k \bY_{m-1}(t)-\bY_m(0) E_\beta(-\lambda_k t^{\beta})
\end{align*}
Applying the Laplace transform and referring to \cref{eq.fract_lap}, we obtain:
\begin{align*}
\calL\left\{D_t^\beta \bY_{m-1}(t)\right\}=s^\beta Y_{m-1}(s)-s^{\beta-1} \bY_{m-1}\left(0\right)
\end{align*}
where $Y_{m-1}(s)$ is the Laplace transform of $\bY_{m-1}(t)$.
For the right-hand side of the differential equation, we have $\calL\left\{\lambda_k \bY_{m-1}(t)\right\}=\lambda_k Y_{m-1}(s)$. Additionally, the Laplace transform of the Mittag-Leffler function $E_\beta\left(-\lambda_k t^\beta\right)$ known to be  $\frac{s^{\beta-1}}{s^\beta+\lambda_k}$ \citep{podlubny1999fractional}[eq 1.80].
Consequently, the equation in the Laplace domain is represented as:
\begin{align*}
s^\beta Y_{m-1}(s)-s^{\beta-1} \bY_{m-1}\left(0\right)=-\lambda_k Y_{m-1}(s)-\bY_m(0) \frac{s^{\beta-1}}{s^\beta+\lambda_k}
\end{align*}
Rearranging this equation to isolate $Y_{m-1}(s)$ yields:
\begin{align*}
Y_{m-1}(s)=\frac{s^{\beta-1} \bY_{m-1}\left(0\right)-\bY_m(0) \frac{s^{\beta-1}}{s^\beta+\lambda_k}}{s^\beta+\lambda_k}
\end{align*}
As $s \rightarrow 0$, it follows that $Y_{m-1}(s) = \Theta(s^{\beta-1}$). Applying the same process recursively, we find that $Y_{i}(s) = \Theta(s^{\beta-1})$ for all $i = 1, \ldots, m$.
Invoking the Hardy–Littlewood Tauberian theorem \citep{feller1991introduction}, we can conclude that for all indices $i = 1, \ldots, m$, the following relationship holds:
\begin{align}
\bY_{i}(t) = \Theta(t^{-\beta}). \label{eq.trfsed}
\end{align}
Consequently, we can deduce that, akin to the scenarios involving diagonalizable matrices, the feature components associated with other eigenspaces in non-diagonalizable cases also diminish at an algebraic rate.

Note that we have proved that $\bpi$ is the unique equilibrium point in the probability space. Consequently, the stationary vector $\mathbf{v}$ referenced in \cref{gadsf} must be $\bpi$. This indicates that all $\bP(t)$, assuming $\bP(0) \neq \bpi$, converge to $\bpi$ at the rate of $\Theta(t^{-\beta})$, i.e.,
 \begin{align}
      \left\|\bP(t) - \bpi\T \right\|_2  \sim  \Theta( t^{-\beta}). 
\end{align}
Our next objective is to rigorously prove that $\mathbf{v} \equiv \bpi$ for any initial $\bP(0)$.

Given that the matrix $\bW \bD^{-1}$ is column stochastic, we observe the following:
\begin{align}
\mathbf{1} \bL &= \mathbf{1} (\bI - \bW \bD^{-1}) = \mathbf{1} - \mathbf{1} = \mathbf{0}
\end{align}
where $\mathbf{1}$ denotes an all-ones row vector and $\mathbf{0}$ an all-zeros row vector. This implies that $\mathbf{1}\T$ is a left eigenvector of $\bL$ associated with the eigenvalue $\lambda_j = 0$.
Moreover,
\begin{align}
   \bL\bpi\T= (\bI- \bW \bD^{-1})\bpi\T= \bpi\T - \bpi\T= \mathbf{0}\T.
\end{align}
This means that $\bpi\T$ is a right eigenvector associated with the eigenvalue $\lambda_j=0$.

According to \citep{horn2012matrix}[Theorem 1.4.7.(b)], we therefore can let $\bS=[\cdots\bpi\T\cdots]$ with $\bpi\T$ being the $j$-th column of $\bS$, and correspondingly $\bS^{-1}=[\cdots\mathbf{1}\T\cdots]\T$ with $\mathbf{1}$ being the $j$-th row of $\bS^{-1}$.  
For any probability vector $\bP(0)$, we have $\mathbf{1}\bP(0) = 1$, and we observe that $\bY_j(t)\equiv\bY_j(0)=[\bS^{-1}\bP(0)]_j\equiv 1$. Consequently, as $\bY(t)$ is transformed by $\bS$, we have $\bS \bY(t)$ converges to $j$-th column of $\bS$, which is $\bpi\T$, since all other components $\bY_k(t)$, $k\ne j$, converge to zero with the order $\Theta(t^{-\beta})$.

The proof now is complete.

\end{proof}

\section*{Limitations}
Our research proposes an advanced graph diffusion framework that integrates \emph{time-fractional derivatives}, effectively encompassing many GNNs. Nonetheless, it presents certain limitations. A crucial element we have overlooked is the application of the \emph{fractional derivative in the spatial domain}. In fractional diffusion equations, this implies substituting the standard second-order spatial derivative with a Riesz-Feller derivative \citep{gorenflo2003fractional}, thus modeling a random walk with space-based long-range jumps. Incorporating such a space-fractional diffusion equation within GNNs could potentially alleviate issues like the bottleneck and over-squashing highlighted in \citep{alonbottleneck}. This represents a current limitation of our work and suggests a compelling future research trajectory that merges both time and space fractional derivatives in GNNs.

\section*{Broader Impact}
The introduction of FROND holds significant potential for applications such as sensor networks, transportation, and manufacturing. FROND's ability to encapsulate long-term memory in neural dynamical processes can enhance the representation of complex interconnections, improving predictive modeling and efficiency. This could lead to more responsive sensor networks, optimized routing in transportation, and improved visibility into manufacturing process networks. However, the advent of FROND and similar models may also have mixed labor implications. While these technologies might render certain repetitive tasks obsolete, potentially displacing jobs, they may also generate new opportunities focused on developing and maintaining such advanced systems. Moreover, the shift from mundane tasks could enable workers to focus more on strategic and creative roles, enhancing job satisfaction and productivity.
It's paramount that the deployment of FROND is done ethically, with ample support for reskilling those whose roles may be affected. This helps ensure that the broader impact of this technology is beneficial to society as a whole.

\end{document}